\documentclass[11pt]{article}

\usepackage[preprint]{acl}

\usepackage{times}
\usepackage{latexsym}
\usepackage{graphicx}
\usepackage[T1]{fontenc}
\usepackage{multirow} 

\usepackage[utf8]{inputenc}

\usepackage{microtype}

\usepackage{inconsolata}

\usepackage{graphicx}
\usepackage{float}

\usepackage{fancyhdr}
\usepackage{adjustbox}
\usepackage{booktabs}
\usepackage[table]{xcolor} 
\usepackage{array}         
\definecolor{dkgreen}{rgb}{0.1, 0.5, 0.1}
\definecolor{dkred}{rgb}{0.7, 0.1, 0.1}
\usepackage[table]{xcolor}
\usepackage{colortbl}
\usepackage{amsmath}
\usepackage[greek, english]{babel}
\usepackage[T1]{fontenc}
\usepackage{amssymb}
\usepackage{tabularx}

\usepackage{listings}

\definecolor{backcolour}{rgb}{0.97,0.97,0.97}
\definecolor{bordercolor}{rgb}{0.85,0.85,0.85}

\lstdefinelanguage{json}{
    basicstyle=\ttfamily\small,
    columns=fullflexible,
    showstringspaces=false,
    backgroundcolor=\color{backcolour},
    frame=single,
    rulecolor=\color{bordercolor},
    breaklines=true,
    breakatwhitespace=true,
    tabsize=2
}

\usepackage[table]{xcolor} 
\usepackage{tcolorbox}      
\tcbuselibrary{breakable, skins}

\definecolor{directbg}{HTML}{FFF2CC} 
\definecolor{pebg}{HTML}{D9EAD3}     
\definecolor{multibg}{HTML}{F4CCCC}  

\newcommand{\CPA}[1]{\cellcolor[rgb]{0.173,0.627,0.173}\textcolor{white}{#1}}  
\newcommand{\CML}[1]{\cellcolor[rgb]{0.565,0.788,0.365}{#1}}                   
\newcommand{\CWK}[1]{\cellcolor[rgb]{0.859,0.937,0.718}{#1}}                   
\newcommand{\CZE}[1]{\cellcolor[rgb]{1.000,1.000,0.749}{#1}}                   
\newcommand{\CWN}[1]{\cellcolor[rgb]{0.996,0.769,0.502}{#1}}                   
\newcommand{\CSN}[1]{\cellcolor[rgb]{0.843,0.188,0.153}\textcolor{white}{#1}}  

\tcbset{promptbase/.style={
    enhanced, 
    breakable,
    boxrule=1pt,
    colframe=black!85!white, 
    coltitle=white,          
    fonttitle=\bfseries\large, 
    sharp corners,          
    boxsep=4pt,
    left=6pt, right=6pt,    
    top=6pt, bottom=6pt,    
    after skip=1.5em,       
}}

\newtcolorbox{directbox}{
    promptbase,
    title={Task 1. Single Translation Prompt Example (\texttt{en-it\_IT})},
    colback=directbg
}

\newtcolorbox{directbox_2}{
    promptbase,
    title={\small{Validation Instance Examples}},
    colback=directbg
}

\newtcolorbox{pebox}{
    promptbase,
    title={Task 3. Post-Edition Prompt Example (\texttt{en-el\_GR})},
    colback=pebg
}

\newtcolorbox{multibox}{
    promptbase,
    title={Task 2. Iterative Translation Prompt Example (\texttt{en-fr\_FR})},
    colback=multibg
}

\title{Testing the Deliteralization Hypothesis \\
in Human and Machine Translation}

 \author{Malik Marmonier \quad Rachel Bawden \quad Benoît Sagot\\
         Inria, Paris, France\\ \{\texttt{firstname.lastname\}@inria.fr}}

\begin{document}
\maketitle
\begin{abstract}
The recent shift from dedicated NMT systems to general-purpose LLMs has reshaped machine translation, with LLMs reported to produce more fluent, less literal output than their predecessors. We test whether this shift extends to the deliteralization hypothesis---the long-standing claim from translation studies that translations become progressively less literal as they are drafted and revised. Using the WMT24++ dataset, we compare the literality of human translations and post-editions to that of two NMT systems and six LLMs across 54 language pairs and three tasks: direct translation, iterative self-revision, and post-editing of human drafts. Literality is measured via a validated Synthetic Literality Index built from six heuristics. We find that (i) human translations remain significantly less literal than those of all tested MT systems, though recent LLMs narrow the gap; (ii) when prompted to iteratively revise their own output, LLMs deliteralize monotonically, providing the first evidence that the hypothesis applies natively to LLM generation; and (iii) as post-editors, LLMs invert the revision triggers of human post-editors, tolerating literal drafts and targeting idiomatic human formulations for revision.
\end{abstract}

\section{Introduction}

Linguistic meaning is compositional. It emerges out of sequences of lexical and grammatical morphemes, which emerge, in turn, out of sequences of phones. The anisomorphy of these morphemes between languages---likely to be greater when the languages considered are neither phylogenetically related nor in contact---means that equivalent meanings in translation must often be determined at a scale higher than the ``word.''

Literality, however, subverts this fundamental principle and has been a topic of contention among translation scholars for centuries \cite{shuttleworth1997dictionary}, \textit{literally}. From Jerome's admonition to translate ``sense for sense'' rather than ``word for word'' \cite{jerome1893pammachius}, to the German Romantics' praise of ``transparent'' translations that let the source language ``shine through'' the target text \cite{benjamin1968task,schleiermacher2012methods}, from \citeauthor{dryden1681preface}'s \shortcite{dryden1681preface} disdain for ``servile, literal translation'' to \citeauthor{venuti1995invisibility}'s \shortcite{venuti1995invisibility} critique of fluency as an instrument of ethnocentric cultural domestication, few attitudes and expectations regarding translation cannot be reduced to matters of literality.

In machine translation (MT) research, literality has, likewise, been a topic of sustained interest, as a defining feature of the dreaded ``translationese'' that long characterized the unnatural output of MT systems \cite{zhang-toral-2019-effect,toral-2019-post,graham-etal-2020-statistical,dankers-etal-2022-transformer,luo-etal-2024-diverge,li-etal-2025-lost}.

\begin{figure}[t]
 \centering
 \includegraphics[width=0.75\columnwidth]{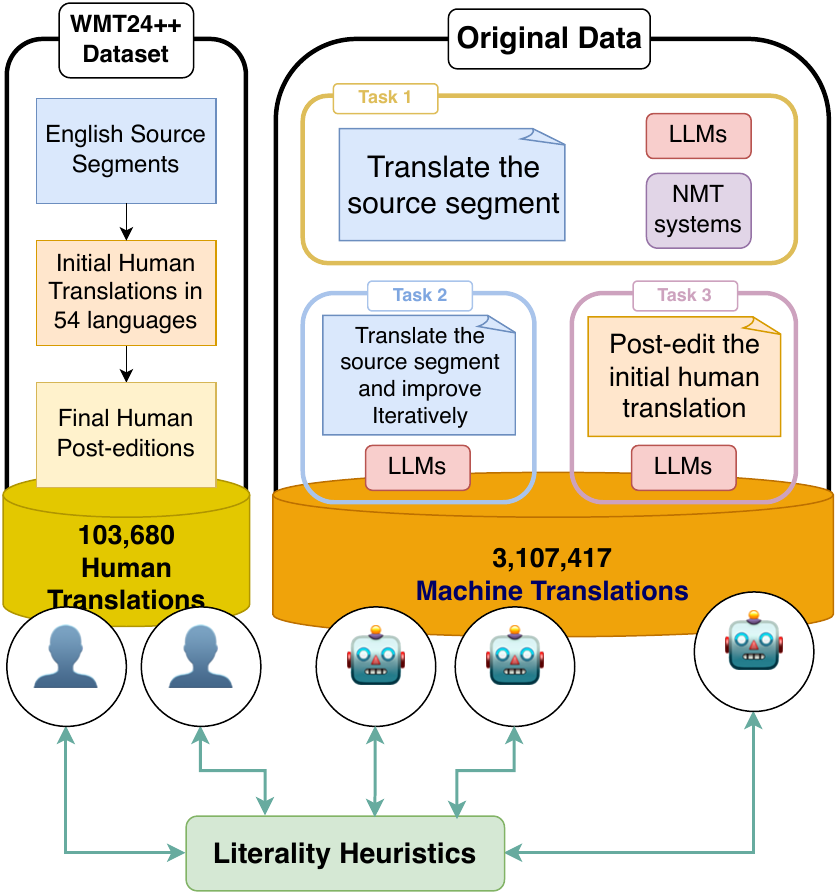}
 \caption{We test the deliteralization hypothesis by comparing the literality of human and LLM translations across 54 languages. LLMs were prompted to perform single translation (Task 1), iterative translation (Task 2), and post-editing of initial human drafts (Task 3). Literality is measured via a weighted ensemble of literality heuristics.}\label{fig:synoptic}
\end{figure}

The present work studies literality in human and machine translation through the lens of the widely-discussed deliteralization hypothesis \cite{chestermanReflections}. Also called the ``literal translation hypothesis,'' it posits that translations, as they are drafted and revised, become progressively less literal. While empirical investigations taking human translators as subjects have provided substantial support for this hypothesis, recent research has also shed light on an opposite ``reliteralizing'' tendency in human translators \cite{PavlovicAntunDesc,borg2022literary}, and the hypothesis---to the best of our knowledge---has never been explored in an MT context.

Our study takes the WMT24++ dataset \cite{deutsch-etal-2025-wmt24} as a starting point. This dataset has the peculiar property of comprising both the ``initial'' human translations of its English source segments and their final human-post-edited version in 54 languages. We obtain translation hypotheses from a variety of Large Language Models (LLMs), representing the state-of-the-art of MT paradigms, alongside two ``traditional'' Neural MT (NMT) systems as a point of reference, prompting both to output a single translation of the source segment (task 1). Additionally, we task the LLMs with iteratively revising their own source translations (task 2), and with post-editing the dataset's initial human translations into the 54 target languages (task 3). We select candidate heuristics to automatically evaluate the degree of literality displayed by a translation in relation to its source segment. We validate these heuristics using a toy corpus of English idioms translated both literally and ``idiomatically'' into diverse languages, as well as human annotators for a subset of the actual data, and devise a synthetic literality index based on a weighted ensemble of these heuristics. We estimate the relative degrees of literality displayed by the human translations (initial and post-edited) and the output of the various MT systems with respect to the source segment, in order to test the deliteralization hypothesis.

Our findings indicate that (i) MT systems--both NMT and LLM-based--produce significantly more literal translations than human translators, though recent LLMs narrow this gap considerably; (ii)~when prompted to iteratively revise their own output, LLMs deliteralize monotonically across successive attempts, providing the first direct evidence that the deliteralization hypothesis applies natively to LLM generation; and (iii) as post-editors of human drafts, LLMs invert the revision priorities of human professionals, tolerating literal drafts and targeting idiomatic human formulations for revision, even as their actual edits remain predominantly deliteralizing in direction. We release our data and code under CC-BY SA and MIT licenses, respectively.

\section{Related Work}
The deliteralization hypothesis, also known as the ``literal translation hypothesis,'' has been widely discussed and empirically tested as a diachronic model of the translation process. The concept has roots in the work of Ivir \shortcite{ivir-formal}, who posited that translators initially seek formal correspondence, only resorting to freer structural or semantic shifts when literal equivalents fail to ensure functional equivalence. Building on this cognitive framework, Tirkkonen-Condit \shortcite{TirkkonenCondit2005TheMM} proposed a ``monitor model,'' arguing that literal translation serves as an automated, default rendering procedure that naturally persists until a cognitive monitor flags a textual issue and prompts conscious revision towards non-literality. The hypothesis was further formalized by Chesterman \shortcite{chestermanReflections}, who advocates for the term ``deliteralization'' to explicitly describe this shift from more literal to less literal draft versions during text production. Empirical investigations have provided substantial support for these theoretical models; for instance, Englund Dimitrova \shortcite{englundDimitrovaExpertise} relied on think-aloud protocols and keystroke logging to demonstrate that translators' syntactic revisions generally move the target text further away from the source text's structure. More recently, Pavlović and Antunović \shortcite{PavlovicAntunDesc} tested the hypothesis by analyzing the ``distance dynamics'' of self-revisions among professional translators and interpreters. While their keystroke-logging data confirmed that deliteralizing changes---moves from more literal to freer renderings---constitute the majority of distance-altering revisions, they also highlighted the complexity of the editing process by revealing a significant occurrence of distance-neutral changes, as well as reliteralizing moves that actually pull the translation back closer to the source text. A similar trend towards reliteralization was notably documented in a case study by Borg \shortcite{borg2022literary} in the context of literary translation.

Within Natural Language Processing (NLP) and MT, researchers' interest in literality is closely linked to the study of ``translationese'' \cite{NidaTaber1969,Gellerstam1986}---a phenomenon most saliently characterized by overly literal, word-for-word translations. 
To better handle parallel corpora and improve MT systems, researchers have developed models to automatically detect translationese at the sentential-level \cite{kurokawa-etal-2009-automatic,koppelTranslationese,riley-etal-2020-translationese}, and literality at the sub-sentential level \cite{zhai-etal-2018-construction,zhai-etal-2019-classification,zhai-etal-2020-detecting}. NMT architectures have thus been shown to exhibit a strong bias toward overly literal outputs. For example, analyses of Transformer-based NMT models \cite{NIPS2017_3f5ee243} reveal that they struggle with non-compositional phrases like idioms \cite{dankers-etal-2022-transformer}. Consequently, MT outputs consistently exhibit more one-to-one token alignments with source texts than human translations, a discrepancy partly attributable to the algorithmic bias of decoding methods such as beam search \cite{luo-etal-2024-diverge}. Recently, the advent of LLMs has shifted these dynamics; research indicates that LLMs generally produce significantly less literal and more fluent translations compared to traditional NMT systems, particularly when handling idiomatic expressions \cite{raunak-etal-2023-gpts}. Nevertheless, LLMs are not entirely immune to translationese---they still generate unexpectedly literal and unnatural renderings \cite{li-etal-2025-lost}. To address this lingering literalism, recent studies propose new mitigation strategies, such as filtering out unnatural training instances and using LLMs to polish training references, thereby encouraging more fluent and target-language-consistent generation \cite{riley-etal-2020-translationese,dutta-chowdhury-etal-2022-towards,li-etal-2025-lost}.

\section{Data}

The following two subsections present the datasets we relied on and compiled in order to test the deliteralization hypothesis in human and machine translation.

\begin{table}[t]
    \centering
    \begin{adjustbox}{width=0.60\columnwidth}
    \begin{tabular}{ll|ll}
        \toprule
        \multicolumn{4}{c}{\bf Target Languages and Locales Represented in WMT24++ } \\
        \midrule
\texttt{ar\_EG} & Arabic (Egypt) & \texttt{lv\_LV} & Latvian (Latvia) \\
\texttt{ar\_SA} & Arabic (Saudi Arabia) & \texttt{ml\_IN} & Malayalam (India) \\
\texttt{bg\_BG} & Bulgarian (Bulgaria) & \texttt{mr\_IN} & Marathi (India) \\
\texttt{bn\_IN} & Bengali (India) & \texttt{nl\_NL} & Dutch (Netherlands) \\
\texttt{ca\_ES} & Catalan (Spain) & \texttt{no\_NO} & Norwegian (Norway) \\
\texttt{cs\_CZ} & Czech (Czechia) & \texttt{pa\_IN} & Punjabi (India) \\
\texttt{da\_DK} & Danish (Denmark) & \texttt{pl\_PL} & Polish (Poland) \\
\texttt{de\_DE} & German (Germany) & \texttt{pt\_BR} & Portuguese (Brazil) \\
\texttt{el\_GR} & Greek (Greece) & \texttt{pt\_PT} & Portuguese (Portugal) \\
\texttt{es\_MX} & Spanish (Mexico) & \texttt{ro\_RO} & Romanian (Romania) \\
\texttt{et\_EE} & Estonian (Estonia) & \texttt{ru\_RU} & Russian (Russia) \\
\texttt{fa\_IR} & Farsi (Iran) & \texttt{sk\_SK} & Slovak (Slovakia) \\
\texttt{fi\_FI} & Finnish (Finland) & \texttt{sl\_SI} & Slovenian (Slovenia) \\
\texttt{fil\_PH} & Filipino (Philippines) & \texttt{sr\_RS} & Serbian (Serbia) \\
\texttt{fr\_CA} & French (Canada) & \texttt{sv\_SE} & Swedish (Sweden) \\
\texttt{fr\_FR} & French (France) & \texttt{sw\_KE} & Swahili (Kenya) \\
\texttt{gu\_IN} & Gujarati (India) & \texttt{sw\_TZ} & Swahili (Tanzania) \\
\texttt{he\_IL} & Hebrew (Israel) & \texttt{ta\_IN} & Tamil (India) \\
\texttt{hi\_IN} & Hindi (India) & \texttt{te\_IN} & Telugu (India) \\
\texttt{hr\_HR} & Croatian (Croatia) & \texttt{th\_TH} & Thai (Thailand) \\
\texttt{hu\_HU} & Hungarian (Hungary) & \texttt{tr\_TR} & Turkish (Turkey) \\
\texttt{id\_ID} & Indonesian (Indonesia) & \texttt{uk\_UA} & Ukrainian (Ukraine) \\
\texttt{it\_IT} & Italian (Italy) & \texttt{ur\_PK} & Urdu (Pakistan) \\
\texttt{ja\_JP} & Japanese (Japan) & \texttt{vi\_VN} & Vietnamese (Vietnam) \\
\texttt{kn\_IN} & Kannada (India) & \texttt{zh\_CN} & Mandarin (China) \\
\texttt{ko\_KR} & Korean (South Korea) & \texttt{zh\_TW} & Mandarin (Taiwan) \\
\texttt{lt\_LT} & Lithuanian (Lithuania) & \texttt{zu\_ZA} & Zulu (South Africa) \\
        \bottomrule
    \end{tabular}
    \end{adjustbox}
    \caption{List of the 54 target languages and locales in WMT24++ used in our experiments.}
    \label{tab:languages}
\end{table}

\subsection{Human Translations}

This study fundamentally relies on the WMT24++ dataset \cite{deutsch-etal-2025-wmt24}, itself built upon the much smaller test set of the WMT24 General Machine Translation Shared Task \cite{kocmi-etal-2024-findings}. It comprises 960 valid
English source segments of various lengths (from short sentences to paragraphs), representing four text domains (literary, news, social media, audio transcripts), their initial human translation into 54 languages,
as well as the final human-post-edited version of this translation. This ``dual-state'' translation data provides an interesting testbed for observing how translations are revised, refined, and potentially deliteralized or reliteralized during the post-editing phase.\footnote{We note that some translation studies scholars \citep[\textit{inter alia}][]{doCarmoMoorkens2020} reserve the term ``post-editing'' to the post-edition of MT output, and the term ``revision'' to the post-edition of human translations. We do not follow this convention and use the terms ``revision'' and ``post-editing'' interchangeably.}

Naturally, the human post-editors chose to preserve the initial translations when judging them acceptable. An average of 42.8\% of the segments were altered during post-edition across the language pairs, though in proportions that greatly vary from one target language to the next (see Appendix \ref{app:sameShare}). Consequently, WMT24++ is also well-suited for studying whether a segment's initial degree of literality correlates with its likelihood of revision.

Table \ref{tab:languages} presents a list of target languages and translation locales represented in the dataset.

\subsection{Machine Translations}
In order to compare the literality of human translations and post-editions to that of current MT output, we selected a total of 6 LLMs, from 3 model families: \texttt{\textbf{Qwen-3-32B}} \cite{yang2025qwen3technicalreport}, a reasoning LLM with 32 billion parameters, and \texttt{\textbf{Qwen-3-8B}}, its distilled 8-billion-parameter variant; \texttt{\textbf{Olmo-3-32B-Think}} \cite{olmo2025olmo3}, a reasoning LLM with 32 billion parameters, and \texttt{\textbf{Olmo-3-32B-Think-SFT}}, its ``primitive'' pre-reinforcement-learning-finetuning variant; \texttt{\textbf{Gemma-4-31B-it}} \cite{gemma4modelcard2026}, a reasoning LLM with 31 billion parameters, and \texttt{\textbf{Gemma-4-E4B-it}}, an edge-optimized variant with an effective inference footprint of 4 billion parameters. We also used \texttt{\textbf{NLLB-200-Distilled-600M}} and \texttt{\textbf{NLLB-200-3.3B}} \cite{nllbteam2022languageleftbehindscaling}, two ``traditional'' encoder-decoder models with 600 million and 3.3 billion parameters, respectively, as a point of reference for NMT systems.

\begin{table}[ht!]
\centering
\small
\setlength{\tabcolsep}{3pt}
\renewcommand{\arraystretch}{1.15}
\resizebox{0.8\columnwidth}{!}{%
\begin{tabular}{lrrrrrrr}
\toprule
\textbf{System} & \multicolumn{1}{c}{\textit{Task 1}} & \multicolumn{5}{c}{\textit{Task 2 Iterative Trans. (\# position)}} & \multicolumn{1}{c}{\textit{Task 3}} \\
\cmidrule(lr){2-2} \cmidrule(lr){3-7} \cmidrule(lr){8-8}
\phantom{System} & \textbf{Trans.} & \textbf{\#{}$1$} & \textbf{\#{}$2$} & \textbf{\#{}$3$} & \textbf{\#{}$4$} & \textbf{\#{}$5$} & \textbf{PE} \\
\midrule
Human & \cellcolor[rgb]{0.689,0.865,0.183}\textcolor[rgb]{0,0,0}{3.015} & — & — & — & — & — & \cellcolor[rgb]{0.8, 0.8, 0.8} \\
NLLB-600M$^{\text{beam}}$ & \cellcolor[rgb]{0.197,0.712,0.479}\textcolor[rgb]{0,0,0}{8.541} & — & — & — & — & — & — \\
NLLB-3.3B$^{\text{beam}}$ & \cellcolor[rgb]{0.312,0.768,0.416}\textcolor[rgb]{0,0,0}{6.865} & — & — & — & — & — & — \\
NLLB-600M$^{\text{anc}}$ & \cellcolor[rgb]{0.168,0.460,0.558}\textcolor[rgb]{1,1,1}{15.301} & — & — & — & — & — & — \\
NLLB-3.3B$^{\text{anc}}$ & \cellcolor[rgb]{0.138,0.537,0.555}\textcolor[rgb]{1,1,1}{13.269} & — & — & — & — & — & — \\
Qwen3-8B$^{-}$ & \cellcolor[rgb]{0.214,0.722,0.470}\textcolor[rgb]{0,0,0}{8.266} & — & — & — & — & — & \cellcolor[rgb]{0.658,0.860,0.203}\textcolor[rgb]{0,0,0}{3.284} \\
Qwen3-8B$^{+}$ & \cellcolor[rgb]{0.220,0.726,0.466}\textcolor[rgb]{0,0,0}{8.140} & \cellcolor[rgb]{0.246,0.739,0.452}\textcolor[rgb]{0,0,0}{7.734} & \cellcolor[rgb]{0.253,0.742,0.448}\textcolor[rgb]{0,0,0}{7.704} & \cellcolor[rgb]{0.246,0.739,0.452}\textcolor[rgb]{0,0,0}{7.777} & \cellcolor[rgb]{0.328,0.774,0.407}\textcolor[rgb]{0,0,0}{6.692} & \cellcolor[rgb]{0.506,0.829,0.300}\textcolor[rgb]{0,0,0}{4.727} & \cellcolor[rgb]{0.637,0.857,0.217}\textcolor[rgb]{0,0,0}{3.508} \\
Qwen3-32B$^{-}$ & \cellcolor[rgb]{0.328,0.774,0.407}\textcolor[rgb]{0,0,0}{6.645} & — & — & — & — & — & \cellcolor[rgb]{0.658,0.860,0.203}\textcolor[rgb]{0,0,0}{3.255} \\
Qwen3-32B$^{+}$ & \cellcolor[rgb]{0.361,0.786,0.388}\textcolor[rgb]{0,0,0}{6.254} & \cellcolor[rgb]{0.378,0.792,0.378}\textcolor[rgb]{0,0,0}{6.077} & \cellcolor[rgb]{0.378,0.792,0.378}\textcolor[rgb]{0,0,0}{6.146} & \cellcolor[rgb]{0.404,0.800,0.363}\textcolor[rgb]{0,0,0}{5.771} & \cellcolor[rgb]{0.431,0.808,0.346}\textcolor[rgb]{0,0,0}{5.553} & \cellcolor[rgb]{0.497,0.826,0.306}\textcolor[rgb]{0,0,0}{4.879} & \cellcolor[rgb]{0.647,0.858,0.210}\textcolor[rgb]{0,0,0}{3.378} \\
OLMo-32B$^{+}$ & \cellcolor[rgb]{0.125,0.640,0.527}\textcolor[rgb]{1,1,1}{10.519} & \cellcolor[rgb]{0.158,0.684,0.502}\textcolor[rgb]{0,0,0}{9.308} & \cellcolor[rgb]{0.191,0.708,0.482}\textcolor[rgb]{0,0,0}{8.674} & \cellcolor[rgb]{0.186,0.705,0.485}\textcolor[rgb]{0,0,0}{8.788} & \cellcolor[rgb]{0.162,0.687,0.499}\textcolor[rgb]{0,0,0}{9.271} & \cellcolor[rgb]{0.143,0.669,0.511}\textcolor[rgb]{1,1,1}{9.707} & \cellcolor[rgb]{0.478,0.821,0.318}\textcolor[rgb]{0,0,0}{5.071} \\
OLMo-32B-SFT$^{+}$ & \cellcolor[rgb]{0.132,0.655,0.520}\textcolor[rgb]{1,1,1}{10.107} & \cellcolor[rgb]{0.202,0.715,0.476}\textcolor[rgb]{0,0,0}{8.456} & \cellcolor[rgb]{0.226,0.729,0.463}\textcolor[rgb]{0,0,0}{8.093} & \cellcolor[rgb]{0.239,0.736,0.456}\textcolor[rgb]{0,0,0}{7.874} & \cellcolor[rgb]{0.253,0.742,0.448}\textcolor[rgb]{0,0,0}{7.694} & \cellcolor[rgb]{0.274,0.752,0.437}\textcolor[rgb]{0,0,0}{7.419} & \cellcolor[rgb]{0.413,0.803,0.357}\textcolor[rgb]{0,0,0}{5.728} \\
Gemma4-E4B$^{-}$ & \cellcolor[rgb]{0.478,0.821,0.318}\textcolor[rgb]{0,0,0}{4.984} & — & — & — & — & — & \cellcolor[rgb]{0.668,0.862,0.196}\textcolor[rgb]{0,0,0}{3.138} \\
Gemma4-E4B$^{+}$ & \cellcolor[rgb]{0.506,0.829,0.300}\textcolor[rgb]{0,0,0}{4.736} & \cellcolor[rgb]{0.487,0.824,0.312}\textcolor[rgb]{0,0,0}{4.914} & \cellcolor[rgb]{0.506,0.829,0.300}\textcolor[rgb]{0,0,0}{4.718} & \cellcolor[rgb]{0.487,0.824,0.312}\textcolor[rgb]{0,0,0}{4.923} & \cellcolor[rgb]{0.459,0.816,0.330}\textcolor[rgb]{0,0,0}{5.245} & \cellcolor[rgb]{0.468,0.819,0.324}\textcolor[rgb]{0,0,0}{5.123} & \cellcolor[rgb]{0.678,0.864,0.190}\textcolor[rgb]{0,0,0}{3.047} \\
Gemma4-31B$^{-}$ & \cellcolor[rgb]{0.627,0.855,0.223}\textcolor[rgb]{0,0,0}{3.581} & — & — & — & — & — & \cellcolor[rgb]{0.699,0.867,0.176}\textcolor[rgb]{0,0,0}{2.898} \\
Gemma4-31B$^{+}$ & \cellcolor[rgb]{0.647,0.858,0.210}\textcolor[rgb]{0,0,0}{3.320} & \cellcolor[rgb]{0.637,0.857,0.217}\textcolor[rgb]{0,0,0}{3.502} & \cellcolor[rgb]{0.647,0.858,0.210}\textcolor[rgb]{0,0,0}{3.353} & \cellcolor[rgb]{0.637,0.857,0.217}\textcolor[rgb]{0,0,0}{3.487} & \cellcolor[rgb]{0.606,0.851,0.237}\textcolor[rgb]{0,0,0}{3.762} & \cellcolor[rgb]{0.576,0.845,0.256}\textcolor[rgb]{0,0,0}{4.101} & \cellcolor[rgb]{0.710,0.869,0.169}\textcolor[rgb]{0,0,0}{2.762} \\
\midrule
\textit{N (all)} & 775,805 & 283,021 & 280,259 & 235,093 & 108,470 & 65,277 & 517,335 \\
\textit{N (MX$\leq$7)} & 482,289 & 194,272 & 195,691 & 164,543 & 65,839 & 35,347 & 473,905 \\
\textit{N (MX$\leq$5)} & 365,865 & 145,160 & 148,071 & 123,665 & 46,256 & 23,776 & 412,109 \\
\textit{N (MX$\leq$3)} & 210,698 & 80,344 & 83,524 & 68,171 & 23,890 & 12,187 & 270,557 \\
\bottomrule
\end{tabular}%
}
\caption{Mean MetricX-24 scores (lower is better) by system and task, computed against the final human-post-edited references in WMT24++, averaged over all 54 target languages. \textit{Single Trans.}: single translation. \textit{Iterative \#1--\#5}: interim translation positions from the iterative translation prompt; \textit{Post-ed.}: post-edition of initial human translations. $^{+}$/$^{-}$ = thinking on/off; anc.\ = ancestral sampling;  — = system was not involved.}
\label{tab:metricx_all_tasks}
\end{table}

For the NLLB models, inference was performed in two ways, once using beam search with a beam size of 4, and once using ancestral sampling \cite{luo-etal-2024-diverge}. Target languages were specified using the NLLB language-specific prefix tokens (e.g., \texttt{ell\_Grek} for Modern Greek) to force the start of the decoded sequence.

For the LLMs, we designed three distinct prompt templates which we populated with segments from the WMT24++ dataset, with each template representing one of three tasks. Prompt wording was carefully neutral with respect to literality: instructions referred only to translation ``quality'' and ``improvement,'' avoiding any explicit framing that could bias models toward either literal or free renderings. \textbf{Task 1 (Single Translation):} Models were instructed to provide a single translation in the target language, with the full source document context included in each prompt. \textbf{Task 2 (Iterative Translation):} Models were prompted to generate between 2 and 5 translations, with each subsequent version ideally improving upon the last, with the full source document context included in each prompt. \textbf{Task 3 (Post-Editing):} Models acted as post-editors of the initial WMT24++ human translation while having access to both the full source document context and the full human-translated initial context.\footnote{For the Qwen-3 and Gemma-4 models, we introduced a secondary variable by devising sets of prompts that explicitly encouraged reasoning, and sets of prompts that forbade it, asking for an immediate output. Olmo models were consistently tested with reasoning-encouraging prompts.} A total of 311,040 prompts were generated. We present examples of the populated prompt templates for each task in Appendix~\ref{app:prompts}. 

LLM inference was conducted using the vLLM engine \cite{kwon2023efficient} and the Transformers library \cite{wolf-etal-2020-transformers}. We used greedy decoding across all tasks, additionally sampling eight outputs per prompt (with temperature 1.0) for the French of France locale in Task 2. Further details regarding data generation and extraction can be found in Appendix~\ref{app:dataGeneration}.

A grand total of 3,107,417 translations were successfully extracted from the MT systems' output, including 723,965 translations in the 54 target languages for Task~1 (99.75\% average extraction rate), 1,866,117 translations for Task~2, and 517,335 post-editions of the initial human translations for task 3 (99.79\% average extraction rate).

We provide overall MetricX-24\footnote{metricx-24-hybrid-xxl-v2p6-bfloat16} scores \cite{juraska-etal-2024-metricx} for the queried systems in Table~\ref{tab:metricx_all_tasks}, with a per-language-pair breakdown and pairwise bootstrap significance testing of mean differences with N=10,000 resamples \citep{koehn-2004-statistical} in Appendix~\ref{app:LPmx24}. MetricX-24 is an error rate ranging from 0 (best) to 25 (worst). While the question as to whether judgments of literality apply to erroneous translation is debated in translation studies \cite{chestermanReflections}, in light of past work \cite{luo-etal-2024-diverge}, we decided to only apply our analysis to translations scoring below a MetricX-24 threshold of 5.0, after manual auditing of data samples.

\section{Methods}
Studies of literality in human translation appear to have mostly relied on manual annotations and the qualitative analyses of human researchers, ensuring their validity but limiting their scope \cite{PavlovicAntunDesc}, with few works borrowing simple NLP heuristics, such as part-of-speech (POS) and character $n$-grams, and applying them to comparable corpora rather than parallel corpora \cite{volanskyFeatures2015}.

Studies of literality in MT have likewise relied on human annotations \cite{li-etal-2025-lost}, using them, at times, to further train automatic taggers \cite{zhai-etal-2018-construction,zhai-etal-2019-classification,zhai-etal-2020-detecting} with uncertain applications to out-of-domain language pairs and text domains. Other studies have relied on controlled testsets with rule-based evaluation and a very narrow scope \cite{dankers-etal-2022-transformer}, whilst other studies have relied on heuristics of various kinds to detect literality \cite{raunak-etal-2023-gpts,luo-etal-2024-diverge}; we follow in their footsteps.

For the present work, we identified seven literality heuristics inspired by previous work or devised by us, validated them, and combined them into a synthetic literality index.

\subsection{Candidate Literality Heuristics}
\label{sec:heuristics}

We tested seven literality heuristics, each capturing a distinct structural or semantic relationship between a source segment~$s$ and its translation hypothesis~$h$.

\paragraph{POS sequence similarity (\textsc{PosSim}).}
Source and hypothesis are each parsed with Stanza \cite{qi-etal-2020-stanza} to obtain their universal POS tag sequences, $P_s = (p_1^s, \ldots, p_m^s)$ and $P_h = (p_1^h, \ldots, p_n^h)$, the edit-distance similarity of these sequences is then computed as the normalized longest-common-subsequence ratio (LCS)\footnote{Computed via the \texttt{difflib} Python library's SequenceMatcher algorithm \cite{python_difflib}.} of these sequences:

\small
\begin{equation}
  \textsc{PosSim}(s, h)
  = \frac{2 \, \lvert\mathrm{LCS}(P_s, P_h)\rvert}{|P_s| + |P_h|}.
\end{equation}
\normalsize

\paragraph{Dependency tree similarity (\textsc{TreeSim}).}
Stanza dependency parses of $s$ and $h$ are each converted into a set of labelled arc types of the form \texttt{HEAD\_POS\,-\,deprel\,-\,DEP\_POS}
(e.g.\ \texttt{VERB-nsubj-NOUN}). Tree similarity is then the Jaccard index over these sets:

\small
\begin{equation}
  \textsc{TreeSim}(s, h) = \frac{|R_s \cap R_h|}{|R_s \cup R_h|},
\end{equation}
\normalsize
where $R_s$ and $R_h$ are the relation-type sets of $s$ and~$h$ respectively.

\paragraph{Alignment density (\textsc{Density}).}
We align $s$ and $h$ at the word level using SimAlign \cite{jalili-sabet-etal-2020-simalign} with the \textit{itermax} algorithm on top of XLM-R \cite{conneau-etal-2020-unsupervised} contextual embeddings.
Let $A \subseteq \{1,\ldots,|s|\} \times \{1,\ldots,|h|\}$ be the resulting
alignment set. Density is then defined as the proportion of tokens that participate in at least one alignment, normalized by the length of the longer sequence:

\small
\begin{equation}
  \textsc{Density}(s, h) = \frac{|A|}{\max(|s|, |h|)}.
\end{equation}
\normalsize

\paragraph{Alignment crossings (\textsc{Crossings}).}
Using the same word-level alignment set $A$ generated by SimAlign, we quantify the extent to which the linear word order is preserved between the source and target segments. Let the elements of $A$ be represented as a sequence of alignment edges $(i_k, j_k)$, sorted primarily by their source token index $i$ such that $k < m \implies i_k \leq i_m$. The crossings metric counts the number of times these alignment links intersect:
\begin{equation}
\small
  \textsc{Crossings}(s, h) = \sum_{k=1}^{|A|-1} \sum_{m=k+1}^{|A|} \mathbb{I}(j_k > j_m),
\end{equation}
\normalsize
where $\mathbb{I}$ is the indicator function that equals 1 if the condition is true and 0 otherwise.

\paragraph{Segment-level semantic similarity (\textsc{SegSemLaBSE}).}
The full source and hypothesis strings are encoded with LaBSE \cite{feng-etal-2022-language} into unit-normalized sentence embeddings $\mathbf{e}_s$ and $\mathbf{e}_h$.
Their cosine similarity,

\small
\begin{equation}
  \textsc{SegSemLaBSE}(s, h) = \frac{\mathbf{e}_s \cdot \mathbf{e}_h}
                                     {\|\mathbf{e}_s\|\,\|\mathbf{e}_h\|},
\end{equation}\normalsize
captures overall cross-lingual semantic similarity.

\paragraph{Token-level lexical similarity (\textsc{TokSimRaw} and \textsc{TokSimPen}).}
For each aligned word pair $(w_s, w_h) \in A$, we compute the LaBSE cosine similarity of the two word embeddings.
The raw signal averages these pair-wise similarities:

\small
\begin{equation}
  \textsc{TokSimRaw}(s, h) =
  \frac{1}{|A|} \sum_{(w_s,\, w_h) \in A} \cos(\mathbf{e}_{w_s},\,\mathbf{e}_{w_h}).
\end{equation}
\normalsize
Because high raw similarity can arise even when large portions of either string
are left unaligned, we additionally penalize by alignment density:

\small
\begin{equation}
  \textsc{TokSimPen}(s, h)
  = \textsc{TokSimRaw}(s, h) \times \textsc{Density}(s, h).
\end{equation}
\normalsize

\subsection{Heuristics Validation}
To validate these heuristics, we devised a dataset consisting of English idioms paired with both highly literal, and fluent idiomatic translations across nine target languages, drawing upon our own linguistic competence and lexicographical resources. The resulting ``base'' validation dataset comprises 163 of these one-sentence triplets (source, literal translation, idiomatic translation), distributed as follows: Spanish (38), French (20), and 15 each for Arabic, Chinese, Italian, Japanese, Modern Greek, Russian, and Turkish. We generated multi-sentence segments by randomly concatenating the base triplets. Taking a subset of $N=2,700$ three-segment concatenations, we also generated four target permutations for each instance: 100\% literal (three literal segments concatenated as the translation), 66\% literal (two literal, one idiomatic), 33\% literal (one literal, two idiomatic), and 0\% literal (the idiomatic translation). The resulting augmented validation dataset comprises 10,963 instances.

\scriptsize
\begin{directbox_2}
\{\texttt{\textbf{``source'':}} ``I have other fish to fry today.'',
  \texttt{\textbf{``literal'':}} ``\textit{J'ai d'autres poissons à frire aujourd'hui.}'',
  \texttt{\textbf{``idiomatic'':}} ``\textit{J'ai d'autres chats à fouetter aujourd'hui.}'',
  \texttt{\textbf{``tgt\_lang'':} ``fr''}\}, \ldots
\\
  \{\texttt{\textbf{``source'':}} ``Stop beating around the bush and answer the question. It is raining cats and dogs outside. I have other fish to fry today.'',
  \texttt{\textbf{``100\%\_literal'':}} ``\textit{Arrête de battre autour du buisson et réponds à la question. Il pleut des chats et des chiens dehors. J'ai d'autres poissons à frire, aujourd'hui.}'',
  \texttt{\textbf{``66\%\_literal'':}} ``\textit{Arrête de battre autour du buisson et réponds à la question. Il pleut des cordes dehors. J'ai d'autres poissons à frire, aujourd'hui.}'',
  \texttt{\textbf{``33\%\_literal'':}} ``\textit{Arrête de tourner autour du pot et réponds à la question. Il pleut des cordes dehors. J'ai d'autres poissons à frire, aujourd'hui.}'',
  \texttt{\textbf{``0\%\_literal'':}} ``\textit{Arrête de tourner autour du pot et réponds à la question. Il pleut des cordes dehors. J'ai d'autres chats à fouetter, aujourd'hui.}'',
  \texttt{\textbf{``tgt\_lang'':} ``fr''}\}
\end{directbox_2}
\normalsize

\begin{figure}[t]
 \centering
 \includegraphics[width=\columnwidth]{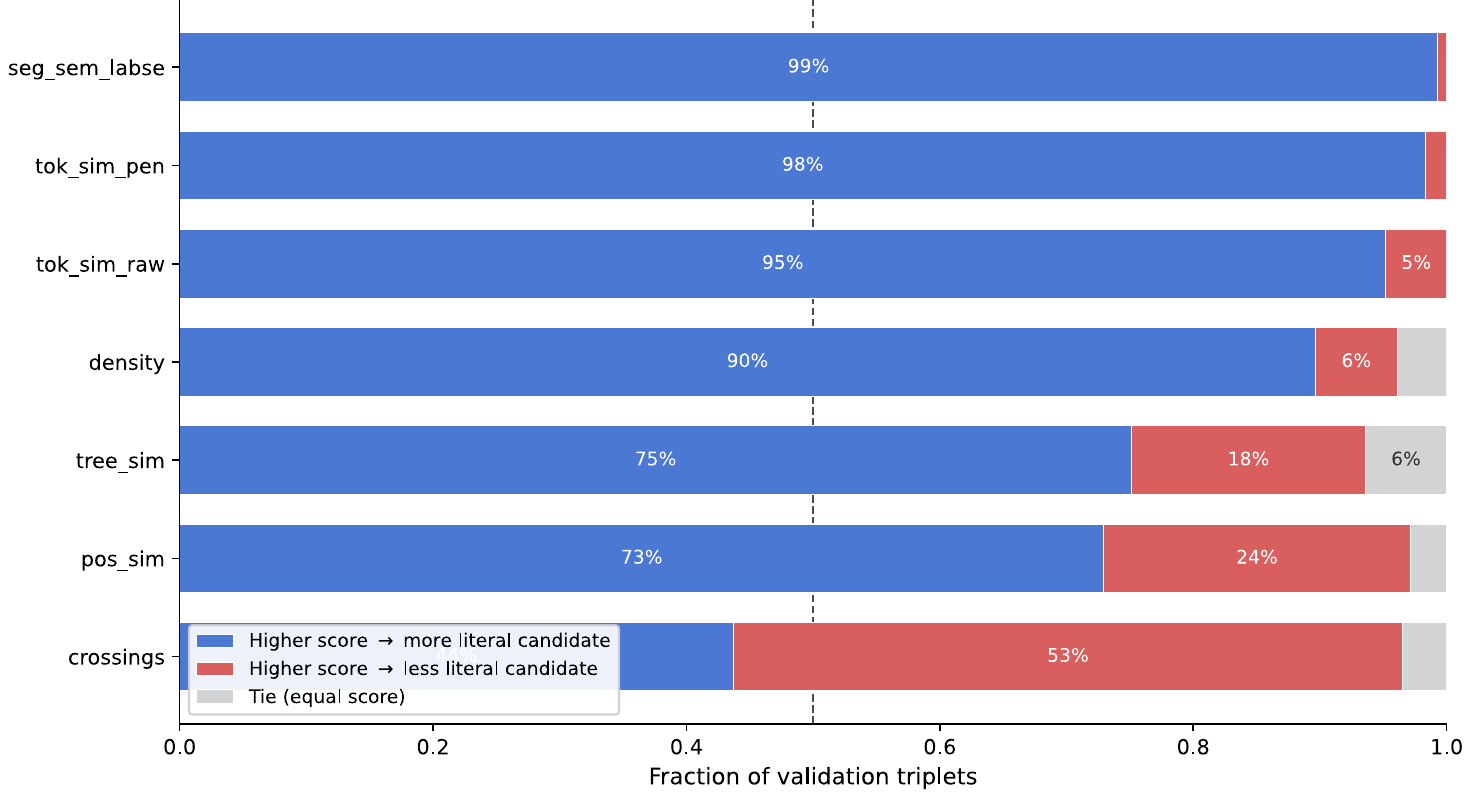}
 \caption{Signal breakdown of literality heuristics across $N=10,963$ validation triplets. The chart displays the fraction of instances where a given heuristic assigned a strictly higher score to the literal candidate (blue), the idiomatic candidate (red), or produced a tie (grey).}\label{fig:signalBreakdown}
\end{figure}

\begin{figure}[t]
 \centering
 \includegraphics[width=\columnwidth]{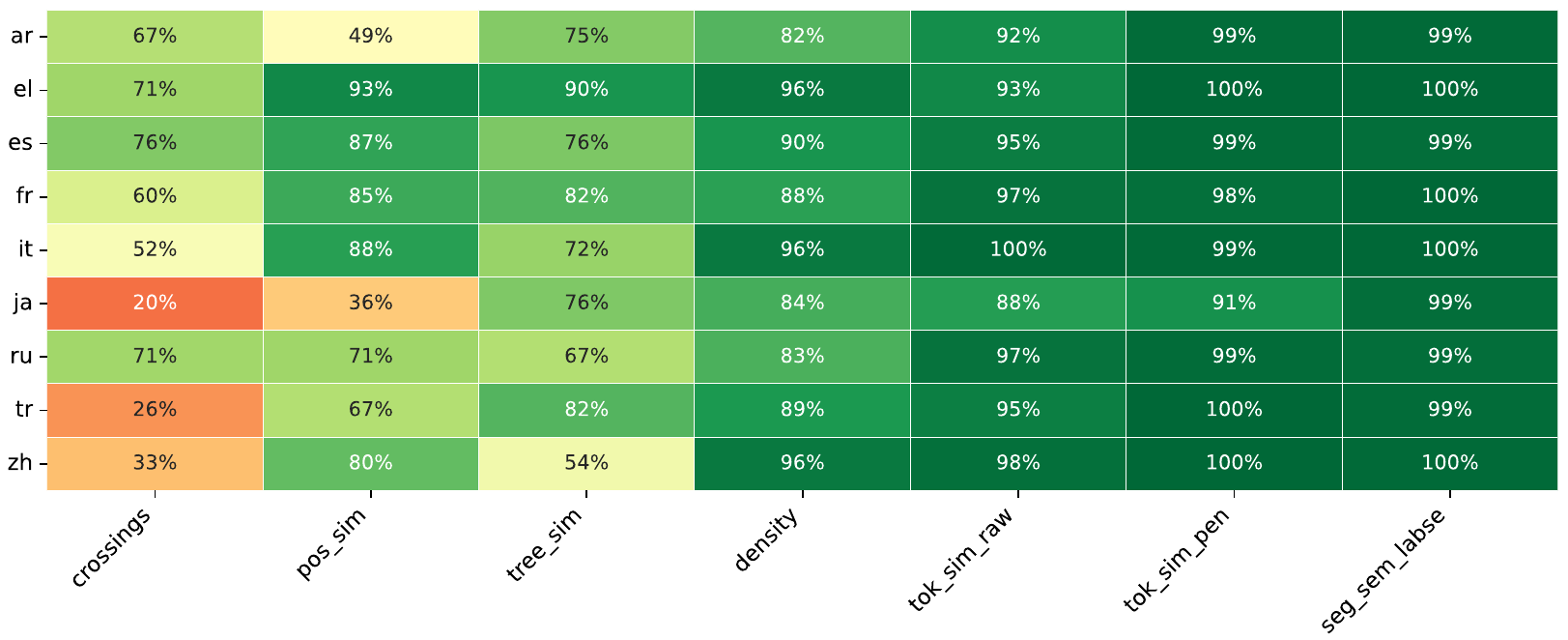}
 \caption{Heuristic hit rates broken down across the nine target languages in the validation set. Values represent the percentage of validation triplets where the heuristic's dominant direction correctly allowed to distinguish the literal translation from the idiomatic one. 
 }
\label{fig:hitrateByLanguage}
\end{figure}

\begin{table}[t]
\centering\scriptsize
\setlength{\tabcolsep}{4pt} 
\renewcommand{\arraystretch}{1.2}
\resizebox{0.66\columnwidth}{!}{
\begin{tabular}{l c c c c c c}
\toprule
 & & \multicolumn{4}{c}{\textbf{Degree of Literality}} & \\
\cmidrule(lr){3-6}
\textbf{Heuristic} & \textbf{$N$} & \textbf{100\%} & \textbf{66\%} & \textbf{33\%} & \textbf{0\%} & \textbf{$p$-value} \\
\midrule
seg\_sem\_labse & 2700 & \cellcolor[rgb]{0.993,0.906,0.144}\textcolor[rgb]{0,0,0}{0.870} & \cellcolor[rgb]{0.260,0.745,0.444}\textcolor[rgb]{0,0,0}{0.837} & \cellcolor[rgb]{0.177,0.438,0.558}\textcolor[rgb]{1,1,1}{0.801} & \cellcolor[rgb]{0.267,0.005,0.329}\textcolor[rgb]{1,1,1}{0.762} & $<$0.001 \\
tok\_sim\_raw & 2700 & \cellcolor[rgb]{0.993,0.906,0.144}\textcolor[rgb]{0,0,0}{0.783} & \cellcolor[rgb]{0.267,0.749,0.441}\textcolor[rgb]{0,0,0}{0.768} & \cellcolor[rgb]{0.177,0.438,0.558}\textcolor[rgb]{1,1,1}{0.751} & \cellcolor[rgb]{0.267,0.005,0.329}\textcolor[rgb]{1,1,1}{0.732} & $<$0.001 \\
tok\_sim\_pen & 2700 & \cellcolor[rgb]{0.993,0.906,0.144}\textcolor[rgb]{0,0,0}{0.673} & \cellcolor[rgb]{0.214,0.722,0.470}\textcolor[rgb]{0,0,0}{0.636} & \cellcolor[rgb]{0.191,0.407,0.556}\textcolor[rgb]{1,1,1}{0.598} & \cellcolor[rgb]{0.267,0.005,0.329}\textcolor[rgb]{1,1,1}{0.560} & $<$0.001 \\
density & 2700 & \cellcolor[rgb]{0.993,0.906,0.144}\textcolor[rgb]{0,0,0}{0.860} & \cellcolor[rgb]{0.208,0.719,0.473}\textcolor[rgb]{0,0,0}{0.828} & \cellcolor[rgb]{0.191,0.407,0.556}\textcolor[rgb]{1,1,1}{0.797} & \cellcolor[rgb]{0.267,0.005,0.329}\textcolor[rgb]{1,1,1}{0.766} & $<$0.001 \\
pos\_sim & 2700 & \cellcolor[rgb]{0.993,0.906,0.144}\textcolor[rgb]{0,0,0}{0.559} & \cellcolor[rgb]{0.246,0.739,0.452}\textcolor[rgb]{0,0,0}{0.538} & \cellcolor[rgb]{0.182,0.426,0.557}\textcolor[rgb]{1,1,1}{0.516} & \cellcolor[rgb]{0.267,0.005,0.329}\textcolor[rgb]{1,1,1}{0.493} & $<$0.001 \\
tree\_sim & 2700 & \cellcolor[rgb]{0.993,0.906,0.144}\textcolor[rgb]{0,0,0}{0.444} & \cellcolor[rgb]{0.186,0.705,0.485}\textcolor[rgb]{0,0,0}{0.424} & \cellcolor[rgb]{0.198,0.392,0.555}\textcolor[rgb]{1,1,1}{0.405} & \cellcolor[rgb]{0.267,0.005,0.329}\textcolor[rgb]{1,1,1}{0.386} & $<$0.001 \\
crossings & 2700 & \cellcolor[rgb]{0.886,0.892,0.095}\textcolor[rgb]{0,0,0}{33.62} & \cellcolor[rgb]{0.993,0.906,0.144}\textcolor[rgb]{0,0,0}{33.57} & \cellcolor[rgb]{0.141,0.530,0.556}\textcolor[rgb]{1,1,1}{34.18} & \cellcolor[rgb]{0.267,0.005,0.329}\textcolor[rgb]{1,1,1}{34.70} & $<$0.001 \\
\bottomrule
\end{tabular}
}
\caption{Gradient of literality heuristics across mixed multi-segment translations. Color scaling is applied row-wise to highlight the gradient effect. The $p$-value represents the result of a Friedman test \cite{friedman_test} evaluating the significance of the difference across all four literality levels.
}
\label{tab:mixed_translation_gradient_rowwise}
\end{table}

\paragraph{Binary hit rates and Monotonicity.} We systematically measured the capacity of each of the candidate heuristics to distinguish between the literal and the free, idiomatic translations in our validation triplets. Figure \ref{fig:signalBreakdown} shows the fraction of triplets for which each heuristic assigned a strictly higher score to either the literal or the idiomatic candidate. It indicates that the embedding-based and alignment-density heuristics are remarkably reliable discriminators of literality in our toy corpus (most notably \textsc{seg\_sem\_labse} and \textsc{tok\_sim\_pen}). The ``syntactic heuristics,'' \textsc{tree\_sim} (75\% hit rate) and \textsc{pos\_sim} (73\%), while still providing a substantial literality signal, show a higher rate of ties and inversions, and the \textsc{crossings} heuristic is the weakest. Figure~\ref{fig:hitrateByLanguage}, which disaggregates the heuristics' hit rates by target language reveals the cause of this weaker performance: \textsc{crossings}, and to a lesser extent \textsc{pos\_sim} and \textsc{tree\_sim}, are not robust to the typological distance between the English source and the target language. We further evaluated the candidate heuristics' sensitivity to more granular degrees of literality. The results, presented in Table~\ref{tab:mixed_translation_gradient_rowwise}, demonstrate that all heuristics except \textsc{crossings} do reflect the degree of literality in a text in their scores: as the proportion of idiomatic text increases, these metrics exhibit a smooth, monotonic cascade downward.

\paragraph{Human validation.} We conducted a human annotation campaign on a subset of our actual data, collecting pairwise judgments of literality. The annotations confirmed the trends revealed by our validation dataset (see Appendix~\ref{app:humanval}).

\paragraph{Synthetic Literality Index (SLI).}
To synthesize these heuristic signals into a unified comparable score for our experiments, we introduce the \textit{Synthetic Literality Index} (SLI). Excluding the underperforming \textsc{crossings}, each score $x_i$ of the six remaining heuristics is first min-max normalized within each language pair (LP), using the empirical range observed across all systems and segments for that LP, giving $Norm_{lp}(x_i) \in [0, 1]$; this controls for the fact that raw signal values vary systematically with source–target typological distance, making cross-LP comparisons otherwise arduous.
The normalized signals are then combined as a weighted sum, where the weight of each signal reflects its predictive validity defined on the basis of the hit rates observed on our validation corpus ($\hat{h} \in \{0.99, 0.98, 0.95, 0.90, 0.75,
0.73\}$ for \textsc{seg\_sem\_labse}, \textsc{tok\_sim\_pen}, \textsc{tok\_sim\_raw}, \textsc{density}, \textsc{tree\_sim}, and \textsc{pos\_sim} respectively), converted to a softmax distribution at temperature
$\tau = 0.5$:

\small
\begin{equation}
  w_i = \frac{\exp(\hat{h}_i / \tau)}{\sum_j \exp(\hat{h}_j / \tau)}.
\end{equation}
\normalsize
The sub-unity temperature sharpens the distribution relative to a uniform baseline, rewarding signals with higher empirical validity without allowing any
single signal to dominate.\footnote{Nine target languages of WMT24++ lack reliable morphosyntactic parsers to the best of our knowledge: Bengali, Filipino, Gujarati, Kannada, Malayalam, Punjabi, Swahili, Thai and Zulu. For WMT24++ language pairs for which no reliable morphosyntactic parser exists and which hence lack \textsc{pos\_sim} and \textsc{tree\_sim} scores, the softmax is recomputed over the available signal subset and renormalized, so that all language pairs can receive an SLI score.} The final index for a given translation hypothesis is thus computed as the weighted sum of its normalized heuristics:

\small
\begin{equation}
\text{SLI} = \sum_{i} w_i Norm_{lp}(x_i),
\end{equation}
\normalsize
and ranges from 0 (maximally free relative to the LP-level distribution) to 1 (maximally literal).

\section{Results}
\label{sec:results}

\begin{table}[ht!]
\centering\small\setlength{\tabcolsep}{4pt}
\renewcommand{\arraystretch}{1.2}
\begin{adjustbox}{max width=0.85\columnwidth}
\begin{tabular}{lr|rrrrr|r}
\toprule
\textbf{System} & \multicolumn{1}{c}{\textit{Task 1}} & \multicolumn{5}{c}{\textit{Task 2 — Iterative (\# position)}} & \multicolumn{1}{c}{\textit{Task 3}} \\
\cmidrule(lr){2-2} \cmidrule(lr){3-7} \cmidrule(lr){8-8}
\phantom{System} & \textbf{Trans.} & \textbf{\#1} & \textbf{\#2} & \textbf{\#3} & \textbf{\#4} & \textbf{\#5} & \textbf{PE} \\
\midrule
Human & \cellcolor[rgb]{0.148,0.512,0.557}\textcolor[rgb]{1,1,1}{0.5819} & — & — & — & — & — & \cellcolor[rgb]{0.164,0.471,0.558}\textcolor[rgb]{1,1,1}{0.5762} \\
NLLB-600M$^{\text{beam}}$ & \cellcolor[rgb]{0.344,0.780,0.397}\textcolor[rgb]{0,0,0}{0.6193} & — & — & — & — & — & — \\
NLLB-3.3B$^{\text{beam}}$ & \cellcolor[rgb]{0.267,0.749,0.441}\textcolor[rgb]{0,0,0}{0.6145} & — & — & — & — & — & — \\
NLLB-600M$^{\text{anc}}$ & \cellcolor[rgb]{0.126,0.571,0.550}\textcolor[rgb]{1,1,1}{0.5897} & — & — & — & — & — & — \\
NLLB-3.3B$^{\text{anc}}$ & \cellcolor[rgb]{0.145,0.519,0.557}\textcolor[rgb]{1,1,1}{0.5825} & — & — & — & — & — & — \\
Qwen3-8B$^{-}$ & \cellcolor[rgb]{0.404,0.800,0.363}\textcolor[rgb]{0,0,0}{0.6230} & — & — & — & — & — & \cellcolor[rgb]{0.123,0.637,0.529}\textcolor[rgb]{1,1,1}{0.5985} \\
Qwen3-8B$^{+}$ & \cellcolor[rgb]{0.312,0.768,0.416}\textcolor[rgb]{0,0,0}{0.6173} & \cellcolor[rgb]{0.274,0.752,0.437}\textcolor[rgb]{0,0,0}{0.6151} & \cellcolor[rgb]{0.186,0.705,0.485}\textcolor[rgb]{0,0,0}{0.6080} & \cellcolor[rgb]{0.143,0.669,0.511}\textcolor[rgb]{1,1,1}{0.6031} & \cellcolor[rgb]{0.121,0.593,0.545}\textcolor[rgb]{1,1,1}{0.5927} & \cellcolor[rgb]{0.176,0.441,0.558}\textcolor[rgb]{1,1,1}{0.5725} & \cellcolor[rgb]{0.122,0.589,0.546}\textcolor[rgb]{1,1,1}{0.5923} \\
Qwen3-32B$^{-}$ & \cellcolor[rgb]{0.267,0.749,0.441}\textcolor[rgb]{0,0,0}{0.6142} & — & — & — & — & — & \cellcolor[rgb]{0.121,0.593,0.545}\textcolor[rgb]{1,1,1}{0.5925} \\
Qwen3-32B$^{+}$ & \cellcolor[rgb]{0.214,0.722,0.470}\textcolor[rgb]{0,0,0}{0.6107} & \cellcolor[rgb]{0.208,0.719,0.473}\textcolor[rgb]{0,0,0}{0.6100} & \cellcolor[rgb]{0.123,0.637,0.529}\textcolor[rgb]{1,1,1}{0.5986} & \cellcolor[rgb]{0.123,0.585,0.547}\textcolor[rgb]{1,1,1}{0.5917} & \cellcolor[rgb]{0.164,0.471,0.558}\textcolor[rgb]{1,1,1}{0.5760} & \cellcolor[rgb]{0.186,0.419,0.557}\textcolor[rgb]{1,1,1}{0.5694} & \cellcolor[rgb]{0.128,0.567,0.551}\textcolor[rgb]{1,1,1}{0.5889} \\
OLMo-32B$^{+}$ & \cellcolor[rgb]{0.143,0.669,0.511}\textcolor[rgb]{1,1,1}{0.6029} & \cellcolor[rgb]{0.162,0.687,0.499}\textcolor[rgb]{0,0,0}{0.6058} & \cellcolor[rgb]{0.123,0.582,0.547}\textcolor[rgb]{1,1,1}{0.5910} & \cellcolor[rgb]{0.126,0.571,0.550}\textcolor[rgb]{1,1,1}{0.5895} & \cellcolor[rgb]{0.131,0.556,0.552}\textcolor[rgb]{1,1,1}{0.5876} & \cellcolor[rgb]{0.131,0.556,0.552}\textcolor[rgb]{1,1,1}{0.5875} & \cellcolor[rgb]{0.123,0.585,0.547}\textcolor[rgb]{1,1,1}{0.5918} \\
OLMo-32B-SFT$^{+}$ & \cellcolor[rgb]{0.208,0.719,0.473}\textcolor[rgb]{0,0,0}{0.6098} & \cellcolor[rgb]{0.208,0.719,0.473}\textcolor[rgb]{0,0,0}{0.6097} & \cellcolor[rgb]{0.121,0.626,0.533}\textcolor[rgb]{1,1,1}{0.5971} & \cellcolor[rgb]{0.119,0.615,0.538}\textcolor[rgb]{1,1,1}{0.5956} & \cellcolor[rgb]{0.120,0.607,0.540}\textcolor[rgb]{1,1,1}{0.5944} & \cellcolor[rgb]{0.120,0.600,0.543}\textcolor[rgb]{1,1,1}{0.5938} & \cellcolor[rgb]{0.122,0.589,0.546}\textcolor[rgb]{1,1,1}{0.5923} \\
Gemma4-31B$^{+}$ & \cellcolor[rgb]{0.120,0.604,0.541}\textcolor[rgb]{1,1,1}{0.5939} & \cellcolor[rgb]{0.147,0.673,0.509}\textcolor[rgb]{1,1,1}{0.6033} & \cellcolor[rgb]{0.165,0.467,0.558}\textcolor[rgb]{1,1,1}{0.5758} & \cellcolor[rgb]{0.184,0.422,0.557}\textcolor[rgb]{1,1,1}{0.5696} & \cellcolor[rgb]{0.264,0.238,0.519}\textcolor[rgb]{1,1,1}{0.5478} & \cellcolor[rgb]{0.267,0.005,0.329}\textcolor[rgb]{1,1,1}{0.5254} & \cellcolor[rgb]{0.126,0.571,0.550}\textcolor[rgb]{1,1,1}{0.5896} \\
Gemma4-31B$^{-}$ & \cellcolor[rgb]{0.120,0.607,0.540}\textcolor[rgb]{1,1,1}{0.5945} & — & — & — & — & — & \cellcolor[rgb]{0.123,0.582,0.547}\textcolor[rgb]{1,1,1}{0.5909} \\
Gemma4-E4B$^{+}$ & \cellcolor[rgb]{0.140,0.666,0.513}\textcolor[rgb]{1,1,1}{0.6026} & \cellcolor[rgb]{0.191,0.708,0.482}\textcolor[rgb]{0,0,0}{0.6085} & \cellcolor[rgb]{0.155,0.493,0.558}\textcolor[rgb]{1,1,1}{0.5792} & \cellcolor[rgb]{0.201,0.384,0.554}\textcolor[rgb]{1,1,1}{0.5649} & \cellcolor[rgb]{0.211,0.364,0.552}\textcolor[rgb]{1,1,1}{0.5624} & \cellcolor[rgb]{0.283,0.105,0.427}\textcolor[rgb]{1,1,1}{0.5347} & \cellcolor[rgb]{0.126,0.571,0.550}\textcolor[rgb]{1,1,1}{0.5898} \\
Gemma4-E4B$^{-}$ & \cellcolor[rgb]{0.246,0.739,0.452}\textcolor[rgb]{0,0,0}{0.6130} & — & — & — & — & — & \cellcolor[rgb]{0.120,0.607,0.540}\textcolor[rgb]{1,1,1}{0.5945} \\
\midrule
Qwen3-8B$^{+}$ (\textsc{fr}, $T\!=\!0$) & \cellcolor[rgb]{0.984,0.905,0.137}\textcolor[rgb]{0,0,0}{0.6517} & \cellcolor[rgb]{0.993,0.906,0.144}\textcolor[rgb]{0,0,0}{0.6524} & \cellcolor[rgb]{0.825,0.885,0.106}\textcolor[rgb]{0,0,0}{0.6437} & \cellcolor[rgb]{0.689,0.865,0.183}\textcolor[rgb]{0,0,0}{0.6374} & \cellcolor[rgb]{0.468,0.819,0.324}\textcolor[rgb]{0,0,0}{0.6266} & \cellcolor[rgb]{0.140,0.666,0.513}\textcolor[rgb]{1,1,1}{0.6024} & — \\
Qwen3-32B$^{+}$ (\textsc{fr}, $T\!=\!0$) & \cellcolor[rgb]{0.856,0.889,0.097}\textcolor[rgb]{0,0,0}{0.6453} & \cellcolor[rgb]{0.886,0.892,0.095}\textcolor[rgb]{0,0,0}{0.6468} & \cellcolor[rgb]{0.647,0.858,0.210}\textcolor[rgb]{0,0,0}{0.6353} & \cellcolor[rgb]{0.497,0.826,0.306}\textcolor[rgb]{0,0,0}{0.6281} & \cellcolor[rgb]{0.378,0.792,0.378}\textcolor[rgb]{0,0,0}{0.6212} & \cellcolor[rgb]{0.336,0.777,0.402}\textcolor[rgb]{0,0,0}{0.6188} & — \\
OLMo-32B$^{+}$ (\textsc{fr}, $T\!=\!0$) & \cellcolor[rgb]{0.846,0.887,0.100}\textcolor[rgb]{0,0,0}{0.6449} & \cellcolor[rgb]{0.825,0.885,0.106}\textcolor[rgb]{0,0,0}{0.6439} & \cellcolor[rgb]{0.378,0.792,0.378}\textcolor[rgb]{0,0,0}{0.6212} & \cellcolor[rgb]{0.369,0.789,0.383}\textcolor[rgb]{0,0,0}{0.6207} & \cellcolor[rgb]{0.369,0.789,0.383}\textcolor[rgb]{0,0,0}{0.6209} & \cellcolor[rgb]{0.352,0.783,0.393}\textcolor[rgb]{0,0,0}{0.6201} & — \\
OLMo-32B-SFT$^{+}$ (\textsc{fr}, $T\!=\!0$) & \cellcolor[rgb]{0.896,0.894,0.096}\textcolor[rgb]{0,0,0}{0.6472} & \cellcolor[rgb]{0.815,0.883,0.110}\textcolor[rgb]{0,0,0}{0.6432} & \cellcolor[rgb]{0.468,0.819,0.324}\textcolor[rgb]{0,0,0}{0.6261} & \cellcolor[rgb]{0.361,0.786,0.388}\textcolor[rgb]{0,0,0}{0.6204} & \cellcolor[rgb]{0.369,0.789,0.383}\textcolor[rgb]{0,0,0}{0.6209} & \cellcolor[rgb]{0.404,0.800,0.363}\textcolor[rgb]{0,0,0}{0.6228} & — \\
Gemma4-31B$^{+}$ (\textsc{fr}, $T\!=\!0$) & \cellcolor[rgb]{0.627,0.855,0.223}\textcolor[rgb]{0,0,0}{0.6341} & \cellcolor[rgb]{0.876,0.891,0.095}\textcolor[rgb]{0,0,0}{0.6463} & \cellcolor[rgb]{0.239,0.736,0.456}\textcolor[rgb]{0,0,0}{0.6124} & \cellcolor[rgb]{0.135,0.659,0.518}\textcolor[rgb]{1,1,1}{0.6018} & \cellcolor[rgb]{0.156,0.490,0.558}\textcolor[rgb]{1,1,1}{0.5787} & — & — \\
Gemma4-E4B$^{+}$ (\textsc{fr}, $T\!=\!0$) & \cellcolor[rgb]{0.876,0.891,0.095}\textcolor[rgb]{0,0,0}{0.6462} & \cellcolor[rgb]{0.984,0.905,0.137}\textcolor[rgb]{0,0,0}{0.6515} & \cellcolor[rgb]{0.267,0.749,0.441}\textcolor[rgb]{0,0,0}{0.6143} & \cellcolor[rgb]{0.121,0.626,0.533}\textcolor[rgb]{1,1,1}{0.5972} & \cellcolor[rgb]{0.125,0.640,0.527}\textcolor[rgb]{1,1,1}{0.5992} & \cellcolor[rgb]{0.214,0.356,0.551}\textcolor[rgb]{1,1,1}{0.5613} & — \\
\midrule
Qwen3-8B$^{+}$ (\textsc{fr}, $T\!=\!1$) & — & \cellcolor[rgb]{0.906,0.895,0.098}\textcolor[rgb]{0,0,0}{0.6475} & \cellcolor[rgb]{0.689,0.865,0.183}\textcolor[rgb]{0,0,0}{0.6373} & \cellcolor[rgb]{0.546,0.838,0.276}\textcolor[rgb]{0,0,0}{0.6303} & \cellcolor[rgb]{0.431,0.808,0.346}\textcolor[rgb]{0,0,0}{0.6243} & \cellcolor[rgb]{0.137,0.662,0.516}\textcolor[rgb]{1,1,1}{0.6022} & — \\
Qwen3-32B$^{+}$ (\textsc{fr}, $T\!=\!1$) & — & \cellcolor[rgb]{0.678,0.864,0.190}\textcolor[rgb]{0,0,0}{0.6368} & \cellcolor[rgb]{0.267,0.749,0.441}\textcolor[rgb]{0,0,0}{0.6147} & \cellcolor[rgb]{0.128,0.648,0.523}\textcolor[rgb]{1,1,1}{0.6002} & \cellcolor[rgb]{0.122,0.589,0.546}\textcolor[rgb]{1,1,1}{0.5919} & \cellcolor[rgb]{0.132,0.552,0.553}\textcolor[rgb]{1,1,1}{0.5873} & — \\
OLMo-32B-SFT$^{+}$ (\textsc{fr}, $T\!=\!1$) & — & \cellcolor[rgb]{0.449,0.814,0.335}\textcolor[rgb]{0,0,0}{0.6254} & \cellcolor[rgb]{0.154,0.680,0.504}\textcolor[rgb]{0,0,0}{0.6045} & \cellcolor[rgb]{0.120,0.604,0.541}\textcolor[rgb]{1,1,1}{0.5941} & \cellcolor[rgb]{0.125,0.574,0.549}\textcolor[rgb]{1,1,1}{0.5901} & \cellcolor[rgb]{0.134,0.549,0.554}\textcolor[rgb]{1,1,1}{0.5868} & — \\
Gemma4-31B$^{+}$ (\textsc{fr}, $T\!=\!1$) & — & \cellcolor[rgb]{0.876,0.891,0.095}\textcolor[rgb]{0,0,0}{0.6460} & \cellcolor[rgb]{0.208,0.719,0.473}\textcolor[rgb]{0,0,0}{0.6102} & \cellcolor[rgb]{0.137,0.662,0.516}\textcolor[rgb]{1,1,1}{0.6022} & \cellcolor[rgb]{0.136,0.541,0.554}\textcolor[rgb]{1,1,1}{0.5855} & \cellcolor[rgb]{0.162,0.687,0.499}\textcolor[rgb]{0,0,0}{0.6054} & — \\
Gemma4-E4B$^{+}$ (\textsc{fr}, $T\!=\!1$) & — & \cellcolor[rgb]{0.906,0.895,0.098}\textcolor[rgb]{0,0,0}{0.6478} & \cellcolor[rgb]{0.154,0.680,0.504}\textcolor[rgb]{0,0,0}{0.6043} & \cellcolor[rgb]{0.153,0.497,0.558}\textcolor[rgb]{1,1,1}{0.5797} & \cellcolor[rgb]{0.167,0.464,0.558}\textcolor[rgb]{1,1,1}{0.5755} & \cellcolor[rgb]{0.132,0.552,0.553}\textcolor[rgb]{1,1,1}{0.5870} & — \\
\bottomrule
\end{tabular}
\end{adjustbox}
\caption{Synthetic Literality Index (SLI, $\tau=0.5$, per-LP normalisation) by system and task. \textit{Trans.}: single direct translation (Task~1, all 54~LPs). \textit{\#1--\#5}: iterative positions (Task~2, $T\!=\!0$, all LPs). \textit{PE}: post-edition of human translation (Task~3); FR rows show \textsc{en-fr\textsubscript{fr}} only. Higher SLI = more literal.}
\label{tab:sli_combined}
\end{table}

Table \ref{tab:sli_combined} reports the SLI scores across all systems and tasks. In direct translation (Task 1), human references establish the least literal baseline (SLI 0.5819). Traditional encoder-decoder systems (NLLB with beam search) and the Qwen-3 family exhibit marked literalism (0.61--0.62), corroborating the well-documented translationese bias of MT systems. The Gemma-4 models, conversely, approximate the human degree of literality much more closely (0.59--0.60). Thinking mode has no significant effect on literality for Qwen3 or Gemma4-31B in direct translation.\footnote{Gemma4-E4B is the exception: the no-thinking variant produces significantly more literal output than its thinking counterpart ($\Delta$SLI = +0.011, p$<$0.001), suggesting that the extended reasoning chain of the smaller Gemma4 model may introduce additional paraphrastic freedom.} Ancestral sampling brings NLLB substantially closer to human levels (0.590, 0.583). This is the strongest within-model effect in Task 1 and corroborates \citeauthor{luo-etal-2024-diverge}'s \citeyear{luo-etal-2024-diverge} finding on beam search literality bias. 

When prompted to iteratively improve their own output (Task 2), all evaluated LLMs exhibit a strict, monotonic decrease in SLI across successive translation attempts (\#1 through \#5). This provides compelling, novel evidence that the deliteralization hypothesis applies natively to LLM generation: given computational reflection, LLMs systematically shed source-aligned syntax and lexicon in favor of freer formulations. This drift is further accelerated when using non-greedy sampling ($T=1$). While all models deliteralize across positions, the dynamics differ sharply. Qwen-3 models show a steady monotonic decline ($\sim$0.617 → 0.572), staying above human throughout. OLMo-3 models show a rapid initial drop followed by a plateau, suggesting the revision process saturates quickly. Gemma-4 models exhibit the most dramatic trajectory — starting slightly above human levels at position 1 and falling well below human by position 5 (0.525–0.535), the only systems to do so. FR-only SLI values are uniformly higher (0.63–0.65), consistent with English–French being a closer typological pair, but the relative ordering of models and the declining position trend replicate faithfully, lending cross-lingual validity to the main results.

However, Task 3 reveals profound behavioral divergences when LLMs act as post-editors of human drafts. As detailed in Appendix~\ref{app:pe_dynamics}, while human post-editors selectively alter 52.2\% of initial segments, LLMs exhibit indiscriminate hyper-activity, overwriting 87\% to 99\% of human drafts across all text domains. LLM post-editors appear incapable of conservative editing, with the Gemma-4 models leaving 6–8\% unchanged versus 1–3\% for Qwen3 — still far more interventionist than the human post-editor, but reflecting their closer-to-human SLI baseline from Task 1 and better performance from the point of view of MetricX-24 scores.

\begin{table}[t]
\centering
\resizebox{0.75\columnwidth}{!}{%
\small\setlength{\tabcolsep}{5pt}
\begin{tabular}{lrrrrrr}
\toprule
\textbf{System} & $n$ & \textbf{PB-}$r$ & $p$ & $\rho$ & $p$ \\
\midrule
Human & 44,525 & $+0.0300$ & $<$0.001 & $+0.0279$ & $<$0.001 \\
Qwen3-8B$^{-}$ & 40,971 & $-0.1048$ & $<$0.001 & $-0.0944$ & $<$0.001 \\
Qwen3-8B$^{+}$ & 39,955 & $-0.1050$ & $<$0.001 & $-0.0943$ & $<$0.001 \\
Qwen3-32B$^{-}$ & 41,330 & $-0.1012$ & $<$0.001 & $-0.0938$ & $<$0.001 \\
Qwen3-32B$^{+}$ & 41,665 & $-0.0709$ & $<$0.001 & $-0.0657$ & $<$0.001 \\
OLMo-32B$^{+}$ & 31,164 & $-0.0894$ & $<$0.001 & $-0.0946$ & $<$0.001 \\
OLMo-32B-SFT$^{+}$ & 27,996 & $-0.0898$ & $<$0.001 & $-0.0821$ & $<$0.001 \\
Gemma4-31B$^{+}$ & 44,792 & $-0.0958$ & $<$0.001 & $-0.0903$ & $<$0.001 \\
Gemma4-31B$^{-}$ & 45,280 & $-0.0972$ & $<$0.001 & $-0.0925$ & $<$0.001 \\
Gemma4-E4B$^{+}$ & 44,273 & $-0.0789$ & $<$0.001 & $-0.0718$ & $<$0.001 \\
Gemma4-E4B$^{-}$ & 43,604 & $-0.0881$ & $<$0.001 & $-0.0829$ & $<$0.001 \\
\bottomrule
\end{tabular}
}
\caption{Correlation between SLI of original translation and likelihood of post-editor alteration. PB-$r$ = point-biserial correlation \cite{Lev1949}; $\rho$ = Spearman rank correlation \cite{Spearman1904}. Positive = more literal originals more likely to be altered.}
\label{tab:pe_corr}
\end{table}

More crucially, Table \ref{tab:pe_corr} exposes a fundamental misalignment in revision triggers. For human post-editors, higher literality in the draft correlates with a significantly higher likelihood of revision (PB-$r = +0.0300$, $p<0.001$), confirming that humans actively police and correct literality in drafts. MT post-editors exhibit the exact opposite behavior, resulting in significant \textit{negative} correlations across all tested models (e.g., $-0.1048$ for Qwen3-8B$^{-}$). As post-editors, LLMs systematically tolerate overly literal drafts while aggressively targeting freer formulations for revision. Paradoxically, once an LLM intervenes, the resulting edit is predominantly deliteralizing (Appendix~\ref{app:pe_dynamics}). For all MT systems except OLMo,\footnote{OLMo-32B is the only system that reliteralizes more than it deliteralizes (49.0\% vs. 43.4\%), with OLMo-SFT close to parity (41.9\% vs. 49.9\%).} deliteralization consistently exceeds reliteralization across all four domains. The human post-editor likewise deliteralizes more than they reliteralize (44.2\% vs. 29.1\%), but with a much higher neutral rate (26.7\%) — roughly a quarter of human edits change the text without shifting its literality beyond the interval threshold used to demarcate the neutral band ($\epsilon = 0.005$; see Appendix~\ref{app:pe_dynamics}), reflecting stylistic, grammatical, or factual corrections invisible to the SLI at this scale. LLMs thus mimic the structural direction of human revisions while failing entirely to replicate their priorities. Interestingly, the highest reliteralizing dynamic for the human post-editors occurs in the literary domain, corroborating past case studies \cite{borg2022literary}.

\section{Conclusion}
We tested the deliteralization hypothesis across 54 language pairs, eight model variants, and three translation tasks, using a validated Synthetic Literality Index alongside MetricX-24 quality scores. Three findings emerge. First, in direct translation, human references remain significantly less literal than NMT systems and most LLMs; only the Gemma-4 family approaches human levels of freedom, and ancestral sampling brings NLLB closer to human literality at substantial quality cost.
Second, when prompted to iteratively revise their own output, every tested LLM exhibits a strict monotonic decrease in literality across successive translations — the first direct evidence, to our knowledge, that the deliteralization hypothesis applies natively to LLM generation.
Third, and most strikingly, post-editing reveals an inversion of revision triggers: human post-editors are significantly more likely to revise highly literal drafts (PB-r = +0.030, p < 0.001), whereas every tested LLM shows the opposite pattern, tolerating literal drafts and targeting idiomatic human formulations for revision. Paradoxically, once an LLM intervenes, the resulting edit is predominantly deliteralizing — mimicking the direction of human revisions while inverting their priorities. LLMs deliteralize when prompted to revise, but they do so indiscriminately, missing the diagnostic cues that guide human post-editors.

\section*{Limitations}

We acknowledge several limitations of this work, while noting that the scale of our experiments---3.1M machine translations and 103,680 human translations across 54 typologically diverse target languages---makes the central trends we report robust to many of the specific design choices discussed below.

\paragraph{English-source-only design.} All 54 language pairs share English as the source, a constraint inherited from WMT24++. Source-language effects on literality are well-documented, and our findings are, strictly speaking, claims about translation out of English. We note, however, that the typological diversity of our target languages---spanning Indo-European, Sino-Tibetan, Afroasiatic, Dravidian, Uralic, Turkic, Austroasiatic, Austronesian, Niger-Congo, Japonic, and Koreanic families---provides substantial coverage of the cross-linguistic conditions under which the deliteralization hypothesis is meant to hold. Replication with non-English source material remains valuable direction for future work.

\paragraph{The ``initial'' human translation is not a first draft.} WMT24++ provides ``initial'' human translations and their post-edited counterparts, but the initial translations are themselves the product of the work of professional translators who have already performed internal revision through interim translations. Our human two-state data therefore captures only the final revision step of the initial drafting phase, in contrast to the cognitive drafting process studied via think-aloud protocols and keystroke logging in the classical translation studies literature. We see this as cutting both ways: it likely understates the deliteralization that occurred during human drafting, but it also makes the residual deliteralization we do observe in human post-editing all the more striking, since it persists even at this late revision stage. We further note that the deliteralization hypothesis can legitimately be taken to apply across the translation–revision continuum rather than being confined to a single agent's drafting process.

\paragraph{Reference-dependent quality filtering.} Our MetricX-24 quality filter is computed against the post-edited human reference, which advantages systems whose output resembles that reference and may interact with literality in non-trivial ways. We adopted a strict MX $\leq$ 5 threshold, after manual auditing, precisely because we judged that, for an analysis of literality, precision should be prioritized over recall: including low-quality translations would conflate genuine signals of ``non-literality'' with hallucination and degenerate output. The volume of surviving segments at this threshold (over 730K LLM translations for Task 2 alone) leaves ample statistical power, and we verified that the trends we report replicate both at the more permissive MX $\leq$ 7 and stricter MX $\leq$ 3 thresholds.

\paragraph{Heuristic coverage gaps.} Nine target languages (Bengali, Filipino, Gujarati, Kannada, Malayalam, Punjabi, Swahili, Thai, Zulu) lack reliable morphosyntactic parsers, so that \textsc{pos\_sim} and \textsc{tree\_sim} are unavailable for the LPs involving these languages and their SLI softmax is renormalized over the remaining four signals. The alignment-based heuristics (\textsc{density}, \textsc{tok\_sim\_raw}, \textsc{tok\_sim\_pen}) are likewise affected to a lesser degree, as their underlying word-level alignments are computed over the output of the Stanza word tokenizers where available and default to whitespace tokenization otherwise. For Thai, tokenization is handled by Stanza's neural word segmenter; for the eight remaining languages, all of which use whitespace as a word delimiter, the alignment-based heuristics (\textsc{density}, \textsc{tok\_sim\_raw}, \textsc{tok\_sim\_pen}) fall back to whitespace tokenization, which may understate alignment density for morphologically rich languages such as Zulu. While these are principled responses to a real-world resource gap, they mean cross-LP comparisons involving the affected languages rest on a narrower signal base. Reassuringly, \textsc{seg\_sem\_labse}---the heuristic with the highest validation hit rate and the strongest correlation with human judgements---operates on full-segment LaBSE embeddings independently of any tokenizer, providing a stable signal for all 54 LPs, further amplified by the softmax temperature.

\section{Ethics Statement}

Our experiments required a diverse set of models and language pairs to gather robust insights. We acknowledge the substantial ecological cost of large-scale MT inference and evaluation. Our work consumed allocations on the Jean Zay HPC cluster, which operates on largely decarbonized electricity, and we release our compiled translation outputs under CC-BY-SA license for re-use by the community.

All models used in this study (Qwen-3, OLMo-3, Gemma-4, NLLB-200) were selected in part for their open licenses, which permit the release of derived data and analyses. We release our compiled translation outputs and code under CC-BY-SA and MIT licenses, respectively, in keeping with open-science practice and to facilitate replication and extension by the community.

The WMT24++ dataset on which we build was released by \citet{deutsch-etal-2025-wmt24} under terms permitting research use; we use it within those terms and add no personally identifying information. Our human validation campaign was conducted by the authors and consenting colleagues on a voluntary basis, with no compensation involved and no personal data collected beyond pairwise literality judgments.

Finally, we note that our findings concerning systematic literality biases in current MT systems have implications for the deployment of these systems in high-stakes translation settings — including legal, medical, and literary contexts — where stylistic register and idiomatic naturalness materially affect the reception and usability of translated text. We caution against deploying MT post-editing pipelines without human oversight, particularly in light of our finding that LLM post-editors invert the revision priorities of human professionals.

\section*{Acknowledgments}

This work was funded by the French \textit{Agence Nationale de la Recherche} (ANR) under the project TraLaLaM (``ANR-23-IAS1-0006''), as well as by Inria under the ``\textit{Défi}''-type project COLaF.  It was also partly funded by Rachel Bawden and Benoît Sagot’s chairs in the PRAIRIE institute, funded by the French national agency ANR, as part of the “Investissements d’avenir” programme under the reference ANR-19-P3IA-0001 and by Benoît Sagot's chair in its follow-up, PRAIRIE-PSAI, also funded by the ANR as part of the “France 2030” strategy under the reference ANR23-IACL-0008.

This work was also granted access to the HPC resources of IDRIS under allocations 2025-AD011015117R2, 2025-AD011012254R5 and 2025-AD011017228 made by GENCI. The authors acknowledge the use of LLMs as coding assistants during the implementation of this work.


\bibliography{custom}

@article{ivir-formal,
 ISSN = {03335372, 15275507},
 URL = {http://www.jstor.org/stable/1772485},
 author = {Vladimir Ivir},
 journal = {Poetics Today},
 number = {4},
 pages = {51--59},
 publisher = {[Duke University Press, Porter Institute for Poetics and Semiotics]},
 title = {Formal Correspondence vs. Translation Equivalence Revisited},
 urldate = {2026-04-09},
 volume = {2},
 year = {1981}
}

@article{TirkkonenCondit2005TheMM,
  title={The Monitor Model Revisited: Evidence from Process Research},
  author={Sonja Tirkkonen-Condit},
  journal={Meta: Translators' Journal},
  year={2005},
  volume={50},
  pages={405-414},
  url={https://api.semanticscholar.org/CorpusID:27965075}
}

@incollection{chestermanReflections,
   author = "Chesterman, Andrew",
   title = "Reflections on the literal translation hypothesis",
   booktitle = "Methods and Strategies of Process Research",
   publisher = "John Benjamins",
   year = "2011",
   pages = "23-35",
   url = "https://www.jbe-platform.com/content/books/9789027285195-btl.94.05che"
}

@book{englundDimitrovaExpertise,
   author = "Englund Dimitrova, Birgitta",
   title = "Expertise and Explicitation in the Translation Process",
   publisher = "John Benjamins",
   year = "2005",
   url = "https://www.jbe-platform.com/content/books/9789027294265",
}

@inbook{PavlovicAntunDesc,
url = {https://doi.org/10.1075/bct.77.06pav},
title = {The effect of interpreting experience on distance dynamics: Testing the literal translation hypothesis},
booktitle = {Describing Cognitive Processes in Translation},
booktitle = {Acts and events},
author = {Nataša Pavlović and Goranka Antunovic},
editor = {Maureen Ehrensberger-Dow and Birgitta Englund Dimitrova and Séverine Hubscher-Davidson and Ulf Norberg},
publisher = {John Benjamins Publishing Company},
pages = {85--103},
doi = {doi:10.1075/bct.77.06pav},
isbn = {9789027268204},
year = {2015},
lastchecked = {2026-04-09}
}

@book{borg2022literary,
  author = "Borg, Claudine",
  title = "A Literary Translation in the Making: A Process-Oriented Perspective",
  edition = "1st",
  publisher = "Routledge",
  year = "2022",
  doi = "10.4324/9781003150909"
}

@inproceedings{zhai-etal-2020-detecting,
    title = "Detecting Non-literal Translations by Fine-tuning Cross-lingual Pre-trained Language Models",
    author = "Zhai, Yuming  and
      Illouz, Gabriel  and
      Vilnat, Anne",
    editor = "Scott, Donia  and
      Bel, Nuria  and
      Zong, Chengqing",
    booktitle = "Proceedings of the 28th International Conference on Computational Linguistics",
    month = dec,
    year = "2020",
    address = "Barcelona, Spain (Online)",
    publisher = "International Committee on Computational Linguistics",
    url = "https://aclanthology.org/2020.coling-main.522/",
    doi = "10.18653/v1/2020.coling-main.522",
    pages = "5944--5956"
}

@inproceedings{zhai-etal-2018-construction,
    title = "Construction of a Multilingual Corpus Annotated with Translation Relations",
    author = "Zhai, Yuming  and
      Max, Aur{\'e}lien  and
      Vilnat, Anne",
    editor = "Machonis, Peter  and
      Barreiro, Anabela  and
      Kocijan, Kristina  and
      Silberztein, Max",
    booktitle = "Proceedings of the First Workshop on Linguistic Resources for Natural Language Processing",
    month = aug,
    year = "2018",
    address = "Santa Fe, New Mexico, USA",
    publisher = "Association for Computational Linguistics",
    url = "https://aclanthology.org/W18-3814/",
    pages = "102--111"
}

@inproceedings{zhai-etal-2019-classification,
    title = "Classification automatique des proc{\'e}d{\'e}s de traduction (Automatic Classification of Translation Processes)",
    author = "Zhai, Yuming  and
      Illouz, Gabriel  and
      Vilnat, Anne",
    editor = "Morin, Emmanuel  and
      Rosset, Sophie  and
      Zweigenbaum, Pierre",
    booktitle = "Actes de la Conf{\'e}rence sur le Traitement Automatique des Langues Naturelles (TALN) PFIA 2019. Volume II : Articles courts",
    month = "7",
    year = "2019",
    address = "Toulouse, France",
    publisher = "ATALA",
    url = "https://aclanthology.org/2019.jeptalnrecital-court.6/",
    pages = "205--214",
    language = "fra"
}

@inproceedings{dankers-etal-2022-transformer,
    title = "Can Transformer be Too Compositional? Analysing Idiom Processing in Neural Machine Translation",
    author = "Dankers, Verna  and
      Lucas, Christopher  and
      Titov, Ivan",
    editor = "Muresan, Smaranda  and
      Nakov, Preslav  and
      Villavicencio, Aline",
    booktitle = "Proceedings of the 60th Annual Meeting of the Association for Computational Linguistics (Volume 1: Long Papers)",
    month = may,
    year = "2022",
    address = "Dublin, Ireland",
    publisher = "Association for Computational Linguistics",
    url = "https://aclanthology.org/2022.acl-long.252/",
    doi = "10.18653/v1/2022.acl-long.252",
    pages = "3608--3626"
}

@article{luo-etal-2024-diverge,
    title = "To Diverge or Not to Diverge: A Morphosyntactic Perspective on Machine Translation vs Human Translation",
    author = "Luo, Jiaming  and
      Cherry, Colin  and
      Foster, George",
    journal = "Transactions of the Association for Computational Linguistics",
    volume = "12",
    year = "2024",
    address = "Cambridge, MA",
    publisher = "MIT Press",
    url = "https://aclanthology.org/2024.tacl-1.20/",
    doi = "10.1162/tacl_a_00645",
    pages = "355--371"
}

@inproceedings{raunak-etal-2023-gpts,
    title = "Do {GPT}s Produce Less Literal Translations?",
    author = "Raunak, Vikas  and
      Menezes, Arul  and
      Post, Matt  and
      Hassan, Hany",
    editor = "Rogers, Anna  and
      Boyd-Graber, Jordan  and
      Okazaki, Naoaki",
    booktitle = "Proceedings of the 61st Annual Meeting of the Association for Computational Linguistics (Volume 2: Short Papers)",
    month = jul,
    year = "2023",
    address = "Toronto, Canada",
    publisher = "Association for Computational Linguistics",
    url = "https://aclanthology.org/2023.acl-short.90/",
    doi = "10.18653/v1/2023.acl-short.90",
    pages = "1041--1050"
}

@inproceedings{li-etal-2025-lost,
    title = "Lost in Literalism: How Supervised Training Shapes Translationese in {LLM}s",
    author = "Li, Yafu  and
      Zhang, Ronghao  and
      Wang, Zhilin  and
      Zhang, Huajian  and
      Cui, Leyang  and
      Yin, Yongjing  and
      Xiao, Tong  and
      Zhang, Yue",
    editor = "Che, Wanxiang  and
      Nabende, Joyce  and
      Shutova, Ekaterina  and
      Pilehvar, Mohammad Taher",
    booktitle = "Proceedings of the 63rd Annual Meeting of the Association for Computational Linguistics (Volume 1: Long Papers)",
    month = jul,
    year = "2025",
    address = "Vienna, Austria",
    publisher = "Association for Computational Linguistics",
    url = "https://aclanthology.org/2025.acl-long.630/",
    doi = "10.18653/v1/2025.acl-long.630",
    pages = "12875--12894",
    ISBN = "979-8-89176-251-0"
}

@book{shuttleworth1997dictionary,
  author = "Shuttleworth, Mark and Cowie, Moira",
  title = "Dictionary of Translation Studies",
  edition = "1st",
  publisher = "Routledge",
  year = "1997",
  doi = "10.4324/9781315760490"
}

@incollection{jerome1893pammachius,
  author    = {{Jerome [395 CE]}},
  title     = {Letter {LVII} to {Pammachius}: On the Best Method of Translating},
  translator= {W. H. Fremantle},
  editor    = {Philip Schaff and Henry Wace},
  booktitle = {A Select Library of Nicene and Post-Nicene Fathers of the Christian Church},
  series    = {Second Series},
  volume    = {6},
  publisher = {Christian Literature Company},
  address   = {New York},
  year      = {1893}
}

@incollection{benjamin1968task,
  author    = {{Benjamin [1923]}},
  title     = {The Task of the Translator},
  translator= {Harry Zohn},
  editor    = {Hannah Arendt},
  booktitle = {Illuminations},
  publisher = {Harcourt, Brace and World},
  address   = {New York},
  year      = {1968}
}

@incollection{schleiermacher2012methods,
  author    = {{Schleiermacher [1813]}},
  title     = {On the Different Methods of Translating},
  translator= {Susan Bernofsky},
  editor    = {Lawrence Venuti},
  booktitle = {The Translation Studies Reader},
  edition   = {3rd},
  publisher = {Routledge},
  address   = {London},
  year      = {2012}
}

@incollection{dryden1681preface,
  author    = {Dryden, John},
  title     = {Preface to {Ovid's} Epistles},
  booktitle = {Ovid's Epistles, Translated by Several Hands},
  publisher = {Jacob Tonson},
  address   = {London},
  year      = {1681}
}

@book{venuti1995invisibility,
  author    = {Venuti, Lawrence},
  title     = {The Translator's Invisibility: A History of Translation},
  publisher = {Routledge},
  address   = {London},
  year      = {1995},
  doi       = {10.4324/9780203360064}
}

@inproceedings{deutsch-etal-2025-wmt24,
    title = "{WMT}24++: Expanding the Language Coverage of {WMT}24 to 55 Languages {\&} Dialects",
    author = "Deutsch, Daniel  and
      Briakou, Eleftheria  and
      Caswell, Isaac Rayburn  and
      Finkelstein, Mara  and
      Galor, Rebecca  and
      Juraska, Juraj  and
      Kovacs, Geza  and
      Lui, Alison  and
      Rei, Ricardo  and
      Riesa, Jason  and
      Rijhwani, Shruti  and
      Riley, Parker  and
      Salesky, Elizabeth  and
      Trabelsi, Firas  and
      Winkler, Stephanie  and
      Zhang, Biao  and
      Freitag, Markus",
    editor = "Che, Wanxiang  and
      Nabende, Joyce  and
      Shutova, Ekaterina  and
      Pilehvar, Mohammad Taher",
    booktitle = "Findings of the Association for Computational Linguistics: ACL 2025",
    month = jul,
    year = "2025",
    address = "Vienna, Austria",
    publisher = "Association for Computational Linguistics",
    url = "https://aclanthology.org/2025.findings-acl.634/",
    doi = "10.18653/v1/2025.findings-acl.634",
    pages = "12257--12284",
    ISBN = "979-8-89176-256-5"
}

@inproceedings{kocmi-etal-2024-findings,
    title = "Findings of the {WMT}24 General Machine Translation Shared Task: The {LLM} Era Is Here but {MT} Is Not Solved Yet",
    author = "Kocmi, Tom  and
      Avramidis, Eleftherios  and
      Bawden, Rachel  and
      Bojar, Ond{\v{r}}ej  and
      Dvorkovich, Anton  and
      Federmann, Christian  and
      Fishel, Mark  and
      Freitag, Markus  and
      Gowda, Thamme  and
      Grundkiewicz, Roman  and
      Haddow, Barry  and
      Karpinska, Marzena  and
      Koehn, Philipp  and
      Marie, Benjamin  and
      Monz, Christof  and
      Murray, Kenton  and
      Nagata, Masaaki  and
      Popel, Martin  and
      Popovi{\'c}, Maja  and
      Shmatova, Mariya  and
      Steingr{\'i}msson, Steinth{\'o}r  and
      Zouhar, Vil{\'e}m",
    editor = "Haddow, Barry  and
      Kocmi, Tom  and
      Koehn, Philipp  and
      Monz, Christof",
    booktitle = "Proceedings of the Ninth Conference on Machine Translation",
    month = nov,
    year = "2024",
    address = "Miami, Florida, USA",
    publisher = "Association for Computational Linguistics",
    url = "https://aclanthology.org/2024.wmt-1.1/",
    doi = "10.18653/v1/2024.wmt-1.1",
    pages = "1--46"
}

@misc{nllbteam2022languageleftbehindscaling,
      title={{No Language Left Behind: Scaling Human-Centered Machine Translation}}, 
      author={{NLLB Team} and Marta R. Costa-jussà and James Cross and Onur Çelebi and Maha Elbayad and Kenneth Heafield and Kevin Heffernan and Elahe Kalbassi and Janice Lam and Daniel Licht and Jean Maillard and Anna Sun and Skyler Wang and Guillaume Wenzek and Al Youngblood and Bapi Akula and Loic Barrault and Gabriel Mejia Gonzalez and Prangthip Hansanti and John Hoffman and Semarley Jarrett and Kaushik Ram Sadagopan and Dirk Rowe and Shannon Spruit and Chau Tran and Pierre Andrews and Necip Fazil Ayan and Shruti Bhosale and Sergey Edunov and Angela Fan and Cynthia Gao and Vedanuj Goswami and Francisco Guzmán and Philipp Koehn and Alexandre Mourachko and Christophe Ropers and Safiyyah Saleem and Holger Schwenk and Jeff Wang},
      year={2022},
      eprint={2207.04672},
      archivePrefix={arXiv},
      primaryClass={cs.CL},
      url={https://arxiv.org/abs/2207.04672}, 
}

@misc{yang2025qwen3technicalreport,
      title={Qwen3 Technical Report}, 
      author={An Yang and Anfeng Li and Baosong Yang and Beichen Zhang and Binyuan Hui and Bo Zheng and Bowen Yu and Chang Gao and Chengen Huang and Chenxu Lv and Chujie Zheng and Dayiheng Liu and Fan Zhou and Fei Huang and Feng Hu and Hao Ge and Haoran Wei and Huan Lin and Jialong Tang and Jian Yang and Jianhong Tu and Jianwei Zhang and Jianxin Yang and Jiaxi Yang and Jing Zhou and Jingren Zhou and Junyang Lin and Kai Dang and Keqin Bao and Kexin Yang and Le Yu and Lianghao Deng and Mei Li and Mingfeng Xue and Mingze Li and Pei Zhang and Peng Wang and Qin Zhu and Rui Men and Ruize Gao and Shixuan Liu and Shuang Luo and Tianhao Li and Tianyi Tang and Wenbiao Yin and Xingzhang Ren and Xinyu Wang and Xinyu Zhang and Xuancheng Ren and Yang Fan and Yang Su and Yichang Zhang and Yinger Zhang and Yu Wan and Yuqiong Liu and Zekun Wang and Zeyu Cui and Zhenru Zhang and Zhipeng Zhou and Zihan Qiu},
      year={2025},
      eprint={2505.09388},
      archivePrefix={arXiv},
      primaryClass={cs.CL},
      url={https://arxiv.org/abs/2505.09388}, 
}

@misc{olmo2025olmo3,
      title={Olmo 3}, 
      author={Team Olmo and Allyson Ettinger and Amanda Bertsch and Bailey Kuehl and David Graham and David Heineman and Dirk Groeneveld and Faeze Brahman and Finbarr Timbers and Hamish Ivison and Jacob Morrison and Jake Poznanski and Kyle Lo and Luca Soldaini and Matt Jordan and Mayee Chen and Michael Noukhovitch and Nathan Lambert and Pete Walsh and Pradeep Dasigi and Robert Berry and Saumya Malik and Saurabh Shah and Scott Geng and Shane Arora and Shashank Gupta and Taira Anderson and Teng Xiao and Tyler Murray and Tyler Romero and Victoria Graf and Akari Asai and Akshita Bhagia and Alexander Wettig and Alisa Liu and Aman Rangapur and Chloe Anastasiades and Costa Huang and Dustin Schwenk and Harsh Trivedi and Ian Magnusson and Jaron Lochner and Jiacheng Liu and Lester James V. Miranda and Maarten Sap and Malia Morgan and Michael Schmitz and Michal Guerquin and Michael Wilson and Regan Huff and Ronan Le Bras and Rui Xin and Rulin Shao and Sam Skjonsberg and Shannon Zejiang Shen and Shuyue Stella Li and Tucker Wilde and Valentina Pyatkin and Will Merrill and Yapei Chang and Yuling Gu and Zhiyuan Zeng and Ashish Sabharwal and Luke Zettlemoyer and Pang Wei Koh and Ali Farhadi and Noah A. Smith and Hannaneh Hajishirzi},
      year={2025},
      eprint={2512.13961},
      archivePrefix={arXiv},
      primaryClass={cs.CL},
      url={https://arxiv.org/abs/2512.13961}, 
}

@inproceedings{kwon2023efficient,
  title={Efficient Memory Management for Large Language Model Serving with PagedAttention},
  author={Woosuk Kwon and Zhuohan Li and Siyuan Zhuang and Ying Sheng and Lianmin Zheng and Cody Hao Yu and Joseph E. Gonzalez and Hao Zhang and Ion Stoica},
  booktitle={Proceedings of the ACM SIGOPS 29th Symposium on Operating Systems Principles},
  year={2023}
}

@inproceedings{juraska-etal-2024-metricx,
    title = "{M}etric{X}-24: The {G}oogle Submission to the {WMT} 2024 Metrics Shared Task",
    author = "Juraska, Juraj  and
      Deutsch, Daniel  and
      Finkelstein, Mara  and
      Freitag, Markus",
    editor = "Haddow, Barry  and
      Kocmi, Tom  and
      Koehn, Philipp  and
      Monz, Christof",
    booktitle = "Proceedings of the Ninth Conference on Machine Translation",
    month = nov,
    year = "2024",
    address = "Miami, Florida, USA",
    publisher = "Association for Computational Linguistics",
    url = "https://aclanthology.org/2024.wmt-1.35/",
    doi = "10.18653/v1/2024.wmt-1.35",
    pages = "492--504"
}

@inproceedings{wolf-etal-2020-transformers,
    title = "Transformers: State-of-the-Art Natural Language Processing",
    author = "Thomas Wolf and Lysandre Debut and Victor Sanh and Julien Chaumond and Clement Delangue and Anthony Moi and Pierric Cistac and Tim Rault and Rémi Louf and Morgan Funtowicz and Joe Davison and Sam Shleifer and Patrick von Platen and Clara Ma and Yacine Jernite and Julien Plu and Canwen Xu and Teven Le Scao and Sylvain Gugger and Mariama Drame and Quentin Lhoest and Alexander M. Rush",
    booktitle = "Proceedings of the 2020 Conference on Empirical Methods in Natural Language Processing: System Demonstrations",
    month = oct,
    year = "2020",
    address = "Online",
    publisher = "Association for Computational Linguistics",
    url = "https://www.aclweb.org/anthology/2020.emnlp-demos.6",
    pages = "38--45"
}

@inproceedings{qi-etal-2020-stanza,
    title = "{S}tanza: A Python Natural Language Processing Toolkit for Many Human Languages",
    author = "Qi, Peng  and
      Zhang, Yuhao  and
      Zhang, Yuhui  and
      Bolton, Jason  and
      Manning, Christopher D.",
    editor = "Celikyilmaz, Asli  and
      Wen, Tsung-Hsien",
    booktitle = "Proceedings of the 58th Annual Meeting of the Association for Computational Linguistics: System Demonstrations",
    month = jul,
    year = "2020",
    address = "Online",
    publisher = "Association for Computational Linguistics",
    url = "https://aclanthology.org/2020.acl-demos.14/",
    doi = "10.18653/v1/2020.acl-demos.14",
    pages = "101--108"
}

@inproceedings{jalili-sabet-etal-2020-simalign,
    title = "{S}im{A}lign: High Quality Word Alignments Without Parallel Training Data Using Static and Contextualized Embeddings",
    author = {Jalili Sabet, Masoud  and
      Dufter, Philipp  and
      Yvon, Fran{\c{c}}ois  and
      Sch{\"u}tze, Hinrich},
    editor = "Cohn, Trevor  and
      He, Yulan  and
      Liu, Yang",
    booktitle = "Findings of the Association for Computational Linguistics: EMNLP 2020",
    month = nov,
    year = "2020",
    address = "Online",
    publisher = "Association for Computational Linguistics",
    url = "https://aclanthology.org/2020.findings-emnlp.147/",
    doi = "10.18653/v1/2020.findings-emnlp.147",
    pages = "1627--1643"
}

@inproceedings{feng-etal-2022-language,
    title = "Language-agnostic {BERT} Sentence Embedding",
    author = "Feng, Fangxiaoyu  and
      Yang, Yinfei  and
      Cer, Daniel  and
      Arivazhagan, Naveen  and
      Wang, Wei",
    editor = "Muresan, Smaranda  and
      Nakov, Preslav  and
      Villavicencio, Aline",
    booktitle = "Proceedings of the 60th Annual Meeting of the Association for Computational Linguistics (Volume 1: Long Papers)",
    month = may,
    year = "2022",
    address = "Dublin, Ireland",
    publisher = "Association for Computational Linguistics",
    url = "https://aclanthology.org/2022.acl-long.62/",
    doi = "10.18653/v1/2022.acl-long.62",
    pages = "878--891"
}

@manual{python_difflib,
  title  = {difflib --- {H}elpers for computing deltas},
  author = {{Python Software Foundation}},
  year   = {2024},
  note   = {Python Standard Library documentation},
  url    = {https://docs.python.org/3/library/difflib.html}
}

@inproceedings{conneau-etal-2020-unsupervised,
    title = "Unsupervised Cross-lingual Representation Learning at Scale",
    author = "Conneau, Alexis  and
      Khandelwal, Kartikay  and
      Goyal, Naman  and
      Chaudhary, Vishrav  and
      Wenzek, Guillaume  and
      Guzm{\'a}n, Francisco  and
      Grave, Edouard  and
      Ott, Myle  and
      Zettlemoyer, Luke  and
      Stoyanov, Veselin",
    editor = "Jurafsky, Dan  and
      Chai, Joyce  and
      Schluter, Natalie  and
      Tetreault, Joel",
    booktitle = "Proceedings of the 58th Annual Meeting of the Association for Computational Linguistics",
    month = jul,
    year = "2020",
    address = "Online",
    publisher = "Association for Computational Linguistics",
    url = "https://aclanthology.org/2020.acl-main.747/",
    doi = "10.18653/v1/2020.acl-main.747",
    pages = "8440--8451"
}

@incollection{Gellerstam1986,
  author    = {Gellerstam, Martin},
  title     = {Translationese in {S}wedish novels translated from {E}nglish},
  booktitle = {Translation studies in Scandinavia: Proceedings from the Scandinavian Symposium on Translation Theory (SSOTT) II},
  editor    = {Wollin, L. and Lindquist, H.},
  year      = {1986},
  series    = {Lund Studies in English},
  number    = {75},
  pages     = {88--95},
  publisher = {CWK Gleerup},
  address   = {Lund}
}

@book{NidaTaber1969,
  author    = {Nida, Eugene A. and Taber, Charles R.},
  title     = {The Theory and Practice of Translation},
  year      = {1969},
  publisher = {E.J. Brill},
  address   = {Leiden},
  series    = {Helps for Translators},
  volume    = {8}
}

@inproceedings{dutta-chowdhury-etal-2022-towards,
    title = "Towards Debiasing Translation Artifacts",
    author = "Dutta Chowdhury, Koel  and
      Jalota, Rricha  and
      Espa{\~n}a-Bonet, Cristina  and
      Genabith, Josef",
    editor = "Carpuat, Marine  and
      de Marneffe, Marie-Catherine  and
      Meza Ruiz, Ivan Vladimir",
    booktitle = "Proceedings of the 2022 Conference of the North American Chapter of the Association for Computational Linguistics: Human Language Technologies",
    month = jul,
    year = "2022",
    address = "Seattle, United States",
    publisher = "Association for Computational Linguistics",
    url = "https://aclanthology.org/2022.naacl-main.292/",
    doi = "10.18653/v1/2022.naacl-main.292",
    pages = "3983--3991"
}

@inproceedings{riley-etal-2020-translationese,
    title = "Translationese as a Language in ``Multilingual'' {NMT}",
    author = "Riley, Parker  and
      Caswell, Isaac  and
      Freitag, Markus  and
      Grangier, David",
    editor = "Jurafsky, Dan  and
      Chai, Joyce  and
      Schluter, Natalie  and
      Tetreault, Joel",
    booktitle = "Proceedings of the 58th Annual Meeting of the Association for Computational Linguistics",
    month = jul,
    year = "2020",
    address = "Online",
    publisher = "Association for Computational Linguistics",
    url = "https://aclanthology.org/2020.acl-main.691/",
    doi = "10.18653/v1/2020.acl-main.691",
    pages = "7737--7746"
}

@inproceedings{kurokawa-etal-2009-automatic,
    title = "Automatic Detection of Translated Text and its Impact on Machine Translation",
    author = "Kurokawa, David  and
      Goutte, Cyril  and
      Isabelle, Pierre",
    booktitle = "Proceedings of Machine Translation Summit XII: Papers",
    month = aug # " 26-30",
    year = "2009",
    address = "Ottawa, Canada",
    url = "https://aclanthology.org/2009.mtsummit-papers.9/"
}

@inproceedings{koppelTranslationese,
author = {Koppel, Moshe and Ordan, Noam},
title = {Translationese and its dialects},
year = {2011},
isbn = {9781932432879},
publisher = {Association for Computational Linguistics},
address = {USA},
booktitle = {Proceedings of the 49th Annual Meeting of the Association for Computational Linguistics: Human Language Technologies - Volume 1},
pages = {1318–1326},
numpages = {9},
location = {Portland, Oregon},
series = {HLT '11}
}

@inproceedings{NIPS2017_3f5ee243,
 author = {Vaswani, Ashish and Shazeer, Noam and Parmar, Niki and Uszkoreit, Jakob and Jones, Llion and Gomez, Aidan N and Kaiser, \L ukasz and Polosukhin, Illia},
 booktitle = {Advances in Neural Information Processing Systems},
 editor = {I. Guyon and U. Von Luxburg and S. Bengio and H. Wallach and R. Fergus and S. Vishwanathan and R. Garnett},
 pages = {},
 publisher = {Curran Associates, Inc.},
 title = {Attention is All you Need},
 url = {https://proceedings.neurips.cc/paper_files/paper/2017/file/3f5ee243547dee91fbd053c1c4a845aa-Paper.pdf},
 volume = {30},
 year = {2017}
}

@inproceedings{zhang-toral-2019-effect,
    title = "The Effect of Translationese in Machine Translation Test Sets",
    author = "Zhang, Mike  and
      Toral, Antonio",
    editor = "Bojar, Ond{\v{r}}ej  and
      Chatterjee, Rajen  and
      Federmann, Christian  and
      Fishel, Mark  and
      Graham, Yvette  and
      Haddow, Barry  and
      Huck, Matthias  and
      Yepes, Antonio Jimeno  and
      Koehn, Philipp  and
      Martins, Andr{\'e}  and
      Monz, Christof  and
      Negri, Matteo  and
      N{\'e}v{\'e}ol, Aur{\'e}lie  and
      Neves, Mariana  and
      Post, Matt  and
      Turchi, Marco  and
      Verspoor, Karin",
    booktitle = "Proceedings of the Fourth Conference on Machine Translation (Volume 1: Research Papers)",
    month = aug,
    year = "2019",
    address = "Florence, Italy",
    publisher = "Association for Computational Linguistics",
    url = "https://aclanthology.org/W19-5208/",
    doi = "10.18653/v1/W19-5208",
    pages = "73--81"
}

@inproceedings{graham-etal-2020-statistical,
    title = "Statistical Power and Translationese in Machine Translation Evaluation",
    author = "Graham, Yvette  and
      Haddow, Barry  and
      Koehn, Philipp",
    editor = "Webber, Bonnie  and
      Cohn, Trevor  and
      He, Yulan  and
      Liu, Yang",
    booktitle = "Proceedings of the 2020 Conference on Empirical Methods in Natural Language Processing (EMNLP)",
    month = nov,
    year = "2020",
    address = "Online",
    publisher = "Association for Computational Linguistics",
    url = "https://aclanthology.org/2020.emnlp-main.6/",
    doi = "10.18653/v1/2020.emnlp-main.6",
    pages = "72--81"
}

@inproceedings{toral-2019-post,
    title = "Post-editese: an Exacerbated Translationese",
    author = "Toral, Antonio",
    editor = "Forcada, Mikel  and
      Way, Andy  and
      Haddow, Barry  and
      Sennrich, Rico",
    booktitle = "Proceedings of Machine Translation Summit XVII: Research Track",
    month = aug,
    year = "2019",
    address = "Dublin, Ireland",
    publisher = "European Association for Machine Translation",
    url = "https://aclanthology.org/W19-6627/",
    pages = "273--281"
}

@incollection{doCarmoMoorkens2020,
  author    = {do Carmo, F\'{e}lix and Moorkens, Joss},
  title     = {Differentiating editing, post-editing, and revision},
  booktitle = {Translation Revision and Post-editing: Industry Practices and Cognitive Processes},
  editor    = {Koponen, Maarit and Mossop, Brian and Robert, Isabelle and Scocchera, Giovanna},
  publisher = {Routledge (Taylor \& Francis)},
  address   = {Abingdon, UK},
  year      = {2020},
  pages     = {35--49},
  isbn      = {9781138549715}
}

@article{volanskyFeatures2015,
    author = {Volansky, Vered and Ordan, Noam and Wintner, Shuly},
    title = {On the features of translationese},
    journal = {Digital Scholarship in the Humanities},
    volume = {30},
    number = {1},
    pages = {98-118},
    year = {2015},
    month = {04},
    issn = {2055-7671},
    doi = {10.1093/llc/fqt031},
    url = {https://doi.org/10.1093/llc/fqt031},
    eprint = {https://academic.oup.com/dsh/article-pdf/30/1/98/21521905/fqt031.pdf},
}

@misc{gemma4modelcard2026,
  author = {{Google DeepMind}},
  title = {Gemma 4 Model Card},
  year = {2026},
  howpublished = {\url{https://ai.google.dev/gemma/docs/core/model_card_4}}
}

@article{friedman_test,
 ISSN = {01621459, 1537274X},
 URL = {http://www.jstor.org/stable/2279372},
 author = {Milton Friedman},
 journal = {Journal of the American Statistical Association},
 number = {200},
 pages = {675--701},
 publisher = {[American Statistical Association, Taylor & Francis, Ltd.]},
 title = {The Use of Ranks to Avoid the Assumption of Normality Implicit in the Analysis of Variance},
 urldate = {2026-05-03},
 volume = {32},
 year = {1937}
}

@article{baroninewapp,
author = {Baroni, Marco and Bernardini, Silvia},
year = {2005},
month = {08},
pages = {},
title = {A New Approach to the Study of Translationese: Machine-learning the Difference between Original and Translated Text},
volume = {21},
journal = {Literary and Linguistic Computing},
doi = {10.1093/llc/fqi039}
}

@inproceedings{van-halteren-2008-source,
    title = "Source Language Markers in {EUROPARL} Translations",
    author = "van Halteren, Hans",
    editor = "Scott, Donia  and
      Uszkoreit, Hans",
    booktitle = "Proceedings of the 22nd International Conference on Computational Linguistics (Coling 2008)",
    month = aug,
    year = "2008",
    address = "Manchester, UK",
    publisher = "Coling 2008 Organizing Committee",
    url = "https://aclanthology.org/C08-1118/",
    pages = "937--944"
}

@article{Cohen1960ACO,
  title={A Coefficient of Agreement for Nominal Scales},
  author={Jacob Cohen},
  journal={Educational and Psychological Measurement},
  year={1960},
  volume={20},
  pages={37 - 46},
  url={https://api.semanticscholar.org/CorpusID:15926286}
}

@article{landisKoch,
 ISSN = {0006341X, 15410420},
 URL = {http://www.jstor.org/stable/2529310},
 author = {J. Richard Landis and Gary G. Koch},
 journal = {Biometrics},
 number = {1},
 pages = {159--174},
 publisher = {International Biometric Society},
 title = {The Measurement of Observer Agreement for Categorical Data},
 urldate = {2026-05-03},
 volume = {33},
 year = {1977}
}

@article{Spearman1904,
  author  = {Spearman, Charles},
  title   = {The Proof and Measurement of Association between Two Things},
  journal = {The American Journal of Psychology},
  year    = {1904},
  volume  = {15},
  number  = {1},
  pages   = {72--101},
  doi     = {10.2307/1412159}
}

@article{Lev1949,
  author  = {Lev, Joseph},
  title   = {The point biserial coefficient of correlation},
  journal = {The Annals of Mathematical Statistics},
  year    = {1949},
  volume  = {20},
  number  = {1},
  pages   = {125--126},
  doi     = {10.1214/aoms/1177730103}
}

@inproceedings{koehn-2004-statistical,
    title = "Statistical Significance Tests for Machine Translation Evaluation",
    author = "Koehn, Philipp",
    editor = "Lin, Dekang  and
      Wu, Dekai",
    booktitle = "Proceedings of the 2004 Conference on Empirical Methods in Natural Language Processing",
    month = jul,
    year = "2004",
    address = "Barcelona, Spain",
    publisher = "Association for Computational Linguistics",
    url = "https://aclanthology.org/W04-3250/",
    pages = "388--395"
}

\appendix

\section{Share of Human-Post-edited Translations Identical to the Initial Human Translation}
\label{app:sameShare}

\begin{figure}[h!]
 \centering
 \includegraphics[width=\columnwidth]{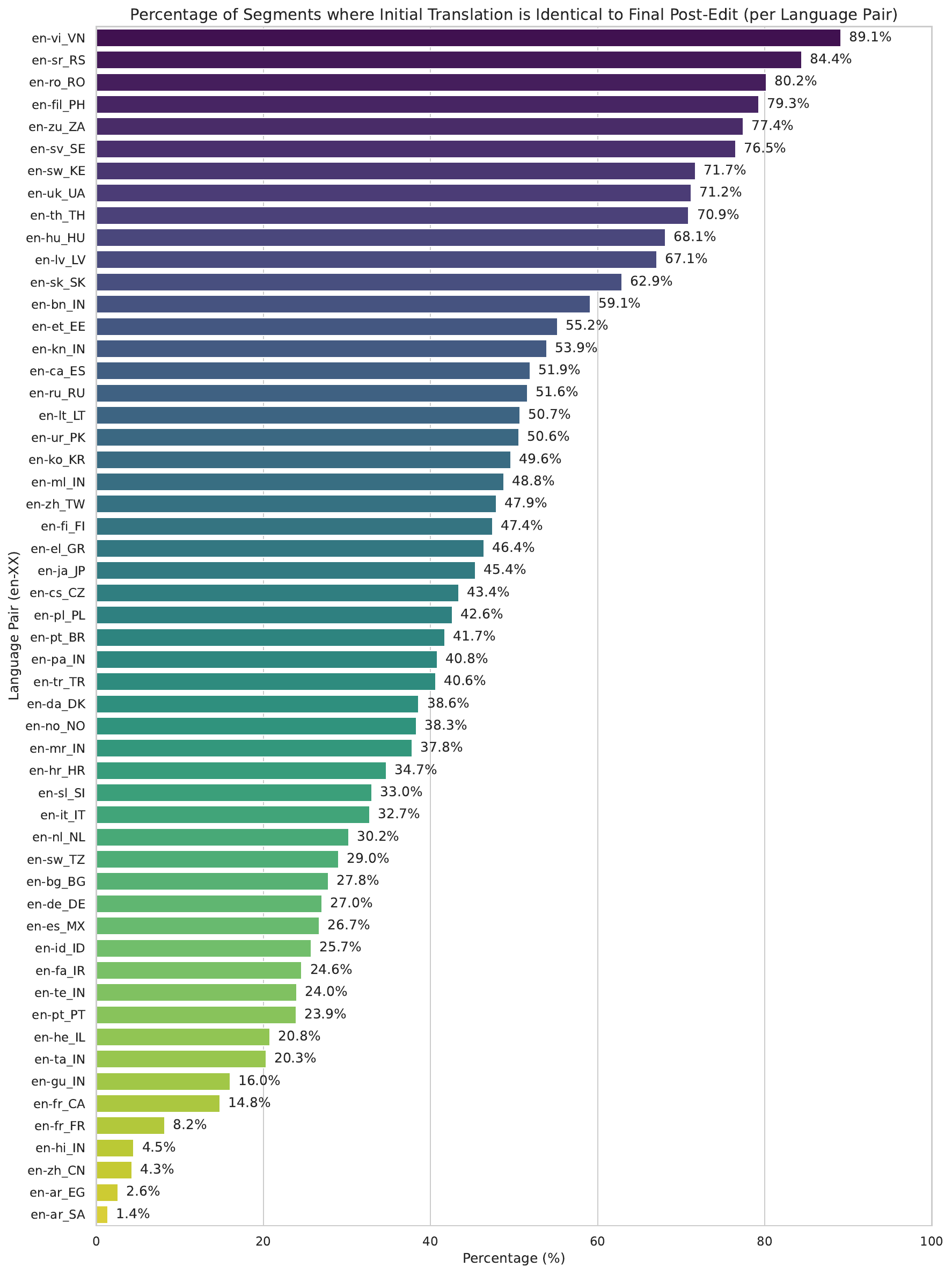}
 \caption{Share of segments for which the initial human translation was deemed acceptable by the human post-editor and left unchanged (per language pair).}\label{fig:identical_share}
\end{figure}

Figure~\ref{fig:identical_share} represents a per-language-pair (en-X\_locale) breakdown of the share of segments for which the initial human translation was left unaltered by the post-editor.

\section{Prompt Template Examples}
\label{app:prompts}

This appendix provides examples of the populated prompt templates fed to the LLMs for Tasks 1, 2, and 3.

To avoid biasing the literality of LLM output, prompt wording was kept neutral on this dimension: instructions referred only to translation ``quality'' and ``improvement,'' with no mention of literality, fluency, idiomaticity, or source-target proximity.

\scriptsize
\begin{directbox}
\textbf{System Prompt:} You are an expert professional translator. Your working languages are English and Italian. You always follow the output format instructions exactly. \\

\textbf{User Prompt (Thinking Mode Enabled):} \\
Translate the following English segment into Italian.
The segment is from a news article. \\

Segment to translate: \\
\textit{Tierra del Sol is pleased to present ``Vicente Siso: Memories of the Land and Water'' at the new gallery location in West Hollywood. Siso has been an artist in the Studio Arts Program since 2012, this marks his debut solo exhibition. Siso was born 1962 in Madrid and raised between Venezuela, Trinidad and Miami; he moved with his family to Southern California in his early 20s.} \\

For context, here are all segments of the document this segment belongs to (the segment to translate is marked with $\longrightarrow$):\\

    \textit{Siso's depictions of land, water center new gallery exhibition \\\\
    ``People Swimming in the Swimming Pool'' from 2022 is one Vicente Siso artwork that will display at Tierra del Sol Gallery beginning Jan. 13. (photo courtesy of Vicente Siso) \\\\
  $\longrightarrow$ Tierra del Sol is pleased to present \"Vicente Siso: Memories of the Land and Water\" at the new gallery location in West Hollywood. Siso has been an artist in the Studio Arts Program since 2012, this marks his debut solo exhibition. Siso was born 1962 in Madrid and raised between Venezuela, Trinidad and Miami; he moved with his family to Southern California in his early 20s. \\\\
    Masterfully working across subject matter, Siso has generated a prolific series of landscapes, portraits, and still-life works rendered in either acrylic, pastel, pencil or watercolor. Drawing from family portraits, his own reference photographs, and recollection, his colorful compositions demonstrate his range of interests and skill across media. Siso's tropical landscapes and seascapes reflect the geographies of his past, employing rich patterns and incorporating people to make meaningful connections between culture, memory and the environment. Siso titles his artworks in a mix of Spanish and English, signifying the celebrated and integral complexities of his life in Los Angeles County. ``Vicente Siso: Memories of the Land and Water'' opens on Saturday, Jan. 13, with a reception from 6-8 p.m. The exhibition is on view through Sunday, March 3. \\\\
    The Tierra del Sol Gallery is located at 7414 Santa Monica Blvd. For information, visit tierradelsolgallery.org. \\\\}

As a reminder, the segment you must translate into Italian is: \\
\textit{Tierra del Sol is pleased to present ``Vicente Siso: Memories of the Land and Water'' at the new gallery location in West Hollywood. Siso has been an artist in the Studio Arts Program since 2012, this marks his debut solo exhibition. Siso was born 1962 in Madrid and raised between Venezuela, Trinidad and Miami; he moved with his family to Southern California in his early 20s.} \\

Important: translate only this segment, not the full document. You may reflect on the task at hand before producing your final answer. When you are done, output your final translation between XML tags, like this:\\
\texttt{<translation>your translation here</translation>}\\\\

\textbf{Thinking-Mode-Disabled Variant:} \\
\ldots \\Important: translate only this segment, not the full document. Do not output any reasoning or chain of thought. Output only your translation between XML tags, like this:\\
\texttt{<translation>your translation here</translation>}

\end{directbox}

\begin{multibox}

\textbf{System Prompt:} You are an expert professional translator. Your working languages are English and French. You always follow the output format instructions exactly. When asked to produce multiple translations, output each one between \texttt{<translation></translation>} tags in sequence, with a brief assessment between each pair. \\

\textbf{User Prompt:} \\
Translate the following English segment into French. The segment is from a news article.\\

Segment to translate:\\
\textit{Siso's depictions of land, water center new gallery exhibition}\\
\\
For context, here are all segments of the document this segment belongs to (the segment to translate is marked with →):\\
\\
  → \textit{Siso's depictions of land, water center new gallery exhibition\\
  \\
    ``People Swimming in the Swimming Pool'' from 2022 is one Vicente Siso artwork that will display at Tierra del Sol Gallery beginning Jan. 13. (photo courtesy of Vicente Siso)\\
    \\
    Tierra del Sol is pleased to present \"Vicente Siso: Memories of the Land and Water\" at the new gallery location in West Hollywood. Siso has been an artist in the Studio Arts Program since 2012, this marks his debut solo exhibition. Siso was born 1962 in Madrid and raised between Venezuela, Trinidad and Miami; he moved with his family to Southern California in his early 20s.\\
    \\
    Masterfully working across subject matter, Siso has generated a prolific series of landscapes, portraits, and still-life works rendered in either acrylic, pastel, pencil or watercolor. Drawing from family portraits, his own reference photographs, and recollection, his colorful compositions demonstrate his range of interests and skill across media. Siso's tropical landscapes and seascapes reflect the geographies of his past, employing rich patterns and incorporating people to make meaningful connections between culture, memory and the environment. Siso titles his artworks in a mix of Spanish and English, signifying the celebrated and integral complexities of his life in Los Angeles County. ``Vicente Siso: Memories of the Land and Water'' opens on Saturday, Jan. 13, with a reception from 6-8 p.m. The exhibition is on view through Sunday, March 3.\\
    \\
    The Tierra del Sol Gallery is located at 7414 Santa Monica Blvd. For information, visit tierradelsolgallery.org.\\}
\\
As a reminder, the segment you must translate into French is:\\
\\
\textit{Siso's depictions of land, water center new gallery exhibition\\}
\\
Important: translate only this segment, not the full document. You may reflect on the task at hand before producing your first translation. When ready, output your first translation between XML tags:\\
\texttt{<translation>your translation here</translation>}\\
Then briefly assess what could be improved.\\
Based on this, produce an improved second translation between XML tags:\\
\texttt{<translation>your improved translation here</translation>}\\
Continue this process for a total of at least 2 and at most 5 translations. Each translation should genuinely attempt to improve on the previous one. If your first translation was already optimal, subsequent translations may be identical. Do not add any text after your final <translation></translation> tag.
\end{multibox}

\begin{pebox}
\textbf{System Prompt:} You are an expert professional translator and post-editor. Your working languages are English and Modern Greek. You always follow the output format instructions exactly. \\

\textbf{User Prompt (Thinking Mode Enabled):} \\
You are given an English source segment and its Modern Greek translation, from a news article.\\\\ 
Post-edit the translation if necessary to improve its quality. \\\\ 
Source segment:\\\\ 
\textit{Tierra del Sol is pleased to present ``Vicente Siso: Memories of the Land and Water'' at the new gallery location in West Hollywood. Siso has been an artist in the Studio Arts Program since 2012, this marks his debut solo exhibition. Siso was born 1962 in Madrid and raised between Venezuela, Trinidad and Miami; he moved with his family to Southern California in his early 20s.}\\\\ 
Modern Greek translation to post-edit:\\\\ 
\textgreek{Η Tierra del Sol με χαρά παρουσιάζει το «Vicente Siso: Memories of the Land and Water» στη νέα τοποθεσία της γκαλερί στο Δυτικό Χόλιγουντ. Ο Siso έχει υπάρξει καλλιτέχνης στο Studio Arts Program από το 2012, και αυτή είναι η πρώτη του ατομική έκθεση. Ο Siso γεννήθηκε το 1962 στη Μαδρίτη και μεγάλωσε στη Βενεζουέλα, το Τρινιντάντ και το Μαϊάμι. Μετακόμισε με την οικογένειά του στη Νότια Καλιφόρνια στις αρχές της δεκαετίας των 20 του.}\\\\ 
\\\\ 
For context, here are all source segments and their translations for the document this segment belongs to (the segment to post-edit is marked with $\longrightarrow$):\\\\ 
\\\\ 
    {[SRC]} \textit{Siso's depictions of land, water center new gallery exhibition}\\\\ 
    {[TGT]} \textgreek{Οι απεικονίσεις της γης, του νερού από τον Siso στο επίκεντρο νέας έκθεσης σε γκαλερί}\\\\ 
    {[SRC]} \textit{``People Swimming in the Swimming Pool'' from 2022 is one Vicente Siso artwork that will display at Tierra del Sol Gallery beginning Jan. 13. (photo courtesy of Vicente Siso)}\\\\ 
    {[TGT]} \textgreek{Το «People Swimming in the Swimming Pool» του 2022 είναι ένα από τα έργα τέχνης του Vicente Siso που θα εκτίθενται στην γκαλερί Tierra del Sol από τις 13 Ιαν. (η φωτογραφία είναι ευγενική παραχώρηση του Vicente Siso)}\\\\ 
    {[SRC]} \textit{Tierra del Sol is pleased to present ``Vicente Siso: Memories of the Land and Water'' at the new gallery location in West Hollywood. Siso has been an artist in the Studio Arts Program since 2012, this marks his debut solo exhibition. Siso was born 1962 in Madrid and raised between Venezuela, Trinidad and Miami; he moved with his family to Southern California in his early 20s.}\\\\ 
  $\longrightarrow$ {[TGT]} \textgreek{Η Tierra del Sol με χαρά παρουσιάζει το «Vicente Siso: Memories of the Land and Water» στη νέα τοποθεσία της γκαλερί στο Δυτικό Χόλιγουντ. Ο Siso έχει υπάρξει καλλιτέχνης στο Studio Arts Program από το 2012, και αυτή είναι η πρώτη του ατομική έκθεση. Ο Siso γεννήθηκε το 1962 στη Μαδρίτη και μεγάλωσε στη Βενεζουέλα, το Τρινιντάντ και το Μαϊάμι. Μετακόμισε με την οικογένειά του στη Νότια Καλιφόρνια στις αρχές της δεκαετίας των 20 του.}\\\\ 
    {[SRC]} \textit{Masterfully working across subject matter, Siso has generated a prolific series of landscapes, portraits, and still-life works rendered in either acrylic, pastel, pencil or watercolor. Drawing from family portraits, his own reference photographs, and recollection, his colorful compositions demonstrate his range of interests and skill across media. Siso's tropical landscapes and seascapes reflect the geographies of his past, employing rich patterns and incorporating people to make meaningful connections between culture, memory and the environment. Siso titles his artworks in a mix of Spanish and English, signifying the celebrated and integral complexities of his life in Los Angeles County. ``Vicente Siso: Memories of the Land and Water'' opens on Saturday, Jan. 13, with a reception from 6-8 p.m. The exhibition is on view through Sunday, March 3.}\\\\ 
    {[TGT]} \textgreek{Δουλεύοντας δεξιοτεχνικά σε διάφορα θέματα, ο Siso έχει δημιουργήσει πολυμελείς σειρές τοπίων, πορτρέτων και έργων νεκρής φύσης, τα οποία έχει αποδώσει είτε με ακρυλικό, παστέλ, μολύβι ή ακουαρέλα. Ζωγραφίζοντας από οικογενειακά πορτρέτα, δικές του φωτογραφίες αναφοράς, και από αναθύμηση, οι πολύχρωμες συνθέσεις του επιδεικνύουν το εύρος των ενδιαφερόντων του και των δεξιοτήτων του σε διάφορα μέσα. Τα τροπικά τοπία και τα θαλασσινά τοπία του Siso αντανακλούν τις γεωγραφίες του παρελθόντος του, μεταχειριζόμενα πλούσια μοτίβα και ενσωματώνοντας ανθρώπους για να κάνουν ουσιαστικές συνδέσεις μεταξύ κουλτούρας, μνήμης και του περιβάλλοντος. Ο Siso τιτλοδοτεί τα έργα τέχνης του με μείγμα Ισπανικών και Αγγλικών, δηλώνοντας τις περίφημες και αναπόσπαστες περιπλοκότητες της ζωής του στην κομητεία του Λος Άντζελες. Το «Vicente Siso: Memories of the Land and Water» ανοίγει το Σάββατο 13 Ιαν., με δεξίωση από τις 6 έως τις 8 μ.μ. Η έκθεση θα προβάλλεται έως την Κυριακή 3 Μαρτίου.}\\\\ 
    {[SRC]} \textit{The Tierra del Sol Gallery is located at 7414 Santa Monica Blvd. For information, visit tierradelsolgallery.org.}\\\\ 
    {[TGT]} \textgreek{Η γκαλερί Tierra del Sol βρίσκεται στο 7414 της Santa Monica Blvd. Για πληροφορίες, επισκεφθείτε το tierradelsolgallery.org.}\\\\ 
\\\\ 
As a reminder, the segment you must post-edit is:\\\\ 
{[SRC]} \textit{Tierra del Sol is pleased to present ``Vicente Siso: Memories of the Land and Water'' at the new gallery location in West Hollywood. Siso has been an artist in the Studio Arts Program since 2012, this marks his debut solo exhibition. Siso was born 1962 in Madrid and raised between Venezuela, Trinidad and Miami; he moved with his family to Southern California in his early 20s.}\\\\ 
{[TGT]} \textgreek{Η Tierra del Sol με χαρά παρουσιάζει το «Vicente Siso: Memories of the Land and Water» στη νέα τοποθεσία της γκαλερί στο Δυτικό Χόλιγουντ. Ο Siso έχει υπάρξει καλλιτέχνης στο Studio Arts Program από το 2012, και αυτή είναι η πρώτη του ατομική έκθεση. Ο Siso γεννήθηκε το 1962 στη Μαδρίτη και μεγάλωσε στη Βενεζουέλα, το Τρινιντάντ και το Μαϊάμι. Μετακόμισε με την οικογένειά του στη Νότια Καλιφόρνια στις αρχές της δεκαετίας των 20 του.}\\\\ 

Important: post-edit only the marked segment, not the full document. You may reflect on the post-editing task before producing your final answer. When you are done, output your final post-edited translation between XML tags, like this:\\ 
\texttt{<translation>your post-edited translation here</translation>}
\\\\
\textbf{Thinking-Mode-Disabled Variant:} \\
\ldots \\Important: post-edit only the marked segment, not the full document. Do not output any reasoning or chain of thought. Output only your post-edited translation between XML tags, like this:\\ 
\texttt{<translation>your post-edited translation here</translation>}

\end{pebox}

\normalsize


\section{Additional Information Regarding Data Generation and Extraction}
\label{app:dataGeneration}
This appendix provides additional information regarding data generation and extraction. All data generation was performed on CUDA-enabled hardware (H100 GPUs) on the Jean Zay HPC cluster.

\subsection{Data Generation}

\textbf{For the NLLB models}, inference was performed through the Transformers library \cite{wolf-etal-2020-transformers} in two ways, once using beam search with a beam size of 4, and once using ancestral sampling \cite{luo-etal-2024-diverge}. Target languages were specified using the NLLB language-specific prefix tokens (e.g., \texttt{ell\_Grek} for Modern Greek) to force the start of the decoded sequence. Due to the predefined language taxonomy of NLLB-200, certain regional varieties in the WMT24++ dataset could not be uniquely targeted. Specifically, Canadian French (\texttt{en-fr\_CA}), Brazilian Portuguese (\texttt{en-pt\_BR}), and regional Swahili variants (\texttt{en-sw\_KE}, \texttt{en-sw\_TZ}) were mapped to their respective base language codes (\texttt{fra\_Latn}, \texttt{por\_Latn}, and \texttt{swh\_Latn}). A manual audit of the resulting outputs confirmed that the French and Portuguese translations predominantly reflected European norms (France and Portugal, respectively), while the specific variety of the Swahili output could not be determined.

\textbf{For the LLMs}, inference was conducted using the vLLM engine \cite{kwon2023efficient} and the Transformers library \cite{wolf-etal-2020-transformers}. We used greedy decoding for Tasks 1 (single translation) and 3 (post-edition); for Task 2 (iterative translation), we used greedy decoding, and, for the fr\_FR target locale only (French of France), we also sampled 8 additional outputs per prompt, using a temperature of 1.0, with all completions sharing a single KV cache prefix via vLLM's prefix caching. For all tasks, the reasoning mode of the Qwen models was enabled or disabled through the ``\texttt{enable\_thinking}'' parameter exposed by the Transformers library for these models. For the Gemma 4 models, we injected or omitted the ``\texttt{<|think|>}'' token at the beginning of the system prompt to switch reasoning on or off. A maximum generation limit of 16,000 tokens was set for the ``thinking'' prompts, while non-reasoning outputs were capped at 1,024 tokens.

\subsection{Data Extraction}

Extraction of the translation candidates from the output of the NLLB models was trivial. 

Extraction of the translation candidates from LLM output ultimately relied on the models following the instruction given to them in the prompt to place all candidate translations between \texttt{<translation>...</translation>} tags.

Table \ref{tab:corpus_stats_tasks13} and Table \ref{tab:corpus_stats_task2} present the extraction yields across all three tasks (single translation, iterative translation, post-edition). Because many of our literality heuristics rely on the XLM-RoBERTa and LaBSE models, which have context windows capped at 512 tokens, and involve complex interactions between these models and the output of the Stanza language-specific word tokenizers, we chose to safely flag all segments longer than 450 tokens according to either of these models tokenizers as ``overlong,'' which also allowed us to discard hallucinations and formatting errors in LLM output. The tables also provide the extracted candidate translation ``survival rates'' at various MetricX-24 score thresholds (lower is better).

\begin{table}[th!]
\centering\small\setlength{\tabcolsep}{4pt}
\begin{adjustbox}{max width=0.95\columnwidth}
\begin{tabular}{llrrrrrrr}
\toprule
\textbf{Task} & \textbf{System} & \textbf{Expected} & \textbf{Extracted} & \textbf{Extr.\%} & \textbf{Flagged} & \textbf{N (MX$\leq$7)} & \textbf{N (MX$\leq$5)} & \textbf{N (MX$\leq$3)} \\
\midrule
\multirow{15}{*}{\textit{Task~1}} & Human (init.) & 51,840 & \cellcolor[rgb]{0.000,0.271,0.161}\textcolor[rgb]{1,1,1}{51,840} & 100.0\% & — & \cellcolor[rgb]{0.000,0.275,0.163}\textcolor[rgb]{1,1,1}{49,690} & \cellcolor[rgb]{0.000,0.387,0.207}\textcolor[rgb]{1,1,1}{44,526} & \cellcolor[rgb]{0.285,0.685,0.380}\textcolor[rgb]{0,0,0}{30,311} \\
 & NLLB-600M & 51,840 & \cellcolor[rgb]{0.000,0.271,0.161}\textcolor[rgb]{1,1,1}{51,840} & 100.0\% & 60 & \cellcolor[rgb]{0.440,0.762,0.459}\textcolor[rgb]{0,0,0}{25,794} & \cellcolor[rgb]{0.731,0.889,0.582}\textcolor[rgb]{0,0,0}{16,943} & \cellcolor[rgb]{0.928,0.972,0.695}\textcolor[rgb]{0,0,0}{8,534} \\
 & NLLB-3.3B & 51,840 & \cellcolor[rgb]{0.000,0.271,0.161}\textcolor[rgb]{1,1,1}{51,840} & 100.0\% & 19 & \cellcolor[rgb]{0.223,0.629,0.337}\textcolor[rgb]{1,1,1}{32,773} & \cellcolor[rgb]{0.533,0.803,0.499}\textcolor[rgb]{0,0,0}{23,098} & \cellcolor[rgb]{0.857,0.944,0.644}\textcolor[rgb]{0,0,0}{12,159} \\
 & Qwen3-8B$^{-}$ & 51,840 & \cellcolor[rgb]{0.000,0.271,0.161}\textcolor[rgb]{1,1,1}{51,840} & 100.0\% & 2 & \cellcolor[rgb]{0.386,0.735,0.431}\textcolor[rgb]{0,0,0}{27,465} & \cellcolor[rgb]{0.687,0.870,0.561}\textcolor[rgb]{0,0,0}{18,334} & \cellcolor[rgb]{0.909,0.964,0.682}\textcolor[rgb]{0,0,0}{9,534} \\
 & Qwen3-8B$^{+}$ & 51,840 & \cellcolor[rgb]{0.000,0.271,0.161}\textcolor[rgb]{1,1,1}{51,840} & 100.0\% & 34 & \cellcolor[rgb]{0.359,0.722,0.418}\textcolor[rgb]{0,0,0}{28,206} & \cellcolor[rgb]{0.663,0.860,0.551}\textcolor[rgb]{0,0,0}{19,167} & \cellcolor[rgb]{0.898,0.960,0.674}\textcolor[rgb]{0,0,0}{9,981} \\
 & Qwen3-32B$^{-}$ & 51,840 & \cellcolor[rgb]{0.000,0.271,0.161}\textcolor[rgb]{1,1,1}{51,840} & 100.0\% & 26 & \cellcolor[rgb]{0.179,0.572,0.299}\textcolor[rgb]{1,1,1}{35,113} & \cellcolor[rgb]{0.454,0.768,0.466}\textcolor[rgb]{0,0,0}{25,338} & \cellcolor[rgb]{0.823,0.929,0.626}\textcolor[rgb]{0,0,0}{13,609} \\
 & Qwen3-32B$^{+}$ & 51,840 & \cellcolor[rgb]{0.000,0.271,0.161}\textcolor[rgb]{1,1,1}{51,840} & 100.0\% & 85 & \cellcolor[rgb]{0.153,0.538,0.276}\textcolor[rgb]{1,1,1}{36,442} & \cellcolor[rgb]{0.427,0.755,0.452}\textcolor[rgb]{0,0,0}{26,172} & \cellcolor[rgb]{0.812,0.924,0.620}\textcolor[rgb]{0,0,0}{14,018} \\
 & OLMo-32B$^{+}$ & 51,840 & \cellcolor[rgb]{0.000,0.271,0.161}\textcolor[rgb]{1,1,1}{51,840} & 100.0\% & 408 & \cellcolor[rgb]{0.533,0.803,0.499}\textcolor[rgb]{0,0,0}{23,110} & \cellcolor[rgb]{0.779,0.910,0.605}\textcolor[rgb]{0,0,0}{15,123} & \cellcolor[rgb]{0.946,0.979,0.709}\textcolor[rgb]{0,0,0}{7,569} \\
 & OLMo-32B-SFT$^{+}$ & 51,840 & \cellcolor[rgb]{0.000,0.271,0.161}\textcolor[rgb]{1,1,1}{51,840} & 100.0\% & 91 & \cellcolor[rgb]{0.552,0.812,0.507}\textcolor[rgb]{0,0,0}{22,466} & \cellcolor[rgb]{0.801,0.920,0.615}\textcolor[rgb]{0,0,0}{14,420} & \cellcolor[rgb]{0.953,0.982,0.714}\textcolor[rgb]{0,0,0}{7,186} \\
 & Gemma4-31B$^{+}$ & 51,840 & \cellcolor[rgb]{0.000,0.288,0.168}\textcolor[rgb]{1,1,1}{50,982} & 98.3\% & — & \cellcolor[rgb]{0.000,0.314,0.178}\textcolor[rgb]{1,1,1}{47,766} & \cellcolor[rgb]{0.044,0.443,0.231}\textcolor[rgb]{1,1,1}{41,668} & \cellcolor[rgb]{0.400,0.742,0.438}\textcolor[rgb]{0,0,0}{27,046} \\
 & Gemma4-31B$^{-}$ & 51,840 & \cellcolor[rgb]{0.000,0.271,0.161}\textcolor[rgb]{1,1,1}{51,807} & 99.9\% & — & \cellcolor[rgb]{0.000,0.327,0.183}\textcolor[rgb]{1,1,1}{47,270} & \cellcolor[rgb]{0.061,0.457,0.237}\textcolor[rgb]{1,1,1}{40,812} & \cellcolor[rgb]{0.427,0.755,0.452}\textcolor[rgb]{0,0,0}{26,129} \\
 & Gemma4-E4B$^{+}$ & 51,840 & \cellcolor[rgb]{0.000,0.271,0.161}\textcolor[rgb]{1,1,1}{51,741} & 99.8\% & — & \cellcolor[rgb]{0.009,0.415,0.219}\textcolor[rgb]{1,1,1}{43,087} & \cellcolor[rgb]{0.186,0.581,0.305}\textcolor[rgb]{1,1,1}{34,768} & \cellcolor[rgb]{0.637,0.849,0.540}\textcolor[rgb]{0,0,0}{20,008} \\
 & Gemma4-E4B$^{-}$ & 51,840 & \cellcolor[rgb]{0.000,0.271,0.161}\textcolor[rgb]{1,1,1}{51,764} & 99.9\% & — & \cellcolor[rgb]{0.035,0.436,0.228}\textcolor[rgb]{1,1,1}{42,025} & \cellcolor[rgb]{0.212,0.615,0.328}\textcolor[rgb]{1,1,1}{33,463} & \cellcolor[rgb]{0.669,0.863,0.553}\textcolor[rgb]{0,0,0}{18,987} \\
 & NLLB-600M$^{\text{anc}}$ & 51,840 & \cellcolor[rgb]{0.000,0.271,0.161}\textcolor[rgb]{1,1,1}{51,838} & 100.0\% & — & \cellcolor[rgb]{0.928,0.972,0.695}\textcolor[rgb]{0,0,0}{8,436} & \cellcolor[rgb]{0.976,0.991,0.768}\textcolor[rgb]{0,0,0}{4,708} & \cellcolor[rgb]{0.989,0.996,0.838}\textcolor[rgb]{0,0,0}{2,195} \\
 & NLLB-3.3B$^{\text{anc}}$ & 51,840 & \cellcolor[rgb]{0.000,0.271,0.161}\textcolor[rgb]{1,1,1}{51,838} & 100.0\% & — & \cellcolor[rgb]{0.850,0.941,0.639}\textcolor[rgb]{0,0,0}{12,646} & \cellcolor[rgb]{0.950,0.981,0.712}\textcolor[rgb]{0,0,0}{7,325} & \cellcolor[rgb]{0.983,0.994,0.806}\textcolor[rgb]{0,0,0}{3,432} \\
\midrule
\multirow{10}{*}{\textit{Task~3}} & Qwen3-8B$^{-}$ & 51,840 & \cellcolor[rgb]{0.000,0.271,0.161}\textcolor[rgb]{1,1,1}{51,840} & 100.0\% & — & \cellcolor[rgb]{0.000,0.292,0.169}\textcolor[rgb]{1,1,1}{48,854} & \cellcolor[rgb]{0.018,0.422,0.222}\textcolor[rgb]{1,1,1}{42,717} & \cellcolor[rgb]{0.366,0.725,0.421}\textcolor[rgb]{0,0,0}{28,041} \\
 & Qwen3-8B$^{+}$ & 51,840 & \cellcolor[rgb]{0.000,0.271,0.161}\textcolor[rgb]{1,1,1}{51,840} & 100.0\% & — & \cellcolor[rgb]{0.000,0.309,0.176}\textcolor[rgb]{1,1,1}{47,941} & \cellcolor[rgb]{0.057,0.453,0.235}\textcolor[rgb]{1,1,1}{41,109} & \cellcolor[rgb]{0.420,0.752,0.449}\textcolor[rgb]{0,0,0}{26,432} \\
 & Qwen3-32B$^{-}$ & 51,840 & \cellcolor[rgb]{0.000,0.271,0.161}\textcolor[rgb]{1,1,1}{51,840} & 100.0\% & — & \cellcolor[rgb]{0.000,0.288,0.168}\textcolor[rgb]{1,1,1}{49,012} & \cellcolor[rgb]{0.013,0.419,0.220}\textcolor[rgb]{1,1,1}{42,947} & \cellcolor[rgb]{0.352,0.718,0.414}\textcolor[rgb]{0,0,0}{28,283} \\
 & Qwen3-32B$^{+}$ & 51,840 & \cellcolor[rgb]{0.000,0.271,0.161}\textcolor[rgb]{1,1,1}{51,840} & 100.0\% & — & \cellcolor[rgb]{0.000,0.296,0.171}\textcolor[rgb]{1,1,1}{48,599} & \cellcolor[rgb]{0.031,0.432,0.226}\textcolor[rgb]{1,1,1}{42,272} & \cellcolor[rgb]{0.386,0.735,0.431}\textcolor[rgb]{0,0,0}{27,306} \\
 & OLMo-32B$^{+}$ & 51,840 & \cellcolor[rgb]{0.000,0.271,0.161}\textcolor[rgb]{1,1,1}{51,840} & 100.0\% & 59 & \cellcolor[rgb]{0.044,0.443,0.231}\textcolor[rgb]{1,1,1}{41,566} & \cellcolor[rgb]{0.230,0.639,0.344}\textcolor[rgb]{1,1,1}{32,421} & \cellcolor[rgb]{0.676,0.866,0.556}\textcolor[rgb]{0,0,0}{18,751} \\
 & OLMo-32B-SFT$^{+}$ & 51,840 & \cellcolor[rgb]{0.000,0.271,0.161}\textcolor[rgb]{1,1,1}{51,840} & 100.0\% & 3 & \cellcolor[rgb]{0.082,0.474,0.244}\textcolor[rgb]{1,1,1}{39,870} & \cellcolor[rgb]{0.238,0.648,0.350}\textcolor[rgb]{1,1,1}{32,024} & \cellcolor[rgb]{0.669,0.863,0.553}\textcolor[rgb]{0,0,0}{19,001} \\
 & Gemma4-31B$^{+}$ & 51,840 & \cellcolor[rgb]{0.000,0.271,0.161}\textcolor[rgb]{1,1,1}{51,840} & 100.0\% & 978 & \cellcolor[rgb]{0.000,0.279,0.164}\textcolor[rgb]{1,1,1}{49,339} & \cellcolor[rgb]{0.000,0.370,0.200}\textcolor[rgb]{1,1,1}{45,333} & \cellcolor[rgb]{0.238,0.648,0.350}\textcolor[rgb]{1,1,1}{32,120} \\
 & Gemma4-31B$^{-}$ & 51,840 & \cellcolor[rgb]{0.000,0.271,0.161}\textcolor[rgb]{1,1,1}{51,840} & 100.0\% & 5 & \cellcolor[rgb]{0.000,0.271,0.161}\textcolor[rgb]{1,1,1}{49,888} & \cellcolor[rgb]{0.000,0.370,0.200}\textcolor[rgb]{1,1,1}{45,299} & \cellcolor[rgb]{0.249,0.663,0.360}\textcolor[rgb]{0,0,0}{31,505} \\
 & Gemma4-E4B$^{+}$ & 51,840 & \cellcolor[rgb]{0.000,0.271,0.161}\textcolor[rgb]{1,1,1}{51,840} & 100.0\% & 12 & \cellcolor[rgb]{0.000,0.275,0.163}\textcolor[rgb]{1,1,1}{49,556} & \cellcolor[rgb]{0.000,0.391,0.209}\textcolor[rgb]{1,1,1}{44,336} & \cellcolor[rgb]{0.298,0.692,0.387}\textcolor[rgb]{0,0,0}{29,999} \\
 & Gemma4-E4B$^{-}$ & 51,840 & \cellcolor[rgb]{0.000,0.271,0.161}\textcolor[rgb]{1,1,1}{51,840} & 100.0\% & 8 & \cellcolor[rgb]{0.000,0.284,0.166}\textcolor[rgb]{1,1,1}{49,280} & \cellcolor[rgb]{0.001,0.408,0.216}\textcolor[rgb]{1,1,1}{43,651} & \cellcolor[rgb]{0.325,0.705,0.400}\textcolor[rgb]{0,0,0}{29,119} \\
\bottomrule
\end{tabular}
\end{adjustbox}
\caption{Corpus statistics for Tasks~1 and~3. \textit{Expected} = number of source segments $\times$ 1 hypothesis. \textit{Extracted} = non-empty hypotheses. \textit{Flagged} = overlength hypotheses likely to be hallucinated or malformed. N columns show hypotheses surviving each MetricX threshold. Color intensity proportional to N.}
\label{tab:corpus_stats_tasks13}
\end{table}

Table \ref{tab:corpus_stats_tasks13} details the statistics for Task 1 (single translation) and Task 3 (post-editing), where the expected output is exactly one hypothesis per source segment (51,840 total expected translations per system). The data indicates robust instruction following across most LLMs, with near 100\% extraction rates. The NLLB models and the Qwen and OLMo families achieve a perfect 100\% extraction rate. The Gemma 4 models show a negligible drop in extraction rate (between 98.3\% and 99.9\%), indicating a few instances where the expected XML tags were either missing or malformed. The table also details the impact of the overlength flag, which, as previously stated, identifies sequence lengths that could cause positional embedding overflows during evaluation. While most models produce very few overlength hypotheses, the base OLMo-32B$^{+}$ model in Task 1 shows a minor spike (408 flagged segments).

The last three columns report the number of hypotheses surviving increasingly strict MetricX-24 score thresholds ($MX \leq 7, MX \leq 5, MX \leq 3$), where lower scores indicate higher quality. For Task 1, initial Human translations naturally retain the highest volume of segments at the strictest threshold (30,311 at $MX \leq 3$). Among the models, the Gemma 4 family, particularly Gemma4-31B$^{+}$, retains the highest number of high-quality translations. Notably, ancestral sampling with the NLLB models drastically reduces the number of segments surviving the MetricX-24 thresholds compared to standard beam search, highlighting the expected drop in translation quality when prioritizing diversity over probability. Task 3 (post-editing) generally shows higher retention rates at strict MetricX thresholds across all models compared to Task 1, which is expected since the initial human translation is given to the models as the starting point for post-editing.

\begin{table}[h!]
\centering\small\setlength{\tabcolsep}{4pt}
\begin{adjustbox}{max width=\columnwidth}
\begin{tabular}{lrrrrrrr}
\toprule
\textbf{System / Setting} & \textbf{Completions} & \textbf{Translations} & \textbf{Avg/comp.} & \textbf{Flagged} & \textbf{N (MX$\leq$7)} & \textbf{N (MX$\leq$5)} & \textbf{N (MX$\leq$3)} \\
\midrule
Gemma4-31B ($T=0$, all LPs) & 51,840 & \cellcolor[rgb]{0.920,0.969,0.690}\textcolor[rgb]{0,0,0}{150,004} & 2.89 & 1,536 & \cellcolor[rgb]{0.533,0.803,0.499}\textcolor[rgb]{0,0,0}{137,869} & \cellcolor[rgb]{0.637,0.849,0.540}\textcolor[rgb]{0,0,0}{118,689} & \cellcolor[rgb]{0.850,0.941,0.639}\textcolor[rgb]{0,0,0}{74,769} \\
Gemma4-31B ($T=1$, fr\textsubscript{FR} only) & 7,680 & \cellcolor[rgb]{0.994,0.998,0.866}\textcolor[rgb]{0,0,0}{22,780} & 2.97 & — & \cellcolor[rgb]{0.982,0.993,0.801}\textcolor[rgb]{0,0,0}{22,036} & \cellcolor[rgb]{0.983,0.994,0.806}\textcolor[rgb]{0,0,0}{19,751} & \cellcolor[rgb]{0.989,0.996,0.838}\textcolor[rgb]{0,0,0}{13,904} \\
Gemma4-E4B ($T=0$, all LPs) & 51,840 & \cellcolor[rgb]{0.905,0.963,0.679}\textcolor[rgb]{0,0,0}{164,919} & 3.18 & 58 & \cellcolor[rgb]{0.546,0.809,0.504}\textcolor[rgb]{0,0,0}{135,675} & \cellcolor[rgb]{0.693,0.873,0.564}\textcolor[rgb]{0,0,0}{108,700} & \cellcolor[rgb]{0.898,0.960,0.674}\textcolor[rgb]{0,0,0}{60,085} \\
Gemma4-E4B ($T=1$, fr\textsubscript{FR} only) & 7,680 & \cellcolor[rgb]{0.993,0.997,0.860}\textcolor[rgb]{0,0,0}{26,085} & 3.40 & 4 & \cellcolor[rgb]{0.980,0.993,0.790}\textcolor[rgb]{0,0,0}{24,185} & \cellcolor[rgb]{0.983,0.994,0.806}\textcolor[rgb]{0,0,0}{20,000} & \cellcolor[rgb]{0.990,0.996,0.844}\textcolor[rgb]{0,0,0}{11,951} \\
OLMo-32B-SFT ($T=0$, all LPs) & 51,840 & \cellcolor[rgb]{0.758,0.901,0.595}\textcolor[rgb]{0,0,0}{270,052} & 5.21 & 19,072 & \cellcolor[rgb]{0.578,0.823,0.517}\textcolor[rgb]{0,0,0}{129,223} & \cellcolor[rgb]{0.812,0.924,0.620}\textcolor[rgb]{0,0,0}{83,068} & \cellcolor[rgb]{0.953,0.982,0.714}\textcolor[rgb]{0,0,0}{41,964} \\
OLMo-32B-SFT ($T=1$, fr\textsubscript{FR} only) & 7,680 & \cellcolor[rgb]{0.985,0.994,0.817}\textcolor[rgb]{0,0,0}{51,692} & 6.73 & 360 & \cellcolor[rgb]{0.974,0.990,0.757}\textcolor[rgb]{0,0,0}{30,747} & \cellcolor[rgb]{0.982,0.993,0.801}\textcolor[rgb]{0,0,0}{21,449} & \cellcolor[rgb]{0.990,0.996,0.844}\textcolor[rgb]{0,0,0}{11,657} \\
OLMo-32B ($T=0$, all LPs) & 51,840 & \cellcolor[rgb]{0.000,0.271,0.161}\textcolor[rgb]{1,1,1}{850,527} & 16.41 & 3,592 & \cellcolor[rgb]{0.000,0.271,0.161}\textcolor[rgb]{1,1,1}{297,065} & \cellcolor[rgb]{0.249,0.663,0.360}\textcolor[rgb]{0,0,0}{187,498} & \cellcolor[rgb]{0.785,0.913,0.608}\textcolor[rgb]{0,0,0}{88,342} \\
Qwen3-32B ($T=0$, all LPs) & 51,840 & \cellcolor[rgb]{0.920,0.969,0.690}\textcolor[rgb]{0,0,0}{152,300} & 2.94 & 1,664 & \cellcolor[rgb]{0.693,0.873,0.564}\textcolor[rgb]{0,0,0}{108,836} & \cellcolor[rgb]{0.828,0.931,0.628}\textcolor[rgb]{0,0,0}{79,475} & \cellcolor[rgb]{0.950,0.981,0.712}\textcolor[rgb]{0,0,0}{43,148} \\
Qwen3-32B ($T=1$, fr\textsubscript{FR} only) & 7,680 & \cellcolor[rgb]{0.992,0.997,0.855}\textcolor[rgb]{0,0,0}{28,986} & 3.77 & 40 & \cellcolor[rgb]{0.980,0.993,0.790}\textcolor[rgb]{0,0,0}{23,861} & \cellcolor[rgb]{0.984,0.994,0.811}\textcolor[rgb]{0,0,0}{18,721} & \cellcolor[rgb]{0.991,0.997,0.849}\textcolor[rgb]{0,0,0}{11,127} \\
Qwen3-8B ($T=0$, all LPs) & 51,840 & \cellcolor[rgb]{0.920,0.969,0.690}\textcolor[rgb]{0,0,0}{151,968} & 2.93 & 2,880 & \cellcolor[rgb]{0.801,0.920,0.615}\textcolor[rgb]{0,0,0}{85,511} & \cellcolor[rgb]{0.902,0.961,0.676}\textcolor[rgb]{0,0,0}{58,579} & \cellcolor[rgb]{0.974,0.990,0.757}\textcolor[rgb]{0,0,0}{30,618} \\
Qwen3-8B ($T=1$, fr\textsubscript{FR} only) & 7,680 & \cellcolor[rgb]{0.993,0.997,0.860}\textcolor[rgb]{0,0,0}{26,059} & 3.39 & 49 & \cellcolor[rgb]{0.980,0.993,0.790}\textcolor[rgb]{0,0,0}{23,237} & \cellcolor[rgb]{0.985,0.994,0.817}\textcolor[rgb]{0,0,0}{18,250} & \cellcolor[rgb]{0.992,0.997,0.855}\textcolor[rgb]{0,0,0}{10,385} \\
\midrule
\textbf{Total} & 349,440 & \textbf{1,895,372} & 5.42 & 29,255 & 1,018,245 & 734,180 & 397,950 \\
\bottomrule
\end{tabular}
\end{adjustbox}
\caption{Corpus statistics for Task~2 (iterative translation). Each \textit{completion} is one inference call; a completion may contain multiple translations. \textit{Avg/comp.} = mean number of translations extracted per completion. $T=0$: greedy decoding, all 54 LPs; $T=1$: sampling with a temperature of 1, \textsc{en-fr\textsubscript{FR}} only. Color intensity proportional to N.}
\label{tab:corpus_stats_task2}
\end{table}

Table \ref{tab:corpus_stats_task2} presents the statistics for Task 2 (iterative translation), where models were instructed to generate multiple distinct translation candidates per prompt, each ideally improving upon the last. The prompts' instruction that the models output between 2 and 5 translations for a given source segments was generally followed by the Qwen-3 and Gemma-4 models, with an average of 3 translations per completion, but systematically ignored by their Olmo-3 counterparts, particularly by OLMo-32B which generated an average of 16.41 translations per source segment at a temperature of 0. This high volume is accompanied by a disproportionately high number of overlength flags (3,592 for OLMo-32B and 19,072 for OLMo-32B-SFT) and a steep drop-off at strict MetricX-24 thresholds, suggesting that many of these extra generated sequences are low quality or hallucinated noise.

\section{Per-Language-Pair MetricX-24 Scores and Significance Testing}
\label{app:LPmx24}

This appendix provides a detailed, per-language-pair breakdown of the MetricX-24 evaluation scores for all machine translation segments generated during our experiments. All scores were computed using the final human-post-edited translations from the WMT24++ dataset as the gold-standard reference. 

\subsection{Task 1: Single Translation}

\begin{figure*}
    \centering
    \includegraphics[width=1\linewidth]{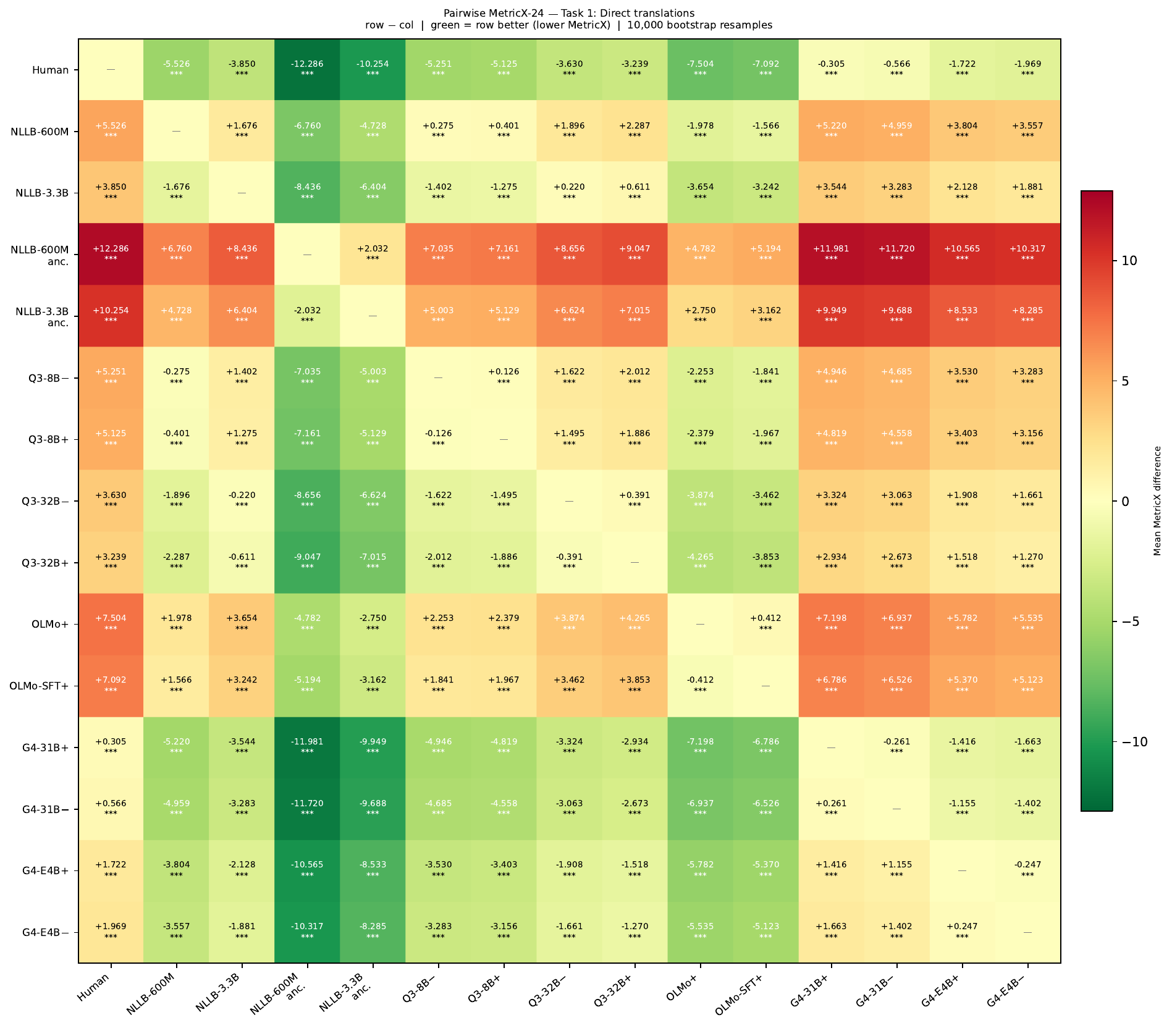}
    \caption{Pairwise comparison of the systems and the human translators for task 1, illustrating the mean difference in MetricX-24 scores alongside statistical significance computed via bootstrap resampling ($N=10,000$).}
    \label{fig:task1_mx}
\end{figure*}

\begin{table*}[t]
\centering
\begin{adjustbox}{max width=\textwidth, max totalheight=\textheight}
\renewcommand{\arraystretch}{0.85}
\setlength{\tabcolsep}{3pt}
\small
\begin{tabular}{lrrrrrrrrrrrrrrr}
\toprule
\textbf{LP} & \textbf{Human} & \textbf{NLLB-600M} & \textbf{NLLB-3.3B} & \textbf{NLLB-600M$^{\text{anc}}$} & \textbf{NLLB-3.3B$^{\text{anc}}$} & \textbf{Qwen3-8B$^{-}$} & \textbf{Qwen3-8B$^{+}$} & \textbf{Qwen3-32B$^{-}$} & \textbf{Qwen3-32B$^{+}$} & \textbf{OLMo-32B$^{+}$} & \textbf{OLMo-32B-SFT$^{+}$} & \textbf{Gemma4-31B$^{+}$} & \textbf{Gemma4-31B$^{-}$} & \textbf{Gemma4-E4B$^{+}$} & \textbf{Gemma4-E4B$^{-}$} \\
\midrule
en-de\_DE & \cellcolor[rgb]{0.866,0.890,0.096}\textcolor[rgb]{0,0,0}{1.338} & \cellcolor[rgb]{0.449,0.814,0.335}\textcolor[rgb]{0,0,0}{5.275} & \cellcolor[rgb]{0.606,0.851,0.237}\textcolor[rgb]{0,0,0}{3.804} & \cellcolor[rgb]{0.138,0.537,0.555}\textcolor[rgb]{1,1,1}{13.210} & \cellcolor[rgb]{0.120,0.622,0.535}\textcolor[rgb]{1,1,1}{11.024} & \cellcolor[rgb]{0.689,0.865,0.183}\textcolor[rgb]{0,0,0}{3.025} & \cellcolor[rgb]{0.710,0.869,0.169}\textcolor[rgb]{0,0,0}{2.775} & \cellcolor[rgb]{0.783,0.879,0.125}\textcolor[rgb]{0,0,0}{2.131} & \cellcolor[rgb]{0.783,0.879,0.125}\textcolor[rgb]{0,0,0}{2.111} & \cellcolor[rgb]{0.369,0.789,0.383}\textcolor[rgb]{0,0,0}{6.242} & \cellcolor[rgb]{0.536,0.836,0.282}\textcolor[rgb]{0,0,0}{4.451} & \cellcolor[rgb]{0.846,0.887,0.100}\textcolor[rgb]{0,0,0}{1.486} & \cellcolor[rgb]{0.846,0.887,0.100}\textcolor[rgb]{0,0,0}{1.548} & \cellcolor[rgb]{0.815,0.883,0.110}\textcolor[rgb]{0,0,0}{1.791} & \cellcolor[rgb]{0.794,0.881,0.120}\textcolor[rgb]{0,0,0}{1.988} \\
\bottomrule
\end{tabular}
\end{adjustbox}
\caption{\small Mean MetricX-24 per LP — single direct translations (MetricX$\leq$25). (lower = better)}
\label{tab:appendix_metricx_trans}
\end{table*}

Table \ref{tab:appendix_metricx_trans} presents the mean MetricX-24 scores for Task 1, where models were prompted to generate a single direct translation of the English source segments (lower scores indicate better translation quality). To rigorously assess the performance hierarchy, Figure~\ref{fig:task1_mx} presents a pairwise comparison of these systems, illustrating the mean difference in MetricX-24 scores alongside statistical significance computed via bootstrap resampling ($N=10,000$). Several distinct trends emerge across system architectures, decoding strategies, and language families:

\paragraph{Overall Performance Hierarchy.} The human translations set the highest quality standard. This is unsurprising, given that the final post-edited translation used as reference is itself based on that initial human translation.  The heatmap confirms that human translations are significantly better than the best-performing MT system (Gemma4-31B$^{+}$), outperforming it by a margin of 0.305 MetricX points ($p < 0.001$). Among the MT systems, the general hierarchy of performance is: Gemma4-31B > Gemma4-E4B > Qwen3-32B > Qwen3-8B > NLLB-3.3B > NLLB-600M > OLMo-32B.

\paragraph{Impact of ``Thinking'' (Reasoning Mode).} The pairwise heatmap provides strong statistical validation for the utility of reasoning during translation. Enabling the thinking mode ($+$) yields a highly significant improvement over the standard generation mode ($-$). For example, Gemma4-31B$^{+}$ significantly outperforms Gemma4-31B$^{-}$ by 0.261 points ($p < 0.001$), and Qwen3-32B$^{+}$ outperforms its non-reasoning counterpart by 0.391 points ($p < 0.001$).

\paragraph{High-Resource vs. Low-Resource Disparities.} As detailed in Table \ref{tab:appendix_metricx_trans}, all systems exhibit quality variations depending on the target language. High-resource European languages, such as German (en-de\_DE) and Spanish (en-es\_MX), see near-human-parity scores from the leading LLMs (e.g., Gemma4-31B$^{+}$ scores 1.486 on German compared to the human 1.338). Conversely, performance degrades sharply on lower-resource languages like Latvian (en-lv\_LV), Zulu (en-zu\_ZA), and Swahili (en-sw\_KE, en-sw\_TZ).

\paragraph{Systemic Failures in Specific LLMs.} While the Gemma-4 and Qwen-3 models handle linguistic diversity relatively well, the OLMo-32B models struggle significantly outside of high-resource languages. This is most visible in Table~\ref{tab:appendix_metricx_trans} through their catastrophic degradation on African languages, scoring upwards of 20.0 on Swahili and Zulu, indicating a near-total failure to produce viable translations in these target locales. The pairwise matrix confirms that OLMo-32B is significantly outperformed by almost all other tested systems, with a paradoxical advantage for the more primitive SFT checkpoint over its final RLVR-finetuned counterpart.

\paragraph{Decoding Strategies in NLLB.} The data starkly illustrates the penalty incurred when optimizing for diversity rather than probability in traditional NMT models. The heatmap shows a massive, statistically significant degradation when applying ancestral sampling to the NLLB models. NLLB-600M with standard beam search outperforms its ancestral sampling variant by a massive 6.760 MetricX points difference ($p < 0.001$).

\subsection{Task 2: Iterative Translation}

\begin{figure}
    \centering
    \includegraphics[width=1\linewidth]{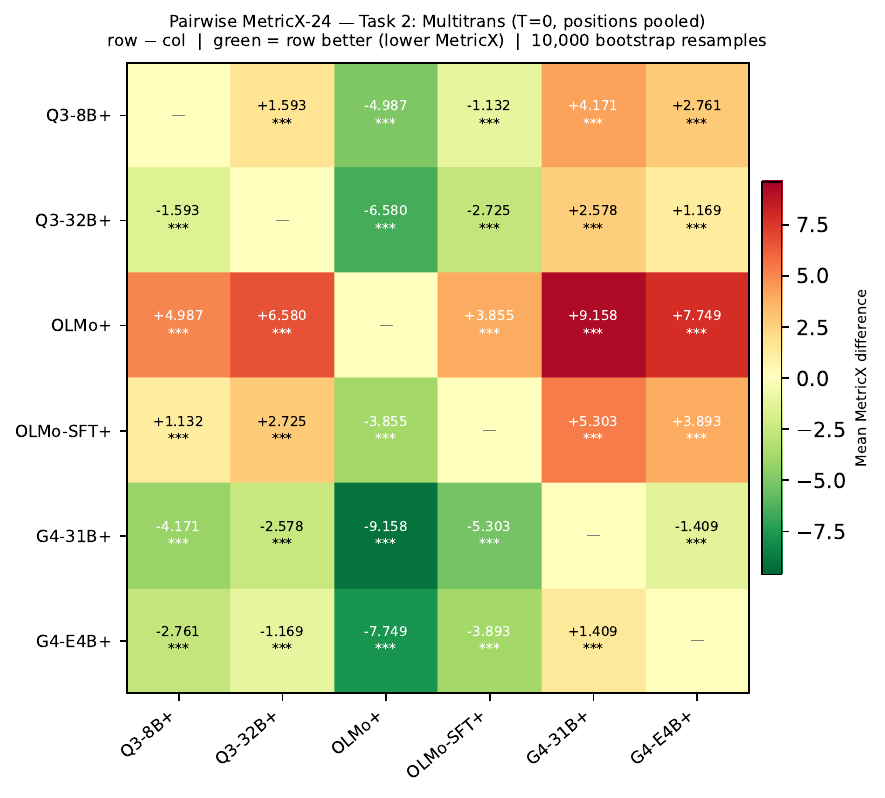}
    \caption{Pairwise comparison of the systems for task 2 (iterative translation) illustrating the mean difference in MetricX-24 scores alongside statistical significance computed via bootstrap resampling ($N=10,000$).}
    \label{fig:metricX_task2_heatmap}
\end{figure}

Table \ref{tab:appendix_metricx_mt} presents the mean MetricX-24 scores for Task 2 (iterative translation), pooling all translation candidates generated per prompt under greedy decoding ($T=0$). Only the thinking-enabled ($+$) setting was used for this task during inference. Figure~\ref{fig:metricX_task2_heatmap} complements this table with a pairwise significance heatmap.

\begin{table}[th!]
\centering
\begin{adjustbox}{max width=\columnwidth, max totalheight=\textheight}
\renewcommand{\arraystretch}{0.85}
\setlength{\tabcolsep}{3pt}
\small
\begin{tabular}{lrrrrrr}
\toprule
\textbf{LP} & \textbf{Qwen3-8B$^{+}$} & \textbf{Qwen3-32B$^{+}$} & \textbf{OLMo-32B$^{+}$} & \textbf{OLMo-32B-SFT$^{+}$} & \textbf{Gemma4-31B$^{+}$} & \textbf{Gemma4-E4B$^{+}$} \\
\midrule
en-zh\_CN & \cellcolor[rgb]{0.731,0.872,0.156}\textcolor[rgb]{0,0,0}{2.609} & \cellcolor[rgb]{0.741,0.873,0.150}\textcolor[rgb]{0,0,0}{2.466} & \cellcolor[rgb]{0.296,0.762,0.424}\textcolor[rgb]{0,0,0}{7.100} & \cellcolor[rgb]{0.647,0.858,0.210}\textcolor[rgb]{0,0,0}{3.361} & \cellcolor[rgb]{0.762,0.876,0.137}\textcolor[rgb]{0,0,0}{2.277} & \cellcolor[rgb]{0.731,0.872,0.156}\textcolor[rgb]{0,0,0}{2.569} \\
en-de\_DE & \cellcolor[rgb]{0.710,0.869,0.169}\textcolor[rgb]{0,0,0}{2.743} & \cellcolor[rgb]{0.773,0.878,0.131}\textcolor[rgb]{0,0,0}{2.196} & \cellcolor[rgb]{0.220,0.726,0.466}\textcolor[rgb]{0,0,0}{8.185} & \cellcolor[rgb]{0.487,0.824,0.312}\textcolor[rgb]{0,0,0}{4.955} & \cellcolor[rgb]{0.835,0.886,0.103}\textcolor[rgb]{0,0,0}{1.595} & \cellcolor[rgb]{0.804,0.882,0.115}\textcolor[rgb]{0,0,0}{1.922} \\
en-zh\_TW & \cellcolor[rgb]{0.699,0.867,0.176}\textcolor[rgb]{0,0,0}{2.880} & \cellcolor[rgb]{0.720,0.870,0.163}\textcolor[rgb]{0,0,0}{2.717} & \cellcolor[rgb]{0.274,0.752,0.437}\textcolor[rgb]{0,0,0}{7.390} & \cellcolor[rgb]{0.536,0.836,0.282}\textcolor[rgb]{0,0,0}{4.481} & \cellcolor[rgb]{0.741,0.873,0.150}\textcolor[rgb]{0,0,0}{2.498} & \cellcolor[rgb]{0.720,0.870,0.163}\textcolor[rgb]{0,0,0}{2.696} \\
en-es\_MX & \cellcolor[rgb]{0.616,0.853,0.230}\textcolor[rgb]{0,0,0}{3.631} & \cellcolor[rgb]{0.689,0.865,0.183}\textcolor[rgb]{0,0,0}{3.016} & \cellcolor[rgb]{0.181,0.701,0.488}\textcolor[rgb]{0,0,0}{8.815} & \cellcolor[rgb]{0.487,0.824,0.312}\textcolor[rgb]{0,0,0}{4.935} & \cellcolor[rgb]{0.720,0.870,0.163}\textcolor[rgb]{0,0,0}{2.713} & \cellcolor[rgb]{0.678,0.864,0.190}\textcolor[rgb]{0,0,0}{3.044} \\
\bottomrule
\end{tabular}
\end{adjustbox}
\caption{\small Mean MetricX-24 per LP — iterative multitrans, $T=0$, all positions pooled (MetricX$\leq$25). Thinking-ON models only. (lower = better)}
\label{tab:appendix_metricx_mt}
\end{table}

The results largely mirror the dynamics observed in Task 1, with a few key takeaways:

\paragraph{Consistent Performance Hierarchy.} The models maintain a strict quality hierarchy, ranking from best to worst as follows: Gemma4-31B$^{+}$ > Gemma4-E4B$^{+}$ > Qwen3-32B$^{+}$ > Qwen3-8B$^{+}$ > OLMo-32B-SFT$^{+}$ > OLMo-32B$^{+}$.

\paragraph{Strong Statistical Significance.} As illustrated by the heatmap, every step up this performance hierarchy represents a highly significant improvement ($p < 0.001$). For instance, Gemma4-31B$^{+}$ significantly outperforms its smaller edge-optimized counterpart (Gemma4-E4B$^{+}$) by 1.409 MetricX points, and beats the best Qwen model (Qwen3-32B$^{+}$) by 2.578 points, both with $p < 0.001$.

\paragraph{The OLMo paradox.} The OLMo-3 family continues to severely lag behind the Gemma-4 and Qwen-3 models. The heatmap also clearly demonstrates that for these models RL finetuning led to degraded performance over the SFT checkpoint, as OLMo-32B-SFT$^{+}$ significantly outperforms the RL-finetuned OLMo-32B$^{+}$ by a margin of 3.855 points ($p < 0.001$).

\subsection{Task 3: Post-Editing}

\begin{figure}
    \centering
    \includegraphics[width=1\linewidth]{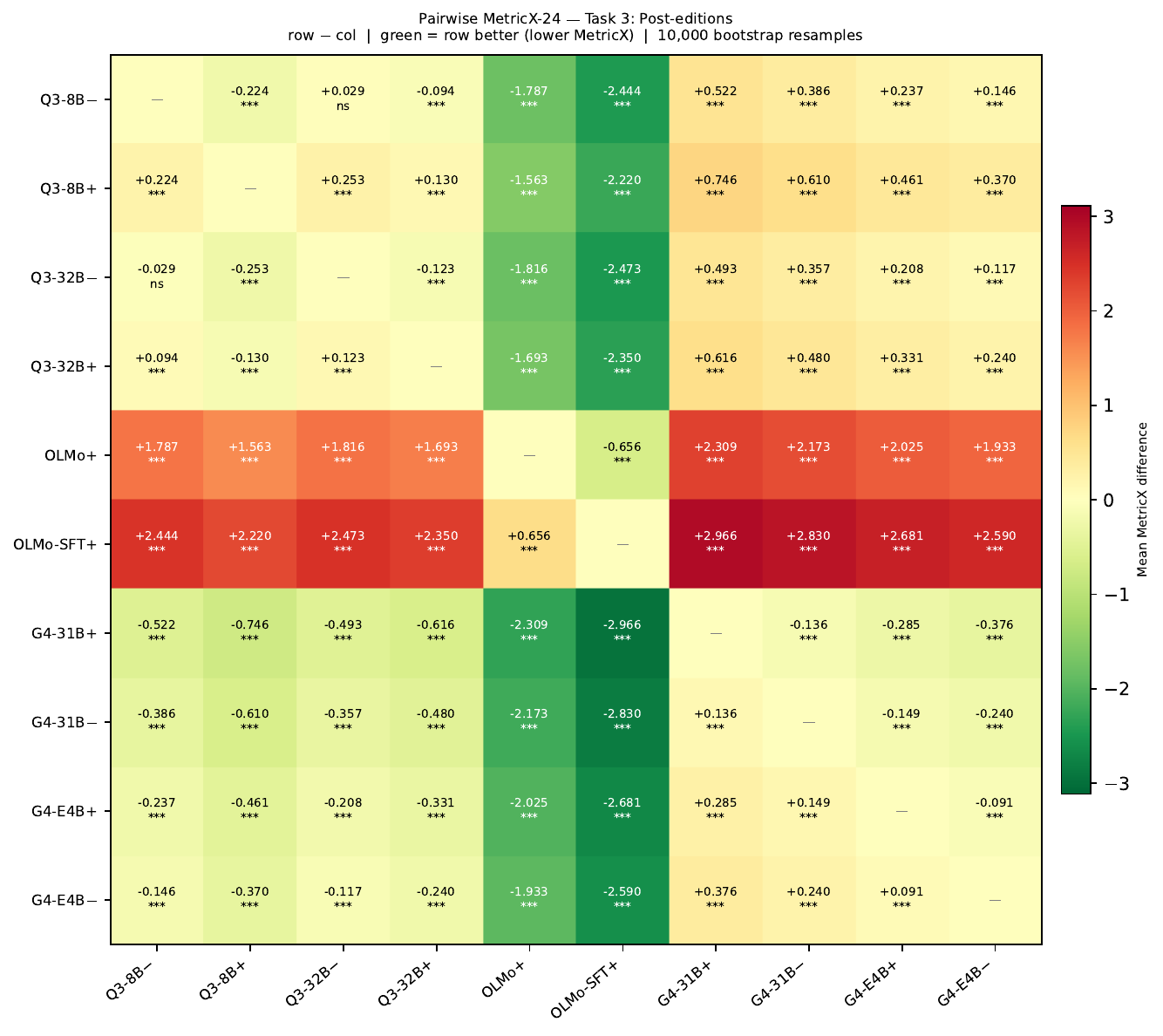}
    \caption{Pairwise comparison of the systems for task 3 (post-editing the initial human translation) illustrating the mean difference in MetricX-24 scores alongside statistical significance computed via bootstrap resampling ($N=10,000$).}
    \label{fig:metricX_task3_heatmap}
\end{figure}

Table \ref{tab:appendix_metricx_pe} details the mean MetricX-24 scores for Task 3, where models were tasked with post-editing the initial human translations. To illustrate the impact of the post-editing phase, we report the delta ($\Delta$) between the system's post-edited output and the initial human translation score. Figure~\ref{fig:metricX_task3_heatmap} provides the corresponding pairwise significance heatmap.

\begin{table*}[t]
\centering
\begin{adjustbox}{max width=\linewidth, max totalheight=\textheight}
\renewcommand{\arraystretch}{0.85}
\setlength{\tabcolsep}{3pt}
\small
\begin{tabular}{lrrrrrrrrrr}
\toprule
\textbf{LP} & \textbf{Qwen3-8B$^{-}$} & \textbf{Qwen3-8B$^{+}$} & \textbf{Qwen3-32B$^{-}$} & \textbf{Qwen3-32B$^{+}$} & \textbf{OLMo-32B$^{+}$} & \textbf{OLMo-32B-SFT$^{+}$} & \textbf{Gemma4-31B$^{+}$} & \textbf{Gemma4-31B$^{-}$} & \textbf{Gemma4-E4B$^{+}$} & \textbf{Gemma4-E4B$^{-}$} \\
\midrule
en-de\_DE & \cellcolor[rgb]{0.856,0.889,0.097}\textcolor[rgb]{0,0,0}{1.462 \textcolor{dkred}{\tiny (+0.124)}} & \cellcolor[rgb]{0.846,0.887,0.100}\textcolor[rgb]{0,0,0}{1.508 \textcolor{dkred}{\tiny (+0.170)}} & \cellcolor[rgb]{0.856,0.889,0.097}\textcolor[rgb]{0,0,0}{1.385 \textcolor{dkred}{\tiny (+0.047)}} & \cellcolor[rgb]{0.856,0.889,0.097}\textcolor[rgb]{0,0,0}{1.441 \textcolor{dkred}{\tiny (+0.103)}} & \cellcolor[rgb]{0.783,0.879,0.125}\textcolor[rgb]{0,0,0}{2.144 \textcolor{dkred}{\tiny (+0.806)}} & \cellcolor[rgb]{0.710,0.869,0.169}\textcolor[rgb]{0,0,0}{2.815 \textcolor{dkred}{\tiny (+1.477)}} & \cellcolor[rgb]{0.866,0.890,0.096}\textcolor[rgb]{0,0,0}{1.314 \textcolor{dkgreen}{\tiny (-0.024)}} & \cellcolor[rgb]{0.856,0.889,0.097}\textcolor[rgb]{0,0,0}{1.437 \textcolor{dkred}{\tiny (+0.099)}} & \cellcolor[rgb]{0.866,0.890,0.096}\textcolor[rgb]{0,0,0}{1.315 \textcolor{dkgreen}{\tiny (-0.023)}} & \cellcolor[rgb]{0.866,0.890,0.096}\textcolor[rgb]{0,0,0}{1.360 \textcolor{dkred}{\tiny (+0.022)}} \\
\bottomrule
\end{tabular}
\end{adjustbox}
\caption{\small Mean MetricX-24 per LP — post-editions (MetricX$\leq$25). $\Delta$ = PE score $-$ initial human translation score ({\color{dkgreen}green} = PE better, {\color{dkred}red} = PE worse). (lower = better)}
\label{tab:appendix_metricx_pe}
\end{table*}

The post-editing task reveals several surprising behavioral divergences among the LLMs that were not present in direct translation:

\paragraph{Gemma 4 Excels and Improves Human Drafts.} The Gemma 4 family, particularly Gemma4-31B$^{+}$, remains the dominant system architecture. Notably, as indicated by the negative $\Delta$ values (green) in Table \ref{tab:appendix_metricx_pe}, Gemma4-31B$^{+}$ frequently succeeds in its intended task: it actually improves upon the initial human translations across a wide variety of language pairs (e.g., \textsc{en-fr\_FR}, \textsc{en-zh\_TW}, \textsc{en-ru\_RU}).

\paragraph{The Reasoning Paradox.} While reasoning capabilities unambiguously improved direct translation, their impact on post-editing is heavily model-dependent. The heatmap confirms that reasoning significantly aids the Gemma 4 architecture; Gemma4-31B$^{+}$ outperforms its non-reasoning counterpart (Gemma4-31B$^{-}$) by 0.136 MetricX points ($p < 0.001$). Conversely, reasoning actually degrades the performance of the Qwen 3 models during post-editing. Qwen3-32B$^{-}$ significantly outperforms the thinking-enabled Qwen3-32B$^{+}$ by 0.123 points ($p < 0.001$), and a similar highly significant inversion is seen between the 8B variants ($p < 0.001$). This suggests that Qwen's reasoning traces may lead to over-correction or unnecessary divergence when provided with a viable human draft.
   
\paragraph{The OLMo-3 Checkpoint Anomaly.} Both OLMo models heavily degrade the initial human translations (large positive $\Delta$ values) across nearly all evaluated language pairs. The pairwise matrix highlights that the RL-finetuned OLMo-32B$^{+}$ significantly outperforms the OLMo-32B-SFT$^{+}$ variant by a margin of 0.656 points ($p < 0.001$), reversing the hierarchy observed for the previous tasks.

\section{Human validation}
\label{app:humanval}

\begin{figure}[h]
    \centering
    \includegraphics[width=\columnwidth]{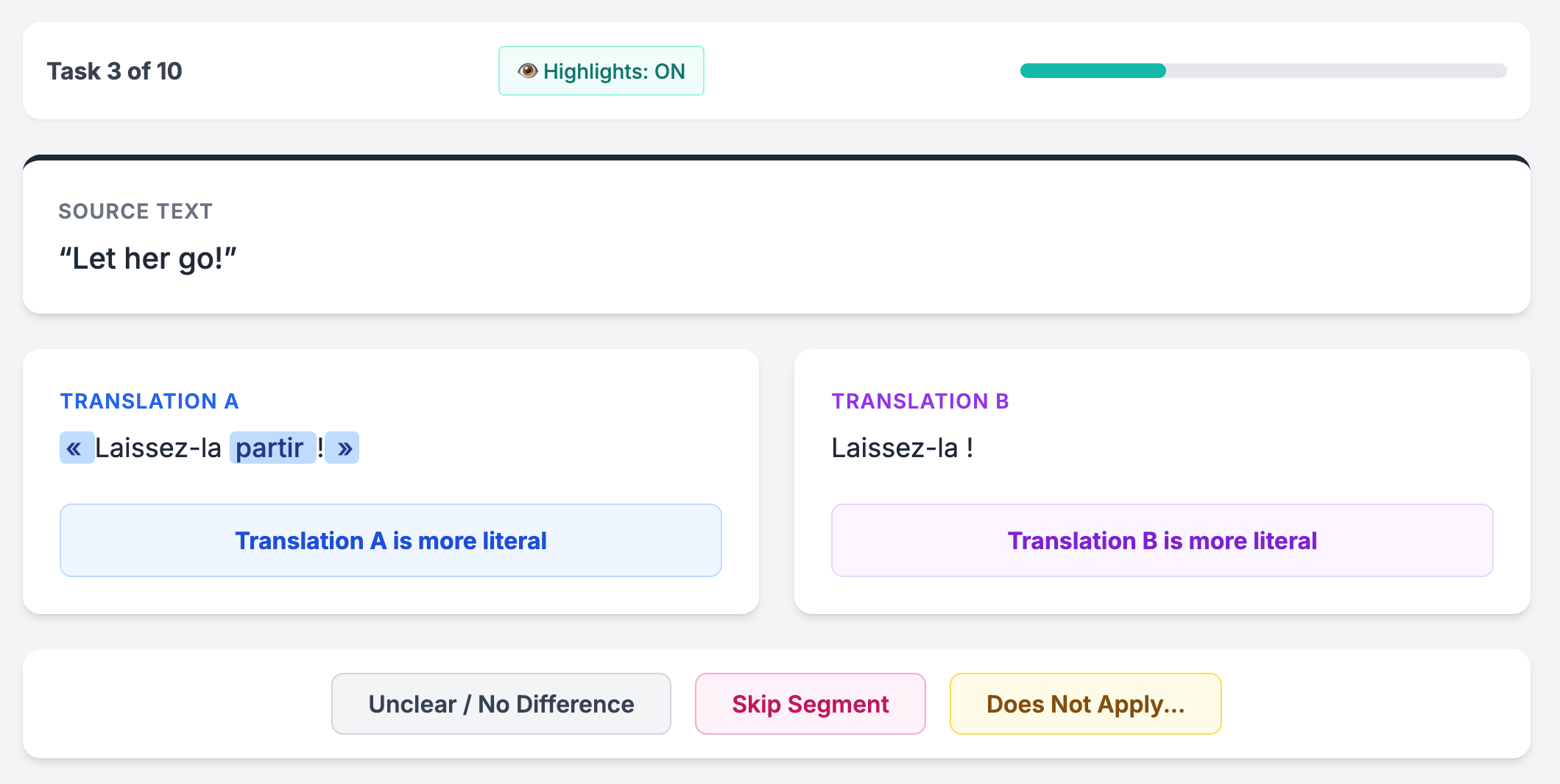}
    \caption{Annotation interface.}
    \label{fig:annotation_UI}
\end{figure}

\begin{figure}[h]
    \centering
    \includegraphics[width=0.96\columnwidth]{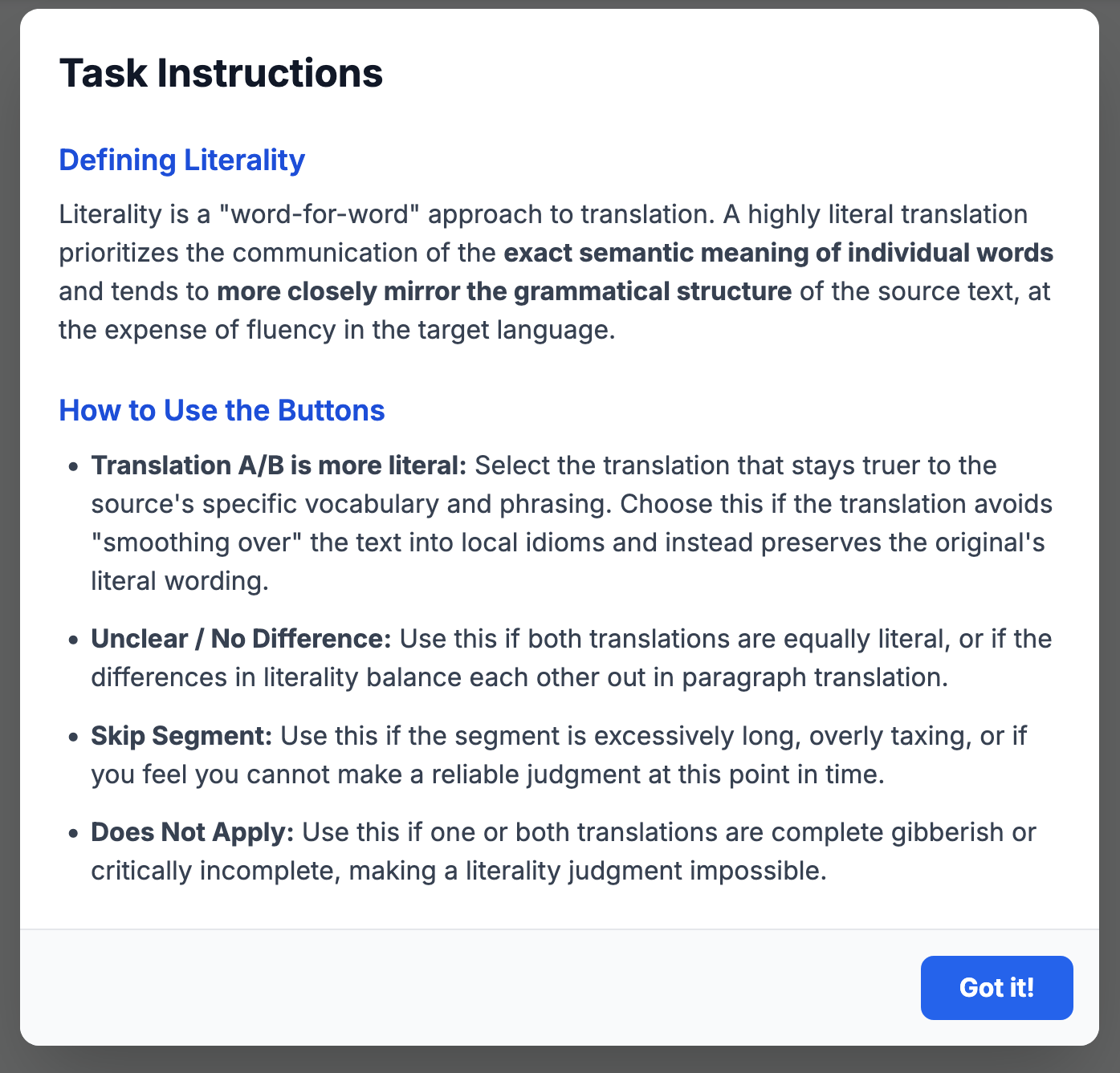}
    \caption{Annotation guidelines.}
    \label{fig:annotation_guidelines}
\end{figure}

To ground our automatic heuristics in human intuition, we conducted a manual annotation campaign on a subset of our actual dataset. We focused on the \texttt{en-fr\_FR} (English-French of France) data partition, evaluating pairs of translation hypotheses (such as the initial human translation, its post-edited counterpart and MT candidates). Segments with a MetricX-24 score higher than 7.0 were excluded from the annotation.

\paragraph{Annotation Setup and Guidelines}
Annotations were performed by the authors and their colleagues. Using a custom web interface (Figure \ref{fig:annotation_UI}), annotators were presented with an English source segment alongside two translation hypotheses. To prevent bias, annotators were completely blinded to the origin of the translations (i.e., MT vs. Human). As detailed in the annotation guidelines (Figure \ref{fig:annotation_guidelines}), annotators were instructed to select the translation that more closely mirrored the grammatical structure and specific vocabulary of the source text in a ``word-for-word'' fashion.

\paragraph{Inter-Annotator Agreement}
Given that literality is a notoriously subjective and relative concept in translation studies \cite{shuttleworth1997dictionary,chestermanReflections}, inter-annotator agreement was surprisingly high. We recorded a raw agreement rate of $74.6\%$, corresponding to a Cohen’s Kappa \cite{Cohen1960ACO} of $\kappa = 0.5322$, indicating moderate agreement \cite{landisKoch}.

\paragraph{Algorithmic vs. Human Reliability}
During the ``post-mortem'' analysis of instances where human annotators disagreed, we observed that the verdicts of the automatic heuristics consistently aligned with principled, structural definitions of literality, breaking the tie in the eyes of the annotator. This suggests that automatic algorithms are highly competitive with, and likely superior to, human annotators in consistently detecting literality shifts. This aligns with previous findings in the broader literature on translationese, where automatic classifiers frequently outperform human judges in identifying translational artifacts \citep[e.g.,][]{baroninewapp,van-halteren-2008-source}.

\paragraph{Correlation with Human Judgements}
Table~\ref{tab:signal_human_corr} details the Spearman correlations between our individual heuristics and the human pairwise judgements. The embedding-based metrics—particularly \textsc{SegSemLaBSE} and \textsc{TokSimRaw}—exhibit robust, statistically significant correlations with human intuition ($p<0.001$). Crucially, these semantic signals remain highly discriminative and stable across increasingly strict MetricX-24 quality thresholds, as well as across varying source segment lengths, validating their heavy weighting in the final SLI formulation.

\begin{table*}[h]
\centering
\begin{adjustbox}{max width=\linewidth, max totalheight=\textheight}
\renewcommand{\arraystretch}{1.05}
\setlength{\tabcolsep}{4pt}
\small

\begin{tabular}{l llll lll lll lll}
\toprule

\textbf{Signal}
 & \multicolumn{4}{c}{\textit{MetricX quality filter}}
 & \multicolumn{3}{c}{\textit{Long segments}}
 & \multicolumn{3}{c}{\textit{Medium segments}}
 & \multicolumn{3}{c}{\textit{Short segments}}
\\
\cmidrule(lr){2-5}
\cmidrule(lr){6-8}
\cmidrule(lr){9-11}
\cmidrule(lr){12-14}

 & ${\leq}7.0$
 & ${\leq}5.0$
 & ${\leq}3.0$
 & ${\leq}1.5$
 & \textit{lg}$\Delta$
 & \textit{md}$\Delta$
 & \textit{sm}$\Delta$
 & \textit{lg}$\Delta$
 & \textit{md}$\Delta$
 & \textit{sm}$\Delta$
 & \textit{lg}$\Delta$
 & \textit{md}$\Delta$
 & \textit{sm}$\Delta$
\\
\midrule

\textsc{Density}
 & \CML{$0.249^{***}$} & \CML{$0.254^{***}$}
 & \CWK{$0.278^{***}$} & \CWK{$0.259^{*}$}
 & \CZE{$-0.107$}   & \CWK{$0.249$}   & \CSN{$-0.492$}
 & \CML{$0.340^{*}$} & \CWK{$0.101$}   & \CWN{$-0.260$}
 & \CML{$0.394^{***}$} & \CWK{$0.108$}   & \CWN{$-0.213$}
\\

\textsc{PosSim}
 & \CWK{$0.284^{***}$} & \CWK{$0.273^{***}$}
 & \CML{$0.308^{***}$} & \CWK{$0.081$}
 & \CML{$0.399$}   & \CPA{$0.488^{**}$} & \CPA{$0.534$}
 & \CML{$0.428^{**}$} & \CWK{$0.204^{*}$} & \CWK{$0.090$}
 & \CPA{$0.470^{***}$} & \CZE{$0.022$}   & \CZE{$-0.014$}
\\

\textsc{SegSemLaBSE}
 & \CPA{$0.586^{***}$} & \CPA{$0.578^{***}$}
 & \CPA{$0.603^{***}$} & \CML{$0.463^{***}$}
 & \CPA{$0.505^{*}$} & \CWK{$0.274$}   & \CML{$0.417$}
 & \CML{$0.464^{**}$} & \CPA{$0.519^{***}$} & \CPA{$0.585^{***}$}
 & \CPA{$0.720^{***}$} & \CPA{$0.748^{***}$} & \CWK{$0.158$}
\\

\textsc{TokSimPen}
 & \CML{$0.399^{***}$} & \CML{$0.395^{***}$}
 & \CML{$0.386^{***}$} & \CWK{$0.273^{*}$}
 & \CZE{$-0.122$}   & \CML{$0.408^{*}$} & \CZE{$-0.129$}
 & \CML{$0.460^{**}$} & \CWK{$0.198$}   & \CWK{$0.058$}
 & \CML{$0.386^{***}$} & \CML{$0.374^{***}$} & \CZE{$0.045$}
\\

\textsc{TokSimRaw}
 & \CML{$0.404^{***}$} & \CML{$0.389^{***}$}
 & \CWK{$0.294^{***}$} & \CWK{$0.146$}
 & \CWK{$0.087$}   & \CML{$0.376^{*}$} & \CWK{$0.101$}
 & \CPA{$0.557^{***}$} & \CWK{$0.249^{*}$} & \CML{$0.352^{**}$}
 & \CML{$0.395^{***}$} & \CPA{$0.469^{***}$} & \CML{$0.356$}
\\

\textsc{TreeSim}
 & \CWK{$0.254^{***}$} & \CWK{$0.285^{***}$}
 & \CWK{$0.290^{***}$} & \CWK{$0.196$}
 & \CWK{$0.227$}   & \CWK{$0.218$}   & \CZE{$0.024$}
 & \CWK{$0.185$}   & \CWK{$0.280^{**}$} & \CWK{$0.137$}
 & \CML{$0.415^{***}$} & \CWK{$0.211$}   & \CWN{$-0.330$}
\\

\midrule
$N$
 & 436 & 367 & 216 & 82
 & 24 & 33 & 13
 & 41 & 94 & 55
 & 74 & 76 & 26
\\
\bottomrule
\end{tabular}
\end{adjustbox}

\caption{Spearman correlations between heuristic literality signals and human
 pairwise literality judgements on a sample of \textit{en}--\textit{fr}
 translations drawn from the main dataset ($N=436$ pairs).
 \textbf{Columns 1--4} (\textit{MetricX quality filter}):
 correlation on subsets retaining only pairs where both hypotheses achieve a MetricX-24 score below threshold $\tau \in
 \{7.0,\,5.0,\,3.0,\,1.5\}$; smaller $\tau$ selects progressively
 higher-quality pairs.
 \textbf{Columns 5--13} (\textit{Long / Medium / Short segments}):
 correlations broken down jointly by source segment length and by the magnitude of the \textsc{TokSimPen} difference between the two
 hypotheses (\textit{lg}$\Delta$ = large, \textit{md}$\Delta$ = medium,
 \textit{sm}$\Delta$ = small), capturing how discriminable the two candidates are according to that signal.
 Significance: $^{***}p<0.001$, $^{**}p<0.01$, $^{*}p<0.05$; unmarked cells are not significant ($p\geq0.05$).}
\label{tab:signal_human_corr}
\end{table*}

\section{Additional SLI score analysis}
\label{app:sliAdd}

\begin{figure*}[h!]
    \centering
    \includegraphics[width=1\linewidth]{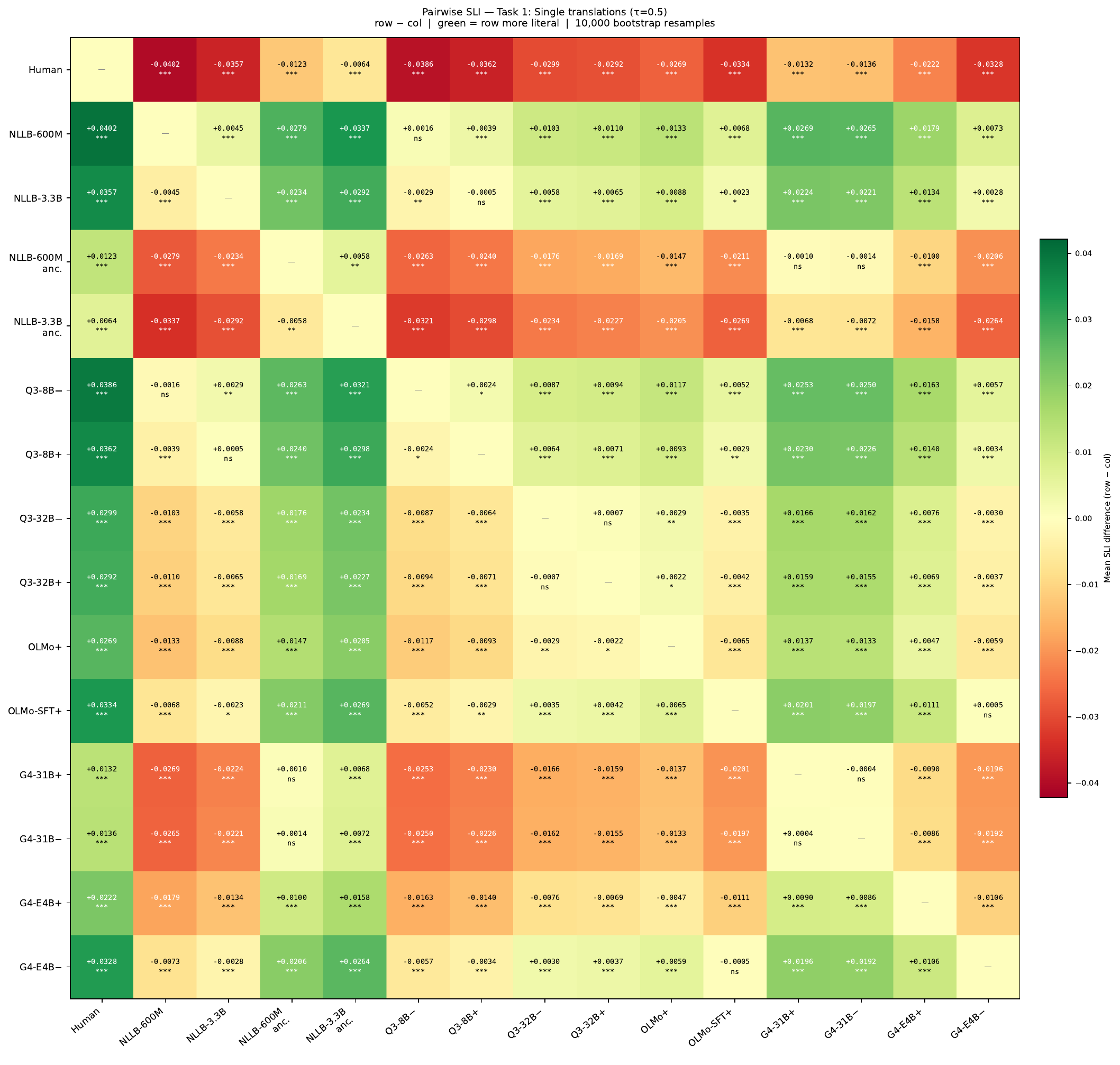}
\caption{Pairwise bootstrap mean SLI difference — Task~1 single translations (row$-$col, 10,000 resamples, segment-level, $\tau=0.5$, per-LP normalisation). Positive = row system more literal. $^{***}p<0.001$, $^{**}p<0.01$, $^{*}p<0.05$.}    \label{fig:t1_sli_heatmap}
\end{figure*}

\begin{figure*}[h!]
    \centering
    \includegraphics[width=1\linewidth]{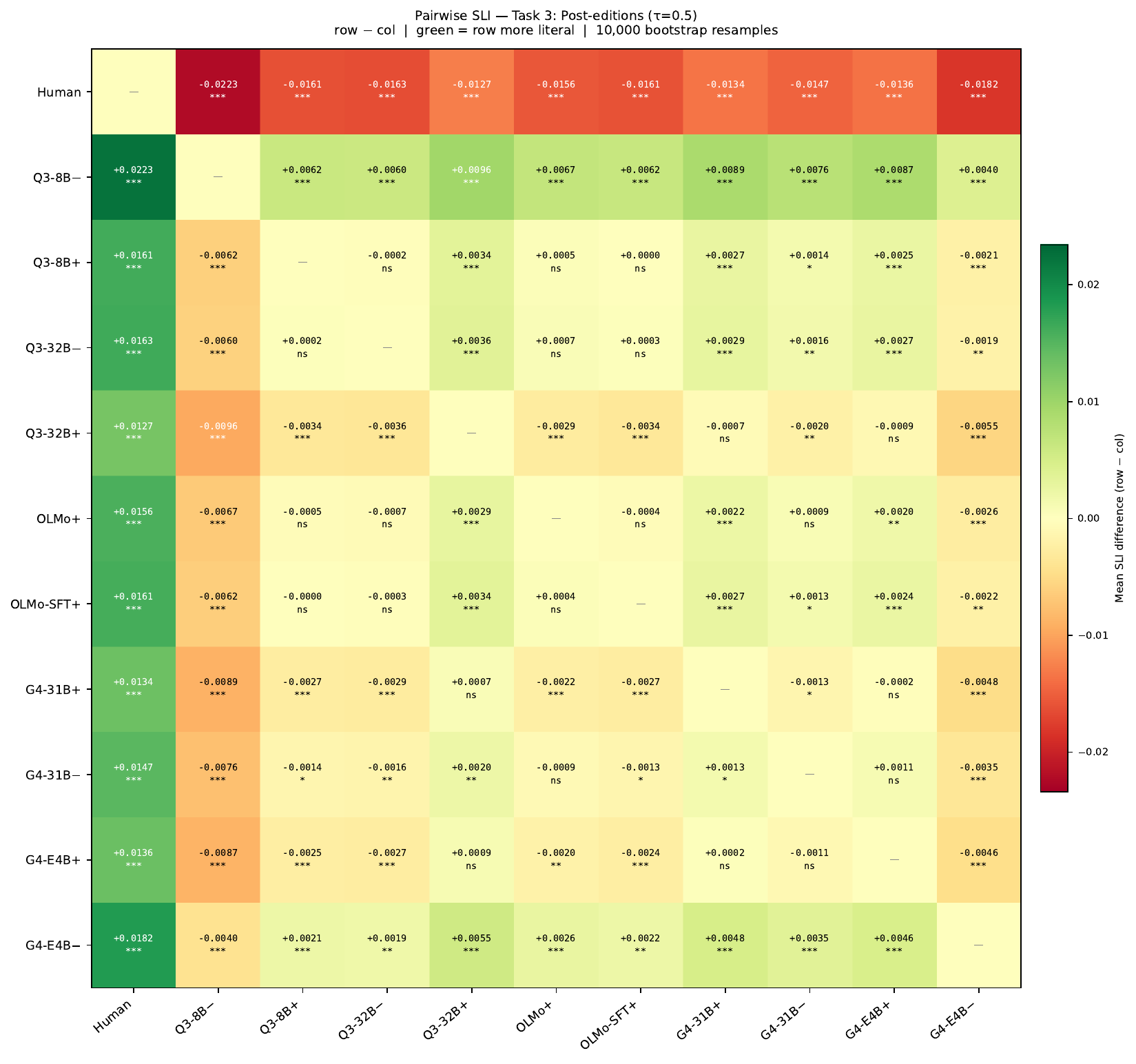}
\caption{Pairwise bootstrap mean SLI difference — Task~3 post-editions (row$-$col, 10,000 resamples, segment-level, $\tau=0.5$, per-LP normalisation). Positive = row system more literal. Human PE computed from WMT24++ \textit{target} literality. $^{***}p<0.001$, $^{**}p<0.01$, $^{*}p<0.05$.}    \label{fig:t3_sli_heatmap}
\end{figure*}

We present the per-language-pair SLI scores for Task 1 in Table \ref{tab:t1_sli_local_abs}, demonstrating the consistency of these patterns across 54 languages. To assess the statistical significance of the differences in literality between systems, Figure \ref{fig:t1_sli_heatmap} presents a pairwise comparison for Task 1, computing the mean difference in SLI and evaluating significance via bootstrap resampling ($N=10,000$). Similarly, Figure \ref{fig:t3_sli_heatmap} displays the pairwise significance testing for the post-editing setting (Task 3). Both heatmaps confirm that the observed differences in literality—particularly the human baseline's freer translations and Gemma-4's closer approximation to human output—are robustly significant across the models evaluated.

\begin{table*}[h]
\centering
\begin{adjustbox}{max width=\textwidth, max totalheight=\textheight}
\renewcommand{\arraystretch}{0.85}
\setlength{\tabcolsep}{3pt}
\small
\begin{tabular}{lrrrrrrrrrrrrrrr}
\toprule
\textbf{LP} & \textbf{Human} & \textbf{NLLB-600M} & \textbf{NLLB-3.3B} & \textbf{NLLB-600M$^{\text{anc}}$} & \textbf{NLLB-3.3B$^{\text{anc}}$} & \textbf{Qwen3-8B$^{-}$} & \textbf{Qwen3-8B$^{+}$} & \textbf{Qwen3-32B$^{-}$} & \textbf{Qwen3-32B$^{+}$} & \textbf{OLMo-32B$^{+}$} & \textbf{OLMo-32B-SFT$^{+}$} & \textbf{Gemma4-31B$^{+}$} & \textbf{Gemma4-31B$^{-}$} & \textbf{Gemma4-E4B$^{+}$} & \textbf{Gemma4-E4B$^{-}$} \\
\midrule
en-no\_NO & \cellcolor[rgb]{0.997,0.697,0.477}\textcolor[rgb]{0,0,0}{0.6856} & \cellcolor[rgb]{0.992,0.884,0.640}\textcolor[rgb]{0,0,0}{0.7208} & \cellcolor[rgb]{0.991,0.906,0.661}\textcolor[rgb]{0,0,0}{0.7252} & \cellcolor[rgb]{0.997,0.734,0.505}\textcolor[rgb]{0,0,0}{0.6919} & \cellcolor[rgb]{0.997,0.712,0.488}\textcolor[rgb]{0,0,0}{0.6881} & \cellcolor[rgb]{0.995,0.834,0.592}\textcolor[rgb]{0,0,0}{0.7117} & \cellcolor[rgb]{0.995,0.841,0.599}\textcolor[rgb]{0,0,0}{0.7133} & \cellcolor[rgb]{0.996,0.806,0.566}\textcolor[rgb]{0,0,0}{0.7058} & \cellcolor[rgb]{0.996,0.791,0.553}\textcolor[rgb]{0,0,0}{0.7028} & \cellcolor[rgb]{0.997,0.697,0.477}\textcolor[rgb]{0,0,0}{0.6861} & \cellcolor[rgb]{0.995,0.820,0.579}\textcolor[rgb]{0,0,0}{0.7083} & \cellcolor[rgb]{0.997,0.734,0.505}\textcolor[rgb]{0,0,0}{0.6930} & \cellcolor[rgb]{0.997,0.748,0.517}\textcolor[rgb]{0,0,0}{0.6954} & \cellcolor[rgb]{0.997,0.770,0.535}\textcolor[rgb]{0,0,0}{0.6994} & \cellcolor[rgb]{0.995,0.820,0.579}\textcolor[rgb]{0,0,0}{0.7092} \\
\bottomrule
\end{tabular}
\end{adjustbox}
\caption{\small SLI (per-LP standardization) — Task~1 single translations (MetricX$\leq$5.0, $\tau=0.5$). Higher = more literal.}
\label{tab:t1_sli_local_abs}
\end{table*}

\section{Post-Editing Dynamics by Domain}
\label{app:pe_dynamics}

Table~\ref{tab:pe_dynamics_fused} reports post-editing dynamics for each system across the four text domains of WMT24++ (news, social media, speech, literary), restricted to segments where the initial human translation passed the MetricX-24 $\leq 5.0$ quality filter. For each system we report the number of segments left unchanged versus altered, and, among altered segments, the proportion classified as deliteralizing (SLI decreases), reliteralizing (SLI increases), or neutral (no change in SLI beyond a threshold $\epsilon = 0.005$). The value $\epsilon = 0.005$ was chosen to be conservative relative to the SLI's $[0, 1]$ range, ensuring that edits classified as deliteralizing or reliteralizing reflect substantive shifts rather than numerical noise.

\begin{table}[h]
\centering\small\setlength{\tabcolsep}{3pt}
\renewcommand{\arraystretch}{1.05}
\begin{adjustbox}{max width=\columnwidth}
\begin{tabular}{llrrrrrrrrrr}
\toprule
\textbf{Domain} & \textbf{System} & \multicolumn{1}{c}{$n$} & \multicolumn{2}{c}{\textbf{Unchanged}} & \multicolumn{2}{c}{\textbf{Altered}} & \multicolumn{2}{c}{\textbf{Delit.}} & \multicolumn{2}{c}{\textbf{Relit.}} & \multicolumn{1}{c}{\textbf{Neut.}} \\
\cmidrule(lr){4-5}\cmidrule(lr){6-7}\cmidrule(lr){8-9}\cmidrule(lr){10-11}
 & & & $n$ & \% & $n$ & \% & $n$ & \% & $n$ & \% & \% \\
\midrule
\multirow{11}{*}{\textit{All domains}} & Human & 44,525 & 21,267 & \cellcolor[rgb]{0.135,0.545,0.554}\textcolor[rgb]{1,1,1}{47.8} & 23,258   & \cellcolor[rgb]{0.123,0.585,0.547}\textcolor[rgb]{1,1,1}{52.2} & 10,271 & \cellcolor[rgb]{0.148,0.512,0.557}\textcolor[rgb]{1,1,1}{44.2} & 6,769 & \cellcolor[rgb]{0.211,0.364,0.552}\textcolor[rgb]{1,1,1}{29.1} & \cellcolor[rgb]{0.222,0.339,0.549}\textcolor[rgb]{1,1,1}{26.7} \\
 & Qwen3-8B$^{-}$ & 40,971 & 936 & \cellcolor[rgb]{0.274,0.031,0.359}\textcolor[rgb]{1,1,1}{2.3} & 40,035   & \cellcolor[rgb]{0.946,0.900,0.113}\textcolor[rgb]{0,0,0}{97.7} & 20,988 & \cellcolor[rgb]{0.122,0.589,0.546}\textcolor[rgb]{1,1,1}{52.4} & 15,750 & \cellcolor[rgb]{0.167,0.464,0.558}\textcolor[rgb]{1,1,1}{39.3} & \cellcolor[rgb]{0.283,0.121,0.441}\textcolor[rgb]{1,1,1}{8.2} \\
 & Qwen3-8B$^{+}$ & 39,955 & 660 & \cellcolor[rgb]{0.273,0.026,0.353}\textcolor[rgb]{1,1,1}{1.7} & 39,295   & \cellcolor[rgb]{0.955,0.901,0.118}\textcolor[rgb]{0,0,0}{98.3} & 21,728 & \cellcolor[rgb]{0.119,0.615,0.538}\textcolor[rgb]{1,1,1}{55.3} & 14,406 & \cellcolor[rgb]{0.177,0.438,0.558}\textcolor[rgb]{1,1,1}{36.7} & \cellcolor[rgb]{0.283,0.116,0.436}\textcolor[rgb]{1,1,1}{8.0} \\
\bottomrule
\end{tabular}
\end{adjustbox}
\caption{Post-edition dynamics by domain (MetricX$\leq$5.0, neutral $\varepsilon=0.005$). \textit{Delit.}, \textit{Relit.}, \textit{Neut.}\ percentages are over altered segments only. Colour intensity proportional to percentage value.}
\label{tab:pe_dynamics_fused}
\end{table}

\end{document}